\documentclass[10pt,journal,compsoc]{IEEEtran}
\ifCLASSOPTIONcompsoc
  \usepackage[nocompress]{cite}
\else
  \usepackage{cite}
\fi
\ifCLASSINFOpdf
\else
\fi

\usepackage{mathrsfs}
\usepackage{color}
\usepackage{booktabs}
\usepackage{multirow}
\usepackage{threeparttable}
\usepackage{color, colortbl}
\usepackage{changes}
\usepackage{amssymb}

\usepackage{graphicx}
\usepackage{amsmath}

\usepackage[pagebackref=true,breaklinks=true,colorlinks,bookmarks=false,citecolor=blue,linkcolor=blue]{hyperref}

\newcommand{\pub}[1]{\color{gray}{\tiny{[{#1}]}}}
\newcommand{\highest}[1]{\color{red}{\textbf{{#1}}}}
\newcommand{\second}[1]{\color{blue}{\underline{{#1}}}}

\begin{document}
%
\title{DVIS++: Improved Decoupled Framework for Universal Video Segmentation}
%
%
%
%

\author{Tao~Zhang, Xingye~Tian, Yikang~Zhou, Shunping~Ji, Xuebo~Wang, Xin~Tao, Yuan~Zhang, \\ Pengfei~Wan, Zhongyuan Wang, and~Yu~Wu
	
\IEEEcompsocitemizethanks{\IEEEcompsocthanksitem Tao Zhang, Yikang Zhou, Shunping Ji and Yu Wu are with the Wuhan University, Wuhan, 430072, China.
\protect\\
	E-mail: \{zhang\_tao, zhouyik, jishunping, wuyucs\}@whu.edu.cn.
	\IEEEcompsocthanksitem Xingye Tian, Xuebo Wang, Xin Tao, Yuan Zhang, Pengfei Wan and Zhongyuan Wang are with the Y-tech, Kuaishou Technology, Beijing, 100085, China.\protect\\
    E-mail: \{tianxingye, wangxuebo, taoxin, zhangyuan03, wanpengfei, wangzhongyuan\}@kuaishou.com.
    \IEEEcompsocthanksitem  Corresponding authors: Shunping Ji.
    \IEEEcompsocthanksitem This work was supported  by the National Natural Science Foundation of China under Grant 42171430 and 42030102.}
}

\IEEEtitleabstractindextext{%
\begin{abstract}
We present the \textbf{D}ecoupled \textbf{VI}deo \textbf{S}egmentation (DVIS) framework, a novel approach for the challenging task of universal video segmentation, including video instance segmentation (VIS), video semantic segmentation (VSS), and video panoptic segmentation (VPS). Unlike previous methods that model video segmentation in an end-to-end manner, our approach decouples video segmentation into three cascaded sub-tasks: segmentation, tracking, and refinement. This decoupling design allows for simpler and more effective modeling of the spatio-temporal representations of objects, especially in complex scenes and long videos. Accordingly, we introduce two novel components: the referring tracker and the temporal refiner. These components track objects frame by frame and model spatio-temporal representations based on pre-aligned features. To improve the tracking capability of DVIS, we propose a denoising training strategy and introduce contrastive learning, resulting in a more robust framework named DVIS++. Furthermore, we evaluate DVIS++ in various settings, including open vocabulary and using a frozen pre-trained backbone. By integrating CLIP with DVIS++, we present OV-DVIS++, the first open-vocabulary universal video segmentation framework. We conduct extensive experiments on six mainstream benchmarks, including the VIS, VSS, and VPS datasets. Using a unified architecture, DVIS++ significantly outperforms state-of-the-art specialized methods on these benchmarks in both close- and open-vocabulary settings. Code:~\url{https://github.com/zhang-tao-whu/DVIS_Plus}.
\end{abstract}

}

\maketitle

\IEEEdisplaynontitleabstractindextext

%
\IEEEpeerreviewmaketitle

\IEEEraisesectionheading{\section{Introduction}\label{sec:introduction}}

\IEEEPARstart{V}{ideo} segmentation is a fundamental task in computer vision, playing a significant role in video understanding, video editing, and autonomous driving~\cite{zhang2016instance}, among other applications~\cite{zhou2022survey}. Most studies, such as~\cite{yang2019video}, \cite{wu2022seqformer}, and \cite{weng2023mask}, focus on designing specialized architectures for specific subdomains of video segmentation. A few researches, \cite{athar2023tarvis} and \cite{li2023tube}, introduce unified architectures but significantly underperform the specialized architectures. Therefore, in this paper, we concentrate on universal video segmentation and design an efficient unified architecture that can achieve state-of-the-art performance across various segmentation tasks, including video instance segmentation (VIS)~\cite{yang2019video}, video semantic segmentation (VSS), and video panoptic segmentation (VPS)~\cite{kim2020video}. Universal video segmentation requires the simultaneous tracking, segmentation, and identification of all instance-level ``\textit{thing}" objects (\textit{e.g}., person, dog) and semantic-level ``\textit{stuff}" objects (\textit{e.g}., sky, road) in the video. In this paper, we achieve universal video segmentation by adopting a unified perspective to describe both ``thing" objects and ``stuff" objects.

\begin{figure}[t!]
    \centering
	\includegraphics[width=0.48\linewidth]{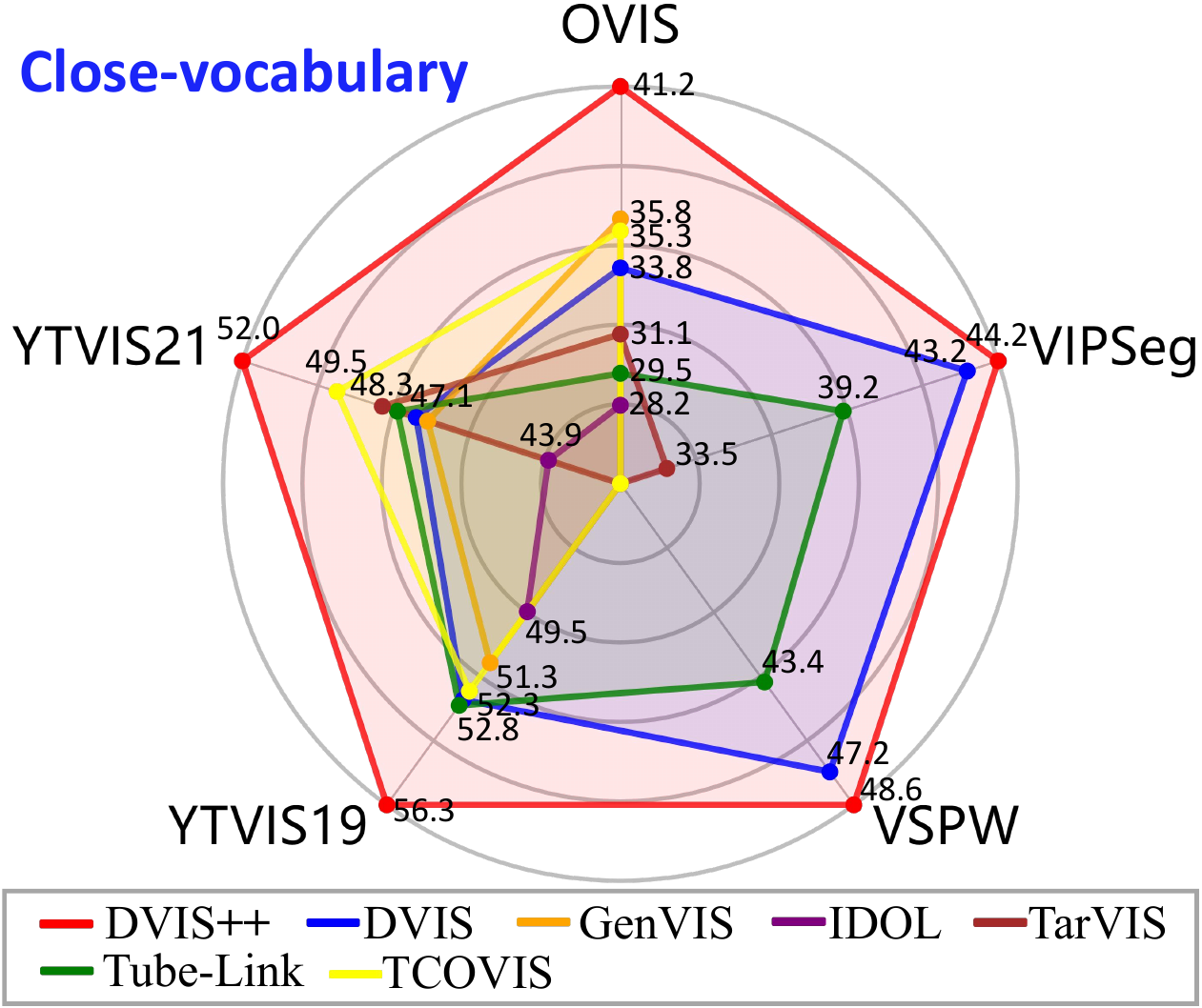}
	\includegraphics[width=0.48\linewidth]{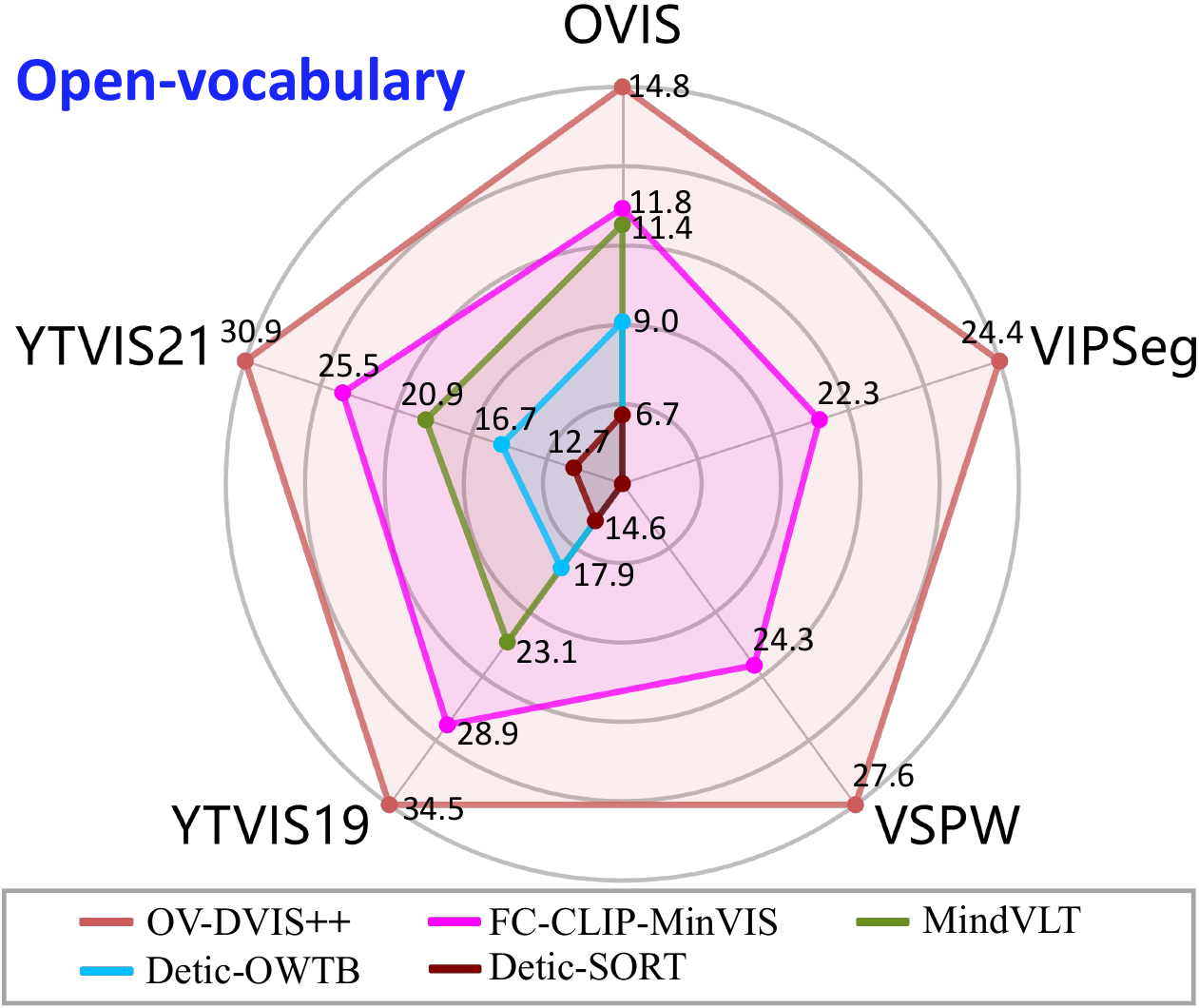}
    \caption{DVIS++ features a uniform architecture that significantly outperforms other methods in VIS, VSS, and VPS tasks, with all methods utilizing a ResNet-50 backbone (left). OV-DVIS++ showcases strong zero-shot video segmentation capabilities, surpassing current SOTA methods in open vocabulary video segmentation (right).}
    \label{fig:radar}
\end{figure}

In the VIS community, previous methods \cite{yang2019video}, \cite{wang2021end}, \cite{hwang2021video}, \cite{wu2022seqformer}, \cite{cheng2021mask2former}, \cite{heo2022vita}, \cite{huang2022minvis}, \cite{wu2022defense}, \cite{heo2023generalized}, \cite{ying2023ctvis} can be divided into two technical pipelines: offline pipeline (\textit{i.e.}, processing the entire video at once) and online pipeline (\textit{i.e.}, processing the video frame by frame). The offline pipeline focuses on studying how to effectively utilize the spatio-temporal information in videos, while the online pipeline focuses on studying how to associate objects frame-by-frame more stably and accurately. 

Offline methods such as VisTR \cite{wang2021end}, IFC \cite{hwang2021video}, SeqFormer~\cite{wu2022seqformer}, Mask2Former-VIS \cite{cheng2021mask2former}, and VITA \cite{heo2022vita} have designed various mechanisms to extract stronger spatio-temporal features of objects. These methods have achieved satisfactory results on simple, short videos \cite{yang2019video}, but they face significant challenges when dealing with complex, long videos \cite{qi2022occluded}. This is because the complexity of object motion trajectories and probability of occlusion dramatically increase when dealing with complex, long videos, making it extremely challenging to directly model the spatio-temporal representation of objects from video features using prior queries. 

On the other hand, online methods like Mask Track R-CNN \cite{yang2019video}, MinVIS \cite{huang2022minvis}, IDOL \cite{wu2022defense}, GenVIS \cite{heo2023generalized}, and CTVIS \cite{ying2023ctvis} focus on obtaining more discriminative representations of objects and designing more powerful object association algorithms. These online methods have achieved good results on both simple, short videos and complex, long videos. However, it is evident that these methods have not attempted to model the long-term spatio-temporal representation of objects. Therefore, there is still room for improvement in these online methods.

In this paper, we rethink the problems in modern video segmentation and propose a solution to overcome these challenges by decoupling the video segmentation task. Directly modeling the spatio-temporal representation of objects from the image features of all frames presents unimaginable challenges and often leads to failure when dealing with complex, long videos. However, if given temporally pre-aligned object representations, modeling the spatio-temporal representation of objects becomes much easier. Therefore, we propose dividing the video segmentation task into three sub-tasks: segmentation, tracking, and refinement. The difficulty of these sub-tasks is independent of the video's length and complexity. Segmentation aims to extract all objects of interest, including ``\textit{thing}" and ``\textit{stuff}" elements, and obtain their representations from a single frame. Tracking seeks to establish the association of the same object across adjacent frames, significantly reducing complexity compared to the direct association across all frames as applied in previous offline methods. Refinement optimizes both segmentation and association results by utilizing the pre-aligned temporal
information of the object.

The segmentation subtask has been well addressed by works on image segmentation \cite{cheng2021per}, \cite{cheng2022masked}, and \cite{li2023mask}. This paper focuses on designing an effective tracker and refiner for the more challenging subtasks of tracking and refinement. 

We propose a learnable tracker, termed the referring tracker, which models the tracking task as a reconstruction or denoising process. Specifically, by utilizing the representation of a specific object from the previous frame as a reference, the referring tracker generates the representation of the same object based on the object representations outputted by the segmenter in the current frame. The pre-aligned object representations between frames can be obtained through the referring tracker by following the above pipeline.

Additionally, we construct a temporal refiner using simple, naive self-attention \cite{vaswani2017attention} and 1D convolution to effectively model the spatio-temporal representations of objects based on the pre-aligned representations. 

Finally, we introduce the \textbf{D}ecoupled \textbf{VI}deo \textbf{S}egmentation (DVIS) framework by cascading the segmenter, referring tracker, and temporal refiner. This framework enables convenient and efficient modeling of spatio-temporal representations of objects and ultimately outperforms all contemporary methods \cite{huang2022minvis}, \cite{wu2022defense}, \cite{heo2023generalized}, \cite{athar2023tarvis}, and \cite{li2023tube}.

Furthermore, we have proposed DVIS++, which improves our previous conference work DVIS \cite{zhang2023dvis} on tracking capabilities. The tracking sub-task is a prerequisite for the refinement sub-task, as good tracking results are fundamental for effectively modeling the spatio-temporal representations of objects. To further enhance the tracking ability of DVIS, we have introduced a denoising training strategy to simulate challenging cases and incorporated contrastive learning to obtain more discriminative object representations.

Specifically, the denoising training strategy, including three noise simulation approaches, mimics the most challenging identification swapping problems in video segmentation. The introduction of the denoising training strategy significantly enhances the tracking capability of the learnable referring tracker. Moreover, effective contrastive learning strategies are newly designed for the referring tracker and the temporal refiner.  

We further validate the effectiveness and universality of DVIS++ under various settings. Initially, we investigate whether DVIS++ remains effective when not allowed to finetune the pre-trained backbone. To this end, we introduce a vision foundation model pre-trained with DINOv2~\cite{oquab2023dinov2}. Secondly, we test our method in an open-vocabulary setting. By integrating CLIP~\cite{radford2021learning} with DVIS++, we create OV-DVIS++, the first open-vocabulary universal video segmentation framework.

We conducted extensive experiments on six datasets, including OVIS \cite{qi2022occluded}, YouTube VIS 2019/2021/2022 \cite{yang2019video}, VIPSeg \cite{miao2022large}, and VSPW \cite{miao2021vspw}, to validate the effectiveness of the proposed DVIS and DVIS++. As shown in Figure~\ref{fig:radar}, DVIS++ achieves comprehensive improvements compared to DVIS and outperforms previous state-of-the-art (SOTA) methods on all six benchmarks. It is worth noting that we have secured the championship in the \textbf{PVUW Challenge} at CVPR 2023 \cite{zhang20231stvps} and the \textbf{LSVOS Challenge} at ICCV 2023 \cite{zhang20231stvis}, utilizing only a subset of the strategies presented in this paper. OV-DVIS++ enables open-vocabulary universal video segmentation, achieving a significant improvement in zero-shot performance on VIS datasets compared to previous SOTA methods, as illustrated on the right side of Figure~\ref{fig:radar}. We believe that DVIS, DVIS++ and OV-DVIS++ can serve as solid baselines in the field of video segmentation.

To summarize, our contributions are as follows:

\begin{enumerate}
  \item Through rethinking problems in modern video segmentation, we propose a decoupling strategy to better model objects' spatio-temporal representations. In line with this strategy, we design DVIS for universal video segmentation, which includes a segmenter, a novel referring tracker, and a novel temporal refiner.
  
  \item To enhance the tracking capability, the foundation of modeling spatio-temporal representations, we incorporate a denoising training strategy and contrastive learning into DVIS, resulting in an improved version, DVIS++. Compared to DVIS, DVIS++ demonstrates enhanced robustness in tracking and segmentation capabilities.
  
  \item We validate the effectiveness and universality of DVIS++ under various settings, including freezing pre-trained backbone and open-vocabulary settings. We introduce the first open-vocabulary universal video segmentation framework, OV-DVIS++, which achieves SOTA performance in the open-vocabulary setting. Furthermore, when utilizing a frozen pre-trained backbone, DVIS++ also works well and shows remarkable performance.
\end{enumerate}

\section{Related Work}
\subsection{Video Segmentation}
\textbf{Specialized Video Segmentation.} Video segmentation is a fundamental task in the field of computer vision. In the past, traditional methods were employed by \cite{bai2009video}, \cite{bai2009geodesic}, and \cite{mu2007automatic} to address video object segmentation and video matting. However, in recent years, deep learning has become the mainstream approach in video segmentation and has achieved remarkable success. There are two distinct communities in video segmentation: video semantic segmentation (VSS) and video instance segmentation (VIS). These communities utilize different technique pipelines and focus on different challenges. VSS methods naturally associate the segmentation results of different frames based on semantic categories. Therefore, VSS methods do not need to address the problem of object association. Instead, the VSS methods \cite{miao2021vspw}, \cite{shelhamer2016clockwork}, \cite{sun2022coarse}, \cite{sun2022mining} and \cite{zhu2017deep} focus on improving the temporal consistency of the segmentation results. As a result, these VSS methods cannot be directly applied to universal video segmentation.

VIS faces the most challenging problem of consistently and accurately associating instance-level objects with the same identity in videos, given significant deformations, large-scale movements, and complex occlusions. The first VIS method, Mask Track R-CNN \cite{yang2019video}, achieves VIS by incorporating a tracking head onto Mask R-CNN \cite{he2017mask} and utilizing multiple cues to associate objects. Subsequently, SipMask \cite{cao2020sipmask}, and CrossVIS \cite{yang2021crossover} improve performance by introducing a more powerful segmenter and a crossover learning scheme, respectively. Additionally, MinVIS \cite{huang2022minvis} and IDOL \cite{wu2022defense} discover that query-based instance representations \cite{carion2020end} can be directly used for object association in adjacent frames, resulting in surprising performance. Furthermore, IDOL \cite{wu2022defense} and CTVIS \cite{ying2023ctvis} introduce contrastive learning into VIS to obtain more discriminative instance representations. GenVIS \cite{heo2023generalized} and GRAtt-VIS \cite{hannan2023gratt} design learnable trackers to handle object association, exhibiting better performance than naive heuristic algorithms.

In contrast to the aforementioned online VIS methods, offline VIS methods VisTR \cite{wang2021end}, IFC \cite{hwang2021video}, Mask2Former-VIS \cite{cheng2021mask2former}, SeqFormer \cite{wu2022seqformer}, and VITA \cite{heo2022vita} focus on enhancing the modeling of spatio-temporal representations of objects and have shown impressive performance on short and simple videos, but their effectiveness diminishes when confronted with complex and lengthy videos. The main reason behind this limitation is that these methods strive to link the same objects in all frames and model their spatio-temporal representations simultaneously, which becomes exceedingly difficult in the case of complex, long videos. As a result, it is not surprising that these VIS methods face challenges when dealing with complex, long videos \cite{qi2022occluded}. GenVIS \cite{heo2023generalized} and NOVIS \cite{meinhardt2023novis} adopt a different approach, decomposing long videos into multiple fixed-length video clips. The short clip typically contains fewer than eight frames. This strategy significantly reduces the temporal association challenge and thus enables the direct modeling of the spatio-temporal representation of the object within each clip. However, it is important to note that GenVIS and NOVIS do not offer significant advantages when limited spatio-temporal information is employed.

Our proposed methods, DVIS \cite{zhang2023dvis} and DVIS++, address all the challenges above through the decoupling design. Firstly, we associate objects frame by frame and subsequently obtain spatio-temporal representations based on pre-aligned object representations. Consequently, our method significantly outperforms all current methods on complex, long videos.

\vspace{2mm}\noindent\textbf{Universal Video Segmentation.} Thanks to the success of universal image segmentation methods such as \cite{wang2021max}, \cite{zhang2021k}, \cite{cheng2021per}, \cite{cheng2022masked}, and \cite{yu2022k}, universal video segmentation methods like Video K-Net \cite{li2022video}, TubeFormer \cite{kim2022tubeformer}, Tube-Link \cite{li2023tube}, TarVIS \cite{athar2023tarvis}, and DVIS \cite{zhang2023dvis} have achieved comparable or even higher performance compared to specialized methods. Among these methods, Video K-Net achieves video segmentation by tracking and associating object kernels, while TubeFormer, Tube-Link, and TarVIS perform video segmentation by directly segmenting objects within clips and associating segmentation results between clips through overlapping frames or similarity clues. However, all the aforementioned methods utilize heuristic algorithms for object association and only leverage temporal information within short clips. In contrast, our proposed DVIS and DVIS++ employ a trainable referring tracker for more elegant and efficient object association. Additionally, they effectively model the spatio-temporal representation of objects throughout the entire video using a temporal refiner.

\subsection{Image Segmentation}

\begin{figure*}[t!]
    \centering
    \includegraphics[width=0.90\linewidth]{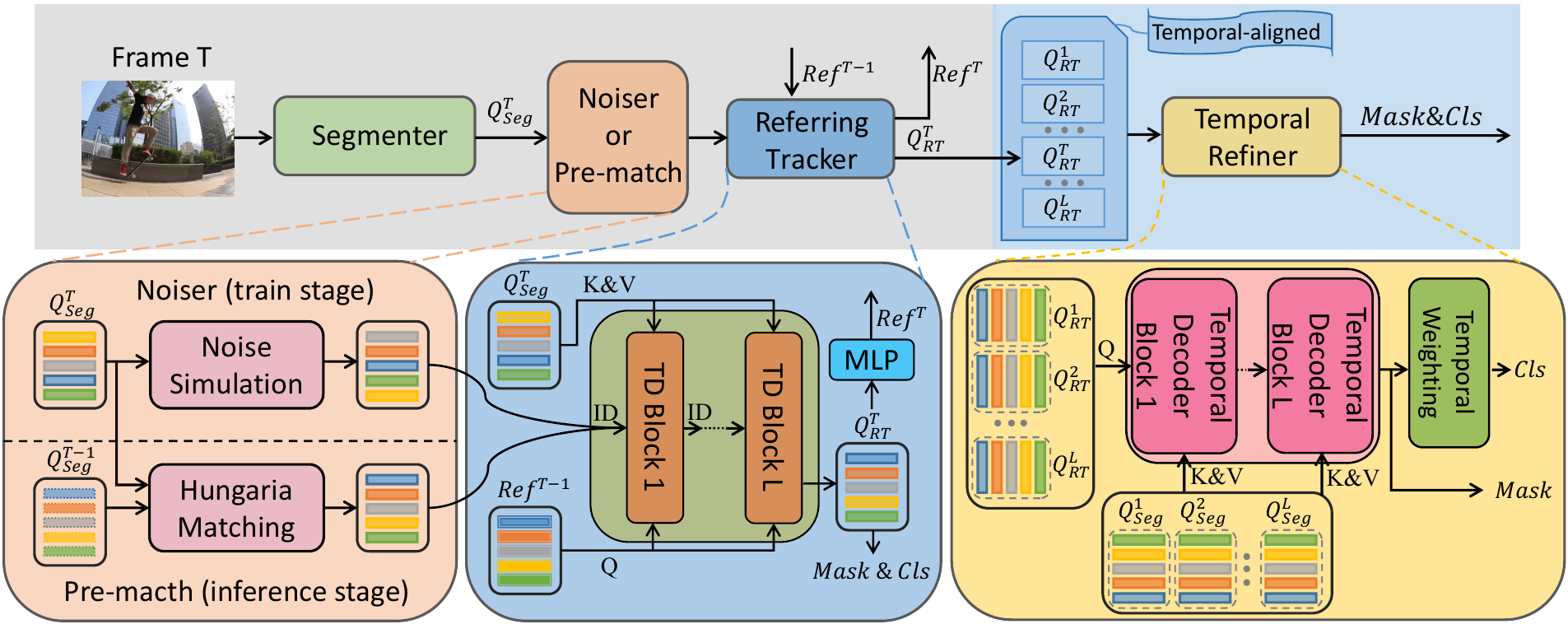} \vspace{-2mm}
    \caption{An \textbf{overview} of DVIS++. DVIS++ comprises three trainable components, namely the Segmenter, Referring Tracker, and Temporal Refiner, as well as a non-trainable component called the Noiser. The Segmenter extracts object representations ($Q_{Seg}$) from each frame. During the training process, the Noiser introduces noise to $Q_{Seg}$. The Referring Tracker generates the corresponding object representation $Q_{RT}^{T}$ for the current frame based on the given reference $Ref^{T-1}$, and it also outputs a new reference $Ref^{T}$ for the next frame. Lastly, the Temporal Refiner constructs the spatio-temporal representation of the object by utilizing the temporal-aligned object representations $\{Q_{RT}^{t} | t\in [1, L]\}$ of all frames. Objects with the same identity index are represented using the same color.}\vspace{-2mm}
    \label{fig:overview}
\end{figure*}

\textbf{Universal Image Segmentation.} Thanks to the success of the transformer \cite{vaswani2017attention} in image detection \cite{carion2020end}, \cite{zhang2022dino} and the emergence of more generic representation forms \cite{tian2020conditional}, \cite{bolya2019yolact} compared to per-pixel classification maps, many universal image segmentation methods \cite{zhang2021k}, \cite{cheng2021per}, \cite{cheng2022masked}, and \cite{li2023mask} have been able to unify semantic segmentation and instance segmentation by adopting the same representation approach. \cite{jain2023oneformer}, \cite{zhang2023simple}, and \cite{li2023semantic} achieve more capabilities, such as open-vocabulary segmentation and multi-granularity segmentation, while implementing universal image segmentation. In this paper, DVIS and DVIS++ adopt the classical universal segmentation method Mask2Former \cite{cheng2022masked} as the segmenter.

\vspace{2mm}\noindent\textbf{Open-Vocabulary Image Segmentation.} Open-vocabulary segmentation aims to segment objects of any category, including those not present in the training set. Recent works \cite{ghiasi2022scaling}, \cite{xu2022simple}, \cite{liang2023open}, \cite{ding2022decoupling}, \cite{xu2022groupvit}, \cite{zhou2022extract}, \cite{zou2023generalized} have utilized large-scale visual language models \cite{radford2021learning}, \cite{jia2021scaling}, \cite{rombach2022high} to achieve open-vocabulary semantic segmentation. MaskCLIP \cite{zhou2022extract} combines a class-agnostic mask proposal network with a frozen CLIP \cite{radford2021learning} encoder to achieve open-vocabulary panoptic segmentation. DenseCLIP \cite{rao2022denseclip} achieves impressive zero-shot segmentation performance by fine-tuning the CLIP encoder. ODISE \cite{xu2023open} utilizes a pre-trained text-image diffusion model as a backbone to achieve open-vocabulary segmentation. OpenSeed \cite{zhang2023simple} achieves more powerful open-vocabulary segmentation performance by combining detection and segmentation datasets. Recently, FC-CLIP \cite{yu2023convolutions} has achieved surprising open-vocabulary segmentation performance by training a mask generation decoder on the frozen CLIP encoder. In this paper, OV-DVIS++ is inspired by FC-CLIP and adopts the same approach to achieve the leading performance of open-vocabulary video segmentation.

\section{METHODOLOGY}

\textbf{The overview of DVIS++.} As shown in Figure~\ref{fig:overview}, DVIS++ consists of four components: a segmenter, a noiser, a referring tracker, and a temporal refiner. Among these components, the noiser is non-trainable, whereas the other three are learnable modules. The segmenter extracts object representations from individual frames. During the training process, the noiser adds significant noise to the object representations produced by the segmenter, thus creating more challenging cases. The noiser effectively increases the difficulty of the object association subtask. During inference, the noiser is replaced by the Hungarian algorithm, which pre-matches the object representations of adjacent frames generated by the segmenter. The referring tracker learns to eliminate the noise introduced by the noiser, thereby achieving precise association of object representations in adjacent frames. Once the referring tracker has obtained temporally aligned object representations for all frames, the temporal refiner effectively utilizes temporal features to model spatio-temporal representations of objects. As a result, more accurate segmentation and tracking results are achieved.

The segmenter will be briefly introduced in Section~\ref{Segmenter}, while the referring tracker and temporal refiner will be introduced in Sections~\ref{Referring Tracker} and \ref{Temporal Refiner}, respectively. The noise training strategy and the details of the noiser will be presented in Section~\ref{Denosing Training Strategy}. In Section~\ref{Contrastive Learning}, we will explain of how the contrastive learning strategy is incorporated into the segmenter, referring tracker, and temporal refiner. In Section~\ref{Vision Foundation Models}, we will describe how the visual foundation models are integrated into DVIS++ to allow DVIS++ to be evaluated in various settings, including open-vocabulary and freezing the pre-trained backbone. Finally, the objective functions will be introduced in Section~\ref{Losses}.

\subsection{Segmenter} \label{Segmenter}

Segmenter extracts object representations from the image and can be any query-based image segmentation method. In this paper, DVIS++ uses Mask2Former \cite{cheng2022masked} as the segmenter. The segmenter can obtain the object representations $Q_{Seg} \in \mathbb{R}^{N\times C}$, confidence scores $S \in \mathbb{R}^{N \times |C|}$, and segmentation masks $M \in \mathbb{R}^{N \times H \times W}$ based on an image $\mathbf{I} \in \mathbb{R}^{H \times W \times 3}$. $N$ represents the number of objects, $C$ represents the feature channels, $|C|$ represents the category size, and $H$ and $W$ represent the height and width of the image, respectively.
\begin{equation}
Q_{Seg}, S, M = \mathcal{S}egmenter(I).
\end{equation}
$Q_{Seg}$ will provide rich semantic information about objects for the referring tracker and temporal refiner. It should be noted that here, the object is a generic term for ``\textit{thing}" and ``\textit{stuff}," thus covering semantic, instance, and panoptic segmentation.

\subsection{Referring Tracker} \label{Referring Tracker}

The referring tracker models the inter-frame association as a task of referring denoising. Without the process of inter-frame object association, the \textit{i-th} object representations in $Q_{Seg}$ across all frames may not correspond to the same object. In other words, there is noise in the frame-by-frame association. Taking $Q_{Seg}$ as the initial values, the referring tracker aims to eliminate this noise from the initial values and output the correct object representations, thereby achieving accurate temporal association results.

The overall architecture of the referring tracker $\mathcal{T}$ is depicted as the blue part in Figure~\ref{fig:overview}. It consists of L transformer denoising (TD) blocks connected in series. The referring tracker takes three inputs: $Ref^{T - 1}$, $Q_{Seg}^{T}$, and $Noiser(Q_{Seg}^{T})$. Here, $Ref^{T - 1}$ represents the reference, containing the object information propagated from the last frame. 
According to $Ref^{T - 1}$, the referring tracker aims to output aligned representations of objects in the \textit{T-th} frame. $Q_{Seg}^{T}$ is the object representation of the \textit{T-th} frame outputted by the segmenter. $Noiser(Q_{Seg}^{T})$ represents the addition of noise to $Q_{Seg}^{T}$ (the process will be detailed below), and the referring tracker aims to accurately filter out the noise added by noiser and the inherent noise in $Q_{Seg}^{T}$. The referring tracker outputs both the object representation $Q_{RT}^{T}$ and a new reference $Ref^{T}$ for the next frame. Additionally, $Q_{RT}^{T}$ also serves as the initial value for the temporal refiner.
\begin{equation}
Q_{RT}^{T}, Ref^{T} =  \mathcal{T}(Ref^{T-1},Q_{Seg}^{T},Noiser(Q_{Seg}^{T})) \,.
\end{equation}
The reference $Ref^{1}$ of the first frame is obtained by transforming $Q_{Seg}^{1}$ using MLP:
\begin{equation}
Ref^{1} =  MLP(Q_{Seg}^{1}) \,.
\end{equation}
\begin{figure}[t!]
    \centering
    \includegraphics[width=0.90\linewidth]{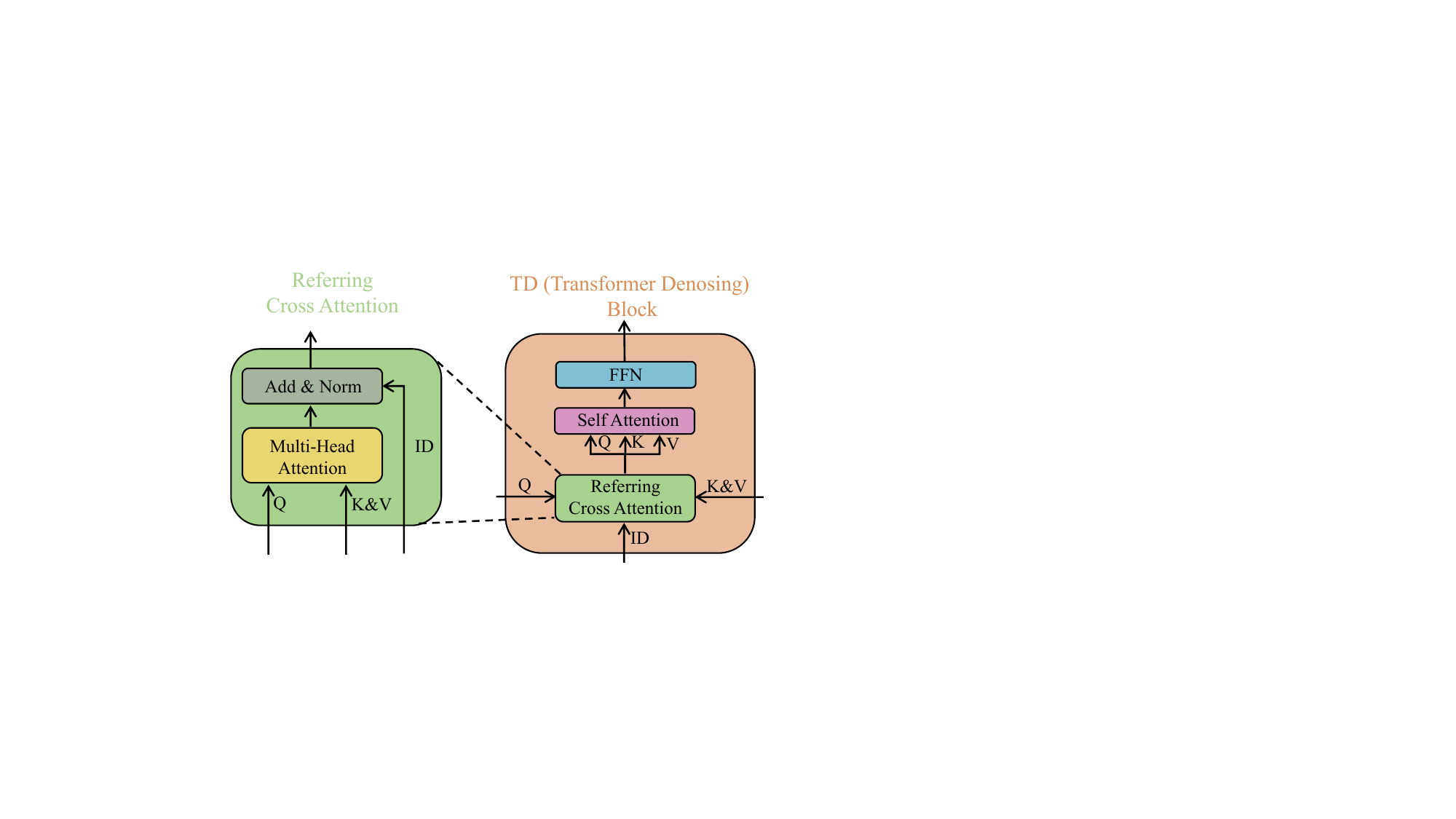}\vspace{-2mm}
    \caption{The architecture of the transformer denoising block, which includes a referring cross-attention, a self-attention, and a Feedforward Neural Network (FFN).}\vspace{-2mm}
    \label{fig:transformer_denosing}
\end{figure}

The core of the referring tracker lies in the transformer denoising (TD) block. The architecture of the TD block is shown in Figure~\ref{fig:transformer_denosing}, which consists of a referring cross-attention (RCA), a self-attention module, and a feedforward neural network (FFN). The RCA plays a crucial role by effectively utilizing the similarity between corresponding objects represented in the previous and the current frame, as illustrated in the green part on the left side of Figure~\ref{fig:transformer_denosing}. The RCA takes three inputs: $Q$, $ID$, and $K\&V$. $ID$ represents the initial values with noise, $Q$ provides reference information, and $K\&V$ provide object features. When $Q$ is set to be equal to $ID$, the $RCA$ degenerates into a standard cross-attention.
\begin{equation}
RCA(ID,Q,K,V)=ID + MHA(Q,K,V) \,.
\end{equation}
MHA denotes the multi-head attention.

\subsection{Temporal Refiner} \label{Temporal Refiner}
\begin{figure}[t!]
    \centering
    \includegraphics[width=0.80\linewidth]{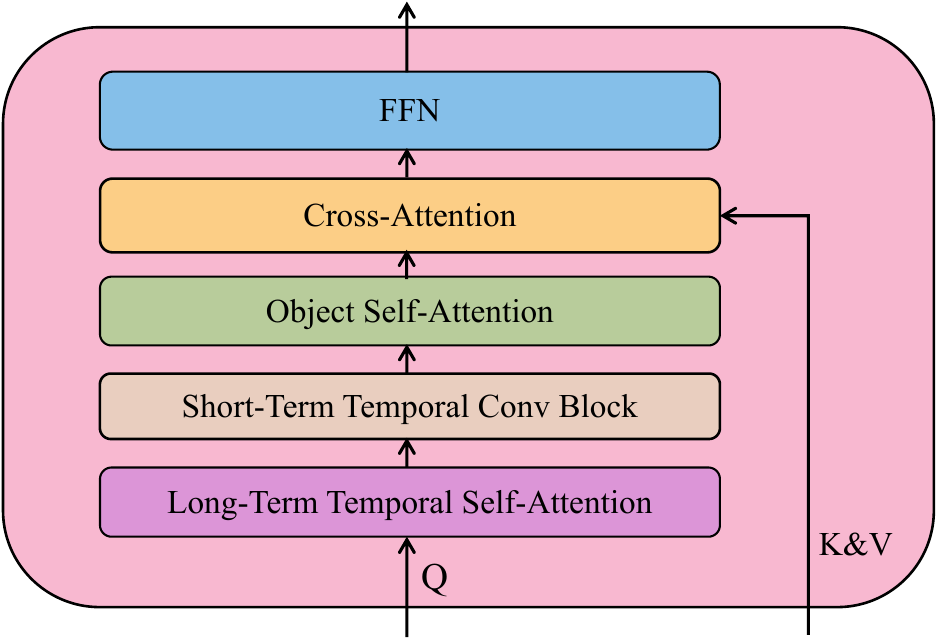}\vspace{-2mm}
    \caption{Architecture of the temporal decoder block.The temporal decoder block utilizes self-attention and 1D convolution to aggregate short-term temporal features and long-term temporal features, respectively.}\vspace{-2mm}
    \label{fig:temporal_decoder}
\end{figure}

The object representation $Q_{RT} \in \mathbb{R}^{N \times T \times C}$, which is aligned frame-by-frame using the referring tracker, provides a solid foundation for modeling the object's spatio-temporal representation. We can apply simple yet effective 1D convolution and self-attention on $Q_{RT}$, a standard 3D tensor, to extract short-term and long-term temporal features. Based on this, we construct a temporal decoder as shown in Figure~\ref{fig:temporal_decoder}. The yellow part in Figure~\ref{fig:overview} represents the temporal refiner $\mathcal{R}$, comprising L temporal decoder blocks cascaded in series. These blocks take $Q_{RT}$ as input and achieve comprehensive temporal feature interaction to obtain the spatio-temporal representation $Q_{TR} \in \mathbb{R}^{N \times T \times C}$ of the object.
\begin{equation}
Q_{TR} =  \mathcal{R}(Q_{RT}, Q_{Seg})\,.
\end{equation}
Since an object should maintain consistent category throughout the entire video, we use temporal weighting to obtain a category representation $\hat{Q}_{TR} \in \mathbb{R}^{N \times C}$ of the object across the video.
\begin{equation}
\hat{Q}_{TR} = \sum^{T}_{t=1} SoftMax(Linear(Q_{TR}^{t}))Q_{TR}^{t} \,.
\end{equation}
Finally, $Q_{TR}$ is used to predict the mask corresponding to the object in each frame, and $\hat{Q}_{TR}$ is used to predict the object's category throughout the entire video.

\begin{figure}[t!]
    \centering
    \includegraphics[width=1.00\linewidth]{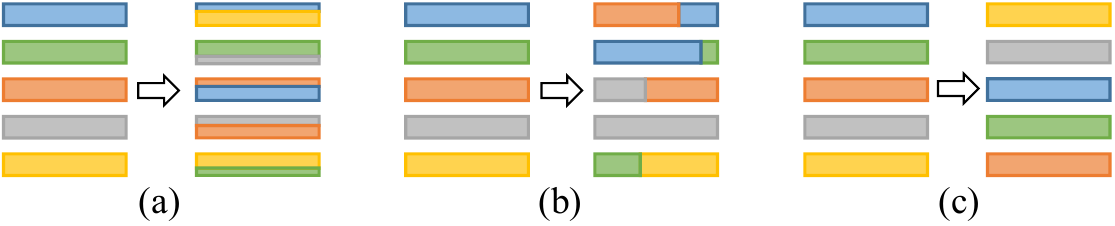} \vspace{-4mm}
    \caption{Noise simulation strategies. From left to right, the strategies are random weighted averaging, random cropping \& concatenating, and random shuffling. Different colors distinguish different instance queries.} \vspace{-2mm}
    \label{fig:noise}
\end{figure}

\begin{figure*}[t!]
    \centering
    \includegraphics[width=0.90\linewidth]{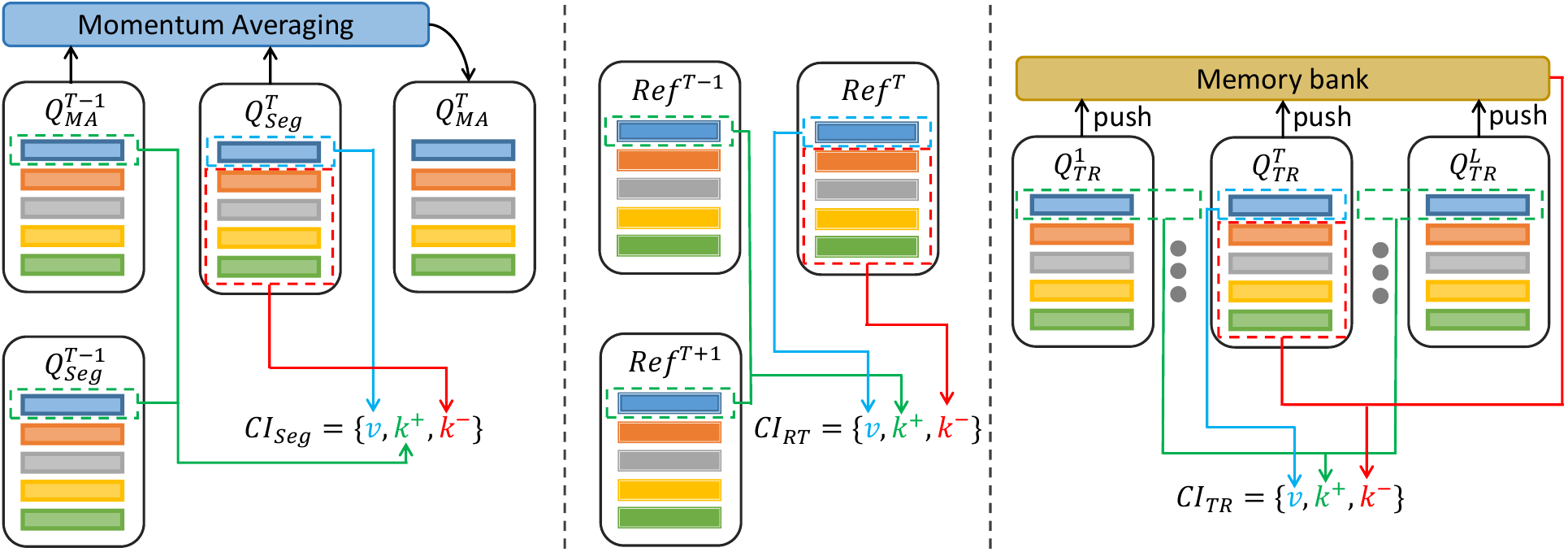}\vspace{-2mm}
    \caption{Contrastive items. $CI_{Seg}$, $CI_{RT}$, and $CI_{TR}$ refer to the contrastive items of the segmenter, referring tracker, and temporal refiner, respectively. The anchor embedding, positive embeddings, and negative embeddings in contrastive items are highlighted in blue, green, and red, respectively. Object representations of the same color indicate the same object identity in the video.}\vspace{-2mm}
    \label{fig:closs}
\end{figure*}

\subsection{Denosing Training Strategy} \label{Denosing Training Strategy}
In DVIS, the object representation $Q_{Seg}$ outputted by the segmenter undergoes a coarse matching process using the Hungarian algorithm \cite{kuhn1955hungarian} before being inputted into the referring tracker. This coarse matching process provides good initial values for the referring tracker, but it may hinder the effectiveness of the referring tracker's training. The coarse matching results from the Hungarian algorithm are usually accurate because most available training video clips are simple. Therefore, the referring tracker learns very little using them as inputs, often leading to an identity mapping shortcut. That is to say, even without changing the inputs, the tracker can achieve a very low loss. To address this issue, we propose a denoising training strategy that enables the referring tracker to primarily learn how to handle challenging cases. Specifically, we have designed various noise simulation strategies to introduce strong noise into the input of the referring tracker. This eliminates the possibility of identity mapping shortcuts, thus compelling the referring tracker to acquire more stable and powerful tracking capabilities. As illustrated in Figure~\ref{fig:overview}, during the training phase, the noiser introduces significant noise to the object representation. However, during the inference phase, noiser is not applied, and instead, the Hungarian algorithm is utilized for coarse matching of the object representation to provide a reliable initial value for the referring tracker.

We propose three simple yet efficient strategies for simulating noise: random weighted averaging, random cropping \& concatenating, and random shuffling. Figure~\ref{fig:noise} illustrates the aforementioned noise simulation strategies, where the input object representations $Q= \left\{ Q^i \in \mathbb{R}^{C}| i \in [1,N] \right\}$ are subjected to the simulated noise, resulting in the output representations $\bar{Q}= \left\{ \bar{Q}^i \in \mathbb{R}^{C}| i \in [1,N] \right\}$.

The random weighted averaging strategy $\mathcal{F}_{w}(.)$ involves randomly combining each object representation with another randomly selected object representation:
\begin{equation}
\left\{
\begin{aligned}
\mathcal{F}_{w}(Q) &= \left\{ f_{w}({Q}^i)| i \in [1,N] \right\} \\
f_{w}({Q}^i) &= \alpha * Q^{i} + (1 - \alpha) * Q^{j} \\
\alpha = rand&(0, 1),\quad j = randint(0, N) \\
\end{aligned} \,.
\right.
\end{equation}
The random cropping \& concatenation strategy $\mathcal{F}_{c}(.)$ is applied to perform both random cropping and concatenating on the object representations:
\begin{equation}
\left\{
\begin{aligned}
\mathcal{F}_{c}(Q) &= \left\{ f_{c}({Q}^i)| i \in [1,N] \right\} \\
f_{c}({Q}^i) &= concatenate(Q^{i}[:k], Q^{j}[k:]) \\
k = ra&ndint(0, C),\quad j = randint(0, N) \\
\end{aligned} \,.
\right.
\end{equation}
The random shuffling strategy $\mathcal{F}_{s}(.)$ involves the random permutation of object representations:
\begin{equation}
\left\{
\begin{aligned}
\mathcal{F}_{s}(Q) &= \left\{ {Q}^{\sigma(i)}| i \in [1,N] \right\} \\
\sigma(i) &= shuffle([1, N])[i] \\
\end{aligned} \,.
\right.
\end{equation}

\subsection{Contrastive Learning} \label{Contrastive Learning}
Contrastive learning aids the network in learning more discriminative object representations by minimizing the distance between anchor embedding $v$ and positive embeddings $k^{+}$ while concurrently enlarging the distance between anchor embedding $v$ and negative embeddings $k^{-}$. The crucial aspect lies in the construction of contrastive items $CI=\left\{ v,k^{+},k^{-} \right\}$ for the segmenter, referring tracker, and temporal refiner. Once the contrastive items are obtained, the contrastive loss $\mathcal{L}_{cl}(.)$ can be computed using the following formula:
\begin{equation}
\begin{aligned}
\mathcal{L}_{cl} &= -log\frac{\sum_{k^{+}} exp(v\cdot k^{+})}{\sum_{k^{+}} exp(v\cdot k^{+})+\sum_{k^{-}}exp(v\cdot k^{-})} \\
 		      &= log[1+\sum_{k^{+}}\sum_{k^{-}}exp(v\cdot k^{-}-v\cdot k^{+})] \,.\\
\end{aligned} 
\end{equation}

\noindent\textbf{Contrastive Items of Segmenter.}
Figure~\ref{fig:closs} illustrates the construction of the contrastive item set $CI_{Seg}$. When selecting an object representation $Q^{T}_{i}$ from the T-th frame as the anchor embedding, other object representations $\left\{Q^{T}_{j} | j\in[1,N],j \ne i\right\}$ from the same frame are chosen as negative embeddings. Additionally, the representation $Q^{T-1}_{i}$ of the same object from the previous frame is selected as the positive embedding. Furthermore, drawing inspiration from CTVIS \cite{ying2023ctvis}, we calculate the momentum average $Q_{MA}^{T}$ of the object representations $\left\{Q^{t} | t \in [1,T] \right\}$ from previous frames using similarity-guided fusion and also utilize  $Q_{MA}^{T}$ as a positive embedding. The process of calculating the momentum average is as follows: 
\begin{equation}
\left\{
\begin{aligned}
&Q_{MA}^{T} = (1-\beta^{T})Q_{MA}^{T-1} + \beta^{T}Q^{T} \\
&\beta^{T} = max \left\{ 0, \frac{1}{T-1}\sum_{t=1}^{T-1} \Psi_{d}(Q^{T},Q^{t}) \right\} \\
\end{aligned} \,,
\right.
\end{equation}
where $\Psi_{d}$ denotes the cosine similarity.

\vspace{2mm}\noindent\textbf{Contrastive Items of Referring Tracker.}
To enhance the between-frame consistency of the references, we construct the contrastive item set $CI_{RT}$, as shown in Figure~\ref{fig:closs}. In this process, when choosing a reference $Ref_{i}^{T}$ from the T-th frame as the anchor embedding, other references $\left\{Ref^{T}_{j} | j\in[1,N],j \ne i\right\}$ from the same frame are selected as negative embeddings. Additionally, positive embeddings are chosen as the reference $Ref^{T-1}_{i}$ from the previous frame and the reference $Ref^{T+1}_{i}$ from the subsequent frame.

\vspace{2mm}\noindent\textbf{Contrastive Items of Temporal Refiner.}
The temporal refiner effectively models the spatio-temporal representation of objects. The introduction of contrastive learning strengthens the discriminative nature of these representations. As shown in Figure~\ref{fig:closs}, the object representation $Q^{T}_{i}$ is chosen as the anchor embedding from the T-th frame. Negative embeddings are selected as other object representations $\left\{Q^{T}_{j} | j\in[1,N],j \ne i\right\}$ from the same frame, while positive embeddings are chosen as the representations $\left\{Q^{t}_{i} | t\in[1,L],t \ne T\right\}$ of the same object from other frames. To address the challenge of identification swapping (i.e., the mismatch of corresponding objects), we maintain a fixed-length memory bank that stores object representations from previous training batches. We then select object representations from the memory bank with the same category as the anchor embedding as additional negative embeddings.

\subsection{Vision Foundation Models} \label{Vision Foundation Models}
In this section, we integrate some vision foundation models, such as DINOv2~\cite{oquab2023dinov2} and CLIP~\cite{radford2021learning}, into DVIS++ to enable evaluation under various settings, such as open-vocabulary and with a frozen pre-trained backbone.

\begin{figure}[t!]
    \centering
    \includegraphics[width=1.0\linewidth]{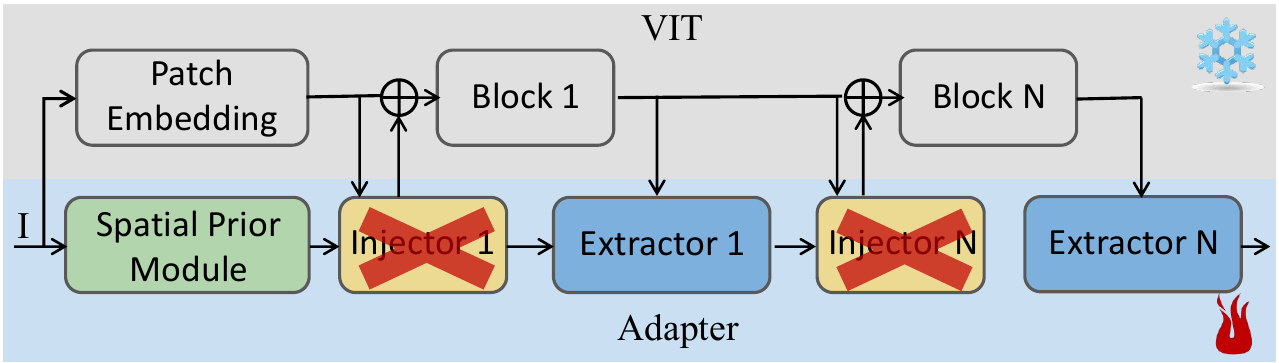}
    \caption{The efficient version of VIT-Adapter. All injectors are eliminated, and the entire VIT is frozen to save a significant amount of GPU memory.}
    \label{fig:adapter}
\end{figure}
\vspace{2mm}\noindent\textbf{DINOv2.} Video segmentation is a dense prediction task that requires the generation of multi-scale features. However, the pre-trained VIT~\cite{dosovitskiy2020image} backbone of DINOv2 cannot output the required multi-scale features. To address this issue, here, the VIT-Adapter \cite{chen2022vision} is incorporated to generate the necessary multi-scale features. It is important to note that video segmentation differs from image segmentation as it involves capturing inter-frame relationships from multiple frames during training. This results in higher GPU memory requirements. Thus, an efficient version of the VIT-Adapter is applied to reduce memory consumption. Figure~\ref{fig:adapter} demonstrates this efficient version, where all injectors are removed, and the DINOv2 VIT backbone is frozen during training. These adaptations enable the DINOv2 pre-trained VIT backbone to be integrated with DVIS++.

\begin{figure}[t!]
    \centering
    \includegraphics[width=1.0\linewidth]{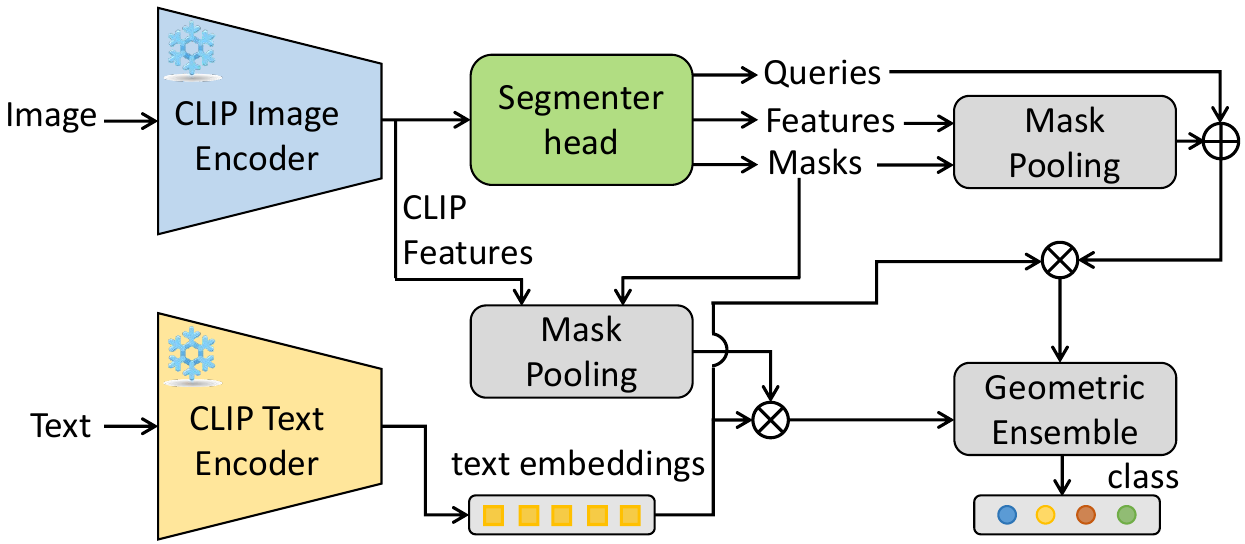}
    \caption{Overview of the open-vocabulary segmenter. Prediction classes are determined by calculating the similarity between object representations and text embeddings. CLIP remains frozen during the training process.}
    \label{fig:open_seg}
\end{figure}
\begin{figure}[t!]
    \centering
    \includegraphics[width=1.0\linewidth]{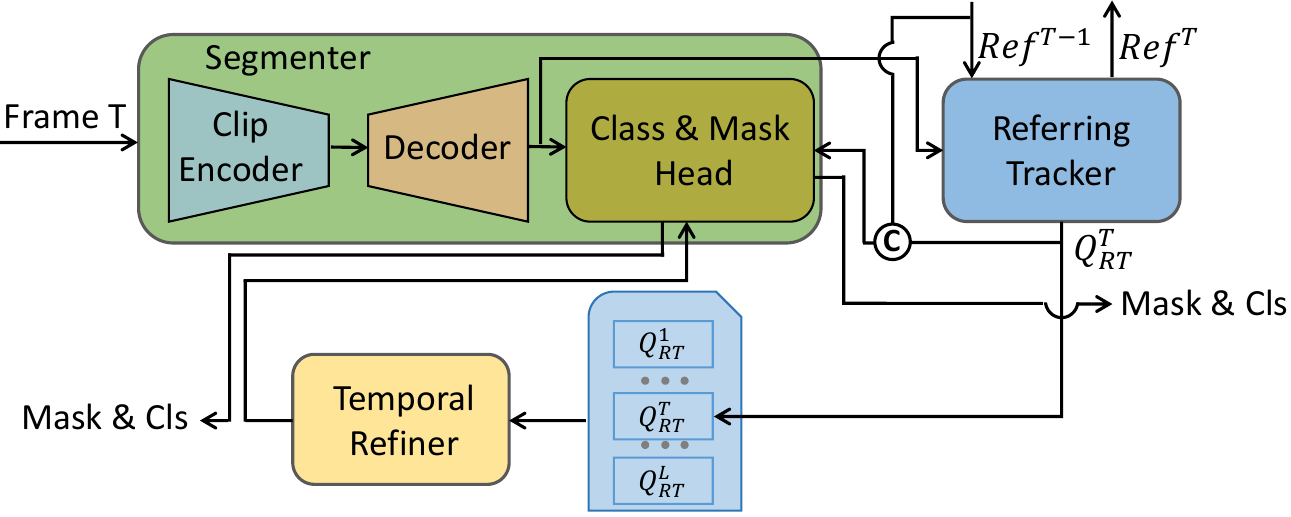}
    \caption{The pipeline of OV-DVIS++. In OV-DVIS++, the referring tracker and temporal refiner share mask and class heads with the segmenter. Additionally, the referring tracker combines $Ref^{T-1}$ and $Q_{RT}^{T}$ to predict object class.}
    \label{fig:open_pipeline}
\end{figure}

\vspace{2mm}\noindent\textbf{CLIP.}
FC-CLIP \cite{yu2023convolutions} demonstrates that training a mask generator from scratch based on a frozen CLIP \cite{radford2021learning} backbone can achieve good open-vocabulary image segmentation capabilities. Inspired by FC-CLIP \cite{yu2023convolutions}, we construct the segmenter to achieve open vocabulary video segmentation. As shown in Figure~\ref{fig:open_seg}, the segmenter is built on the frozen CLIP backbone, where object category prediction is obtained by calculating the similarity between object representations and text embeddings. In addition, both the referring tracker and the temporal refiner are slightly adjusted to better adapt to open-vocabulary segmentation, as shown in Figure ~\ref{fig:open_pipeline}. Firstly, we observe that in some frames, objects may not display any discriminative features, such as a rabbit only showing its back with white fur. This may pose a challenge to object semantics recognition. To address this issue, we combine the reference and the query to predict the category in the referring tracker. Secondly, we no longer retrain the mask head and class head for the referring tracker and temporal refiner. Instead, we use the frozen, pre-trained mask head and class head from the segmenter, meaning that the referring tracker and temporal refiner only implement object representation mapping. Considering the improvements above, we have developed OV-DVIS++, which facilitates open-vocabulary universal video segmentation.

\subsection{Objective Functions} \label{Losses}
Due to the requirement of modeling the relationship between multiple frames (usually more than 5 frames) for video segmentation, end-to-end training would require prohibitively high GPU memory requirements. Therefore, we adopt a separate training approach for the segmenter, referring tracker, and temporal refiner to alleviate the GPU memory demands. Specifically, we initially train the segmenter for the image segmentation task, followed by training the referring tracker with the frozen segmenter. Finally, we train the temporal refiner while keeping the segmenter and referring tracker frozen.

\vspace{2mm}\noindent\textbf{Objective Functions of Segmenter.}
The Segmenter is trained using the same objective functions as Mask2Former \cite{cheng2022masked}. Firstly, the predicted results $\hat{Y}_{Seg}=\left\{ \hat{y}_{i} | i \in [1, N]\right\}$ are matched one-to-one with the ground truth $Y=\left\{y_{i} | i \in [1, M]\right\}$ using the Hungarian algorithm \cite{kuhn1955hungarian}:
\begin{equation}
  \hat{\sigma}=\arg \underset{\sigma}{\min} \ \ \sum^{N}_{i=1} \ \mathcal{L}_{match}(y_{i},\hat{y}_{\sigma(i)})  \,,
\end{equation}
where $\mathcal{L}_{match}$ is the matching cost used in \cite{cheng2022masked}.
The predicted results that do not match with any ground truth will be assigned to the background. 

The objective function $\mathcal{L}_{Seg}$ is comprised of a contrastive item $\mathcal{L}_{cl}$, a mask item $\mathcal{L}_{mask}$, and a classification item $\mathcal{L}_{cls}$. The mask item $\mathcal{L}_{mask}$ encompasses both the dice loss \cite{milletari2016v} $\mathcal{L}_{dice}$ and the cross-entropy loss $\mathcal{L}_{ce}$:
\begin{equation}
\left\{
\begin{aligned}
 \mathcal{L}_{Seg} &= \lambda_{cl}\mathcal{L}_{cl} + \mathcal{L}_{mask} + \lambda_{cls}\mathcal{L}_{cls} \\
\mathcal{L}_{mask} &=  \lambda_{ce}\mathcal{L}_{ce} + \lambda_{dice}\mathcal{L}_{dice}\\
\end{aligned} \,,
\right.
\end{equation}
where $\lambda_{cl}$, $\lambda_{cls}$, $\lambda_{dice}$, and $\lambda_{ce}$ are set as 2.0, 2.0, 5.0, and 5.0, respectively.

\vspace{2mm}\noindent\textbf{Objective Functions of Referring Tracker.}
The referring tracker tracks objects frame by frame, and as such, the network is supervised using an objective function that aligns with this paradigm. Specifically, the prediction results $\hat{Y}_{RT}=\left\{ \hat{y}_{i}^{t} | i \in [1, N], t \in [1,L]\right\}$ are only matched with the ground truth $Y=\left\{y_{i}^{\bar{t}} | i \in [1, M], \bar{t} \in [1,L]\right\}$ on the frame where the object first appears. To expedite convergence during the early training phase, the prediction results of the frozen segmenter $\hat{Y}_{Seg}$ are used for matching instead of the referring tracker's prediction results.
\begin{equation}
\left\{
\begin{aligned}
& \hat{\sigma}=\arg \underset{\sigma}{\min} \ \ \sum^{N}_{i=1} \ \mathcal{L}_{match}(y_{i}^{f(i)},\hat{y}_{\sigma(i)}^{f(i)}) \\
& \hat{y} = \hat{y}_{RT} \ \ if \ \ Iter \geq \frac{Max\_Iter}{2} \ \ else \ \ \hat{y}_{Seg}
\end{aligned}
\right.
\end{equation}
where $f(i)$ represents the frame in which the $i$-th object first appears. The objective function $\mathcal{L}_{RT}$ is exactly the same as $\mathcal{L}_{Seg}$:
\begin{equation}
\mathcal{L}_{RT} = \sum^{T}_{t=1} \ \sum^{N}_{i=1} \mathcal{L}_{Seg}(y_{i}^{t}, {\hat{y}}_{\hat{\sigma}(i)}^{t}) \,.
\end{equation}

\vspace{2mm}\noindent\textbf{Objective Functions of Temporal Refiner.}
The temporal refiner provides mask predictions $M \in \mathbb{R}^{N \times T \times H \times W}$ for objects across all frames and category predictions $C \in \mathbb{R}^{N \times |C|+1}$ for the entire video, enabling holistic matching of the predicted results $\hat{Y}_{TR}=(M,C)$ for the entire video. To simplify the computation, we reshape the masks $M\in \mathbb{R}^{N \times T \times H \times W}$ as $\bar{M}\in \mathbb{R}^{N \times TH \times W}$. By doing this, the entire video can be treated as a large image $\mathcal{I} \in \mathbb{R}^{TH \times W \times 3}$, allowing us to use the same matching process and objective function as the segmenter.

\section{Implementation Details}
\textbf{Settings.} DVIS++ employs Mask2Former~\cite{cheng2022masked} as the segmenter. In DVIS++, the referring tracker utilizes six transformer denoising blocks, and the temporal refiner also employs six temporal decoder blocks. The channel dimensions of the object representations in both the tracker and refiner are consistent with those in the segmenter. The noiser applies a random weighted averaging strategy to introduce noise into the object representations output by the segmenter with probabilities of 0.5 and 0.8 when training for 40K and 160K iterations, respectively. These settings are uniform across all datasets.

OV-DVIS++ employs FC-CLIP~\cite{yu2023convolutions} as the segmenter. Besides the classification branch, the configurations of OV-DVIS++'s referring tracker and temporal refiner are identical to those of DVIS++. The classification process in OV-DVIS++ mirrors that of FC-CLIP, with the classification result derived from the concatenation of the query and reference. Additionally, FC-CLIP does not account for multi-dataset joint training. To facilitate this, we introduce a unique void embedding for each dataset, enabling joint training across multiple datasets.

\vspace{2mm}\noindent\textbf{Training \& Inference.} We employ the AdamW optimizer \cite{loshchilov2017decoupled} with an initial learning rate of 1e-4 and a weight decay of 5e-2 for training. The segmenter is initialized with weights pre-trained on the COCO dataset. In DVIS++, the segmenter undergoes fine-tuning on the video dataset before being frozen. Conversely, in OV-DVIS++, the segmenter is frozen directly without any fine-tuning. Notably, OV-DVIS++ is trained exclusively on the COCO dataset~\cite{lin2014microsoft} and performs zero-shot inference on video datasets without utilizing any video dataset-specific training.

For VIS datasets, such as YouTube-VIS 2019, 2021, and 2022, as well as OVIS, we utilize COCO pseudo videos \cite{wu2022seqformer} for joint training with DVIS++. For VSS and VPS datasets, no additional datasets are employed. We implement random resize and random crop augmentations. During training, the videos in the VIS datasets are resized to a range of 320p to 640p, and to 480p for inference. For the VPS datasets, the videos are resized to a range of 480p to 800p during training and to 720p for inference. Unless otherwise specified, the referring tracker and temporal refiner are trained for 20K iterations, with the learning rate reduced to one-tenth at 14K iterations. If COCO pseudo videos are utilized for joint training, the number of iterations is increased to 40K, with the learning rate decay occurring at 28K iterations. The input for training the referring tracker comprises 5 consecutive frames sampled from the video while training the temporal refiner involves 21 consecutive frames.

\section{Experiments}
\begin{table*}[t!]
\centering
 \caption{Results on the validation set of YouTube-VIS 2019 \& 2021 and OVIS. The best metrics in each group are bolded and highlighted in red, while the second-best metrics are underlined and highlighted in blue. $\ddag$ denotes that these methods are semi-offline rather than pure offline, and significant performance degradation occurs when inferring in pure offline mode.}
 \label{tab:ytvis}
\setlength{\tabcolsep}{3.0pt}
\begin{tabular}{l l | c c c c c| c c c c c | c c c c c}
\toprule[1.5pt]
	\multirow{2}{*}{Method} & \multirow{2}{*}{Backbone} &  \multicolumn{5}{c|}{Youtube-VIS 2019} &  \multicolumn{5}{c|}{Youtube-VIS 2021} & \multicolumn{5}{c}{OVIS} \\
	~ & ~ & AP &  AP$_{\rm 50}$ & AP$_{\rm 75}$ &  AR$_{\rm 1}$ & AR$_{\rm 10}$ & AP &  AP$_{\rm 50}$ & AP$_{\rm 75}$ &  AR$_{\rm 1}$ & AR$_{\rm 10}$ & AP &  AP$_{\rm 50}$ & AP$_{\rm 75}$ &  AR$_{\rm 1}$ & AR$_{\rm 10}$ \\
	\midrule[1pt]
	\multicolumn{17}{c}{\textit{Online}} \\
	\midrule[1pt]
	MaskTrack R-CNN~\pub{CVPR19}~\cite{yang2019video} & ResNet-50 & 30.3 & 51.1 & 32.6 & 31.0 & 35.5 & 28.6 & 48.9 & 29.6 & 26.5 & 33.8 & 10.8 & 25.3 & 8.5 & 7.9 & 14.9 \\
	SipMask~\pub{ECCV20}~\cite{cao2020sipmask} & ResNet-50 & 33.7 & 54.1 & 35.8 & 35.4 & 40.1 & 31.7 & 52.5 & 34.0 & 30.8 & 37.8 & 10.2 & 24.7 & 7.8 & 7.9 & 15.8 \\
	CrossVIS~\pub{ICCV21}~\cite{yang2021crossover}& ResNet-50 & 36.3 & 56.8 & 38.9 & 35.6 & 40.7 & 34.2 & 54.4 & 37.9  & 30.4 & 38.2 & 14.9 & 32.7 & 12.1 & 10.3 & 19.8 \\
	VISOLO~\pub{CVPR22}~\cite{han2022visolo}& ResNet-50 & 38.6 & 56.3 & 43.7 & 35.7 & 42.5 & 36.9 & 54.7 & 40.2 & 30.6 & 40.9 & 15.3 & 31.0 & 13.8 & 11.1 & 21.7  \\
	MinVIS~\pub{NIPS22}~\cite{huang2022minvis}& ResNet-50 & 47.4 & 69.0 & 52.1 & 45.7 & 55.7 & 44.2 & 66.0 & 48.1 & 39.2 & 51.7 & 25.0 & 45.5 & 24.0 & 13.9 & 29.7  \\
	IDOL~\pub{ECCV22}~\cite{wu2022defense}& ResNet-50 & 49.5 & \second{74.0} & 52.9 & 47.7 & 58.7 & 43.9 & 68.0 & 49.6 & 38.0 & 50.9 & 28.2 & 51.0 & 28.0 & 14.5 & 38.6 \\
	GenVIS~\pub{CVPR23}~\cite{heo2023generalized} & ResNet-50 & 50.0 & 71.5 & 54.6 & 49.5 & 59.7 & 47.1 & 67.5 & 51.5 & 41.6 & 54.7 & 35.8 & \second{60.8} & 36.2 & \second{16.3} & 39.6 \\
	GRAtt-VIS~\pub{ArXiv23}~\cite{hannan2023gratt} & ResNet-50 & 50.4 & 70.7 & 55.2 & 48.4 & 58.7 & 48.9 & 69.2 & 53.1 & \second{41.8} & \second{56.0} & \second{36.2} & \second{60.8} & \second{36.8} & \highest{16.8} & \second{40.0} \\ 
	TCOVIS~\pub{ICCV23}~\cite{li2023tcovis} & ResNet-50 & \second{52.3} & 73.5 & 57.6 & \second{49.8} & \second{60.2} & \second{49.5} & \second{71.2} & \second{53.8} & 41.3 & 55.9 & 35.3 & 60.7 & 36.6 & 15.7 & 39.5 \\
	UNINEXT~\pub{CVPR23}~\cite{yan2023universal} & ResNet-50 & 53.0 & 75.2 & \second{59.1} & - & - & - & - & - & - & -& 34.0 & 55.5 & 35.6 & - & - \\
        \rowcolor{gray!35}
	DVIS~\pub{ICCV23}~\cite{zhang2023dvis} & ResNet-50 & 51.2 & 73.8 & 57.1 & 47.2 & 59.3 & 46.4 & 68.4 & 49.6 & 39.7 & 53.5 & 30.2 & 55.0 & 30.5 & 14.5 & 37.3 \\
        \rowcolor{gray!35}
	DVIS++ & ResNet-50 & \highest{55.5} & \highest{80.2} & \highest{60.1} & \highest{51.1} & \highest{62.6} & \highest{50.0} & \highest{72.2} & \highest{54.5} & \highest{42.8} & \highest{58.4} & \highest{37.2} & \highest{62.8} & \highest{37.3} & 15.8 & \highest{42.9} \\
	\hline
	MinVIS~\pub{NIPS22}~\cite{huang2022minvis}& Swin-L & 61.6 & 83.3 & 68.6 & 54.8 & 66.6 & 55.3 & 76.6 & 62.0 & 45.9 & 60.8 & 39.4 & 61.5 & 41.3 & 18.1 & 43.3 \\
	IDOL~\pub{ECCV22}~\cite{wu2022defense}& Swin-L & 64.3 & 87.5 & 71.0 & 55.6 & 69.1 & 56.1 & 80.8 & 63.5 & 45.0 & 60.1 & 40.0 & 63.1 & 40.5 & 17.6 & 46.4 \\
	GenVIS~\pub{CVPR23}~\cite{heo2023generalized} & Swin-L & 64.0 & 84.9 & 68.3 & 56.1 & 69.4 & 59.6 & 80.9 & 65.8 & 48.7 & 65.0 & 45.2 & 69.1 & 48.4 & 19.1 & 48.6  \\
	GRAtt-VIS~\pub{ArXiv23}~\cite{hannan2023gratt} & Swin-L & 63.1 & 85.6 & 67.2 & 55.5 & 67.8 & 60.3 & 81.3 & 67.1 & \second{48.8} & 64.5 & 45.7 & 69.1 & 47.8 & \second{19.2} & 49.4  \\ 
	RefineVIS~\pub{ArXiv23}~\cite{abrantes2023refinevis} & Swin-L & 64.3 & 87.6 & 70.9 & 55.8 & 68.2 & \second{61.4} & \highest{84.1} & 68.5 & 48.3 & 65.2 & 46.1 & 69.7 & 47.8 & 19.0 & 50.8 \\
	CTVIS~\pub{ICCV23}~\cite{ying2023ctvis} & Swin-L & 65.6 & \second{87.7} & 72.2 & \second{56.5} & \second{70.4} & 61.2 & \second{84.0} & \second{68.8} & 48.0 & \second{65.8} & 46.9 & 71.5 & 47.5 & 19.1 & \second{52.1} \\
	TCOVIS~\pub{ICCV23}~\cite{li2023tcovis} & Swin-L & 64.1 & 86.6 & 69.5 & 55.8 & 69.0 & 61.3 & 82.9 & 68.0 & 48.6 & 65.1 & 46.7 & 70.9 & 49.5 & 19.1 & 50.8 \\
	UNINEXT~\pub{CVPR23}~\cite{yan2023universal} & ConvNeXt-L & 64.3 & 87.2 & 71.7 & - & - & - & - & - & - & - & 41.1 & 65.8 & 42.0 & - & - \\
	UNINEXT~\pub{CVPR23}~\cite{yan2023universal} & VIT-H & \second{66.9} & 87.5 & \second{75.1} & - & - & - & - & - & - & - & \second{49.0} & \highest{72.5} & \second{52.2} & - & - \\
        \rowcolor{gray!35}
	DVIS~\pub{ICCV23}~\cite{zhang2023dvis} & Swin-L & 63.9 & 87.2 & 70.4 & 56.2 & 69.0 & 58.7 & 80.4 & 66.6 & 47.5 & 64.6 & 45.9 & 71.1 & 48.3 & 18.5 & 51.5 \\ 
        \rowcolor{gray!35}
	DVIS++ & VIT-L & \highest{67.7} & \highest{88.8} & \highest{75.3} & \highest{57.9} & \highest{73.7} & \highest{62.3} & 82.7 & \highest{70.2} & \highest{49.5} & \highest{68.0} & \highest{49.6} & \highest{72.5} & \highest{55.0} & \highest{20.8} & \highest{54.6} \\
	\bottomrule[1.5pt]
	\multicolumn{17}{c}{\textit{Offline}} \\
	\midrule[1pt]
	EfficientVIS~\pub{CVPR22}~\cite{wu2022efficient} & ResNet-50 & 37.9 & 59.7 & 43.0 & 40.3 & 46.6 & 34.0 & 57.5 & 37.3 & 33.8 & 42.5 & - & - & - & - & - \\
	IFC~\pub{NIPS21}~\cite{hwang2021video}& ResNet-50 & 41.2 & 65.1 & 44.6 & 42.3 & 49.6 & 35.2 & 55.9 & 37.7 & 32.6 & 42.9 & 13.1 & 27.8 & 11.6 & 9.4 & 23.9 \\
	Mask2Former-VIS~\pub{ArXiv22}~\cite{cheng2021mask2former}& ResNet-50 & 46.4 & 68.0 & 50.0 & - & - & 40.6 & 60.9 & 41.8 & - & - & 17.3 & 37.3 & 15.1 & 10.5 & 23.5 \\
	SeqFormer~\pub{ECCV22}~\cite{wu2022seqformer}& ResNet-50 & 47.4 & 69.8 & 51.8 & 45.5 & 54.8 & 40.5 & 62.4 & 43.7 & 36.1 & 48.1 & 15.1 & 31.9 & 13.8 & 10.4 & 27.1 \\
	VITA~\pub{NIPS22}~\cite{heo2022vita}& ResNet-50 & 49.8 & 72.6 & 54.5 & 49.4 & \second{61.0} & 45.7 & 67.4 & 49.5 & 40.9 & 53.6 & 19.6 & 41.2 & 17.4 & 11.7 & 26.0\\
	GenVIS$^{\ddag}$~\pub{CVPR23}~\cite{heo2023generalized}& ResNet-50 & 51.3 & 72.0 & 57.8 & 49.5 & 60.0 & 46.3 & 67.0 & 50.2 & 40.6 & 53.2 & \second{34.5} & 59.4 & \second{35.0} & \second{16.6} & 38.3 \\ 
	MDQE$^{\ddag}$~\pub{CVPR23}~\cite{li2023mdqe} & ResNet-50 & 47.8 & - & - & - & - & 44.5 & 67.1 & 48.7 & 37.9 & 49.8 & 29.2 & 55.2 & 27.1 & 14.5 & 34.2 \\
	NOVIS$^{\ddag}$~\pub{ICML23}~\cite{meinhardt2023novis} & ResNet-50 & \second{52.8} & 75.7 & 56.9 & \second{50.3} & 60.6 & 47.2 & 69.4 & 50.0 & \second{41.3} & 54.4 & 30.8 & 54.4 & 31.0 & 15.2 & 35.3\\
	RefineVIS$^{\ddag}$~\pub{ArXiv23}~\cite{abrantes2023refinevis} & ResNet-50 & 52.2 & 76.3 & 57.7 & 47.5 & 57.6 & \second{50.2} & \second{72.8} & \second{55.4} & 41.2 & \second{56.3} & 33.7 & 56.2 & 34.8 & 15.6 & \second{39.8} \\
        \rowcolor{gray!35}
	DVIS~\pub{ICCV23}~\cite{zhang2023dvis} & ResNet-50 & 52.6 & \second{76.5} & \second{58.2} & 47.4 & 60.4 & 47.4 & 71.0 & 51.6 & 39.9 & 55.2 & 33.8 & \second{60.4} & 33.5 & 15.3 & 39.5 \\
        \rowcolor{gray!35}
	DVIS++ & ResNet-50 & \highest{56.7} & \highest{81.4} & \highest{62.0} & \highest{51.8} & \highest{64.7} & \highest{52.0} & \highest{75.4} & \highest{57.8} & \highest{42.2} & \highest{59.6} & \highest{41.2} & \highest{68.9} & \highest{40.9} & \highest{16.8} & \highest{47.3} \\
	\hline
	SeqFormer~\pub{ECCV22}~\cite{wu2022seqformer}& Swin-L & 59.3 & 82.1 & 66.4 & 51.7 & 64.4 & 51.8 & 74.6 & 58.2 & 42.8 & 58.1 & - & - & - & - & - \\
	Mask2Former-VIS~\pub{ArXiv22}~\cite{cheng2021mask2former}& Swin-L & 60.4 & 84.4 & 67.0 & - & - & 52.6 & 76.4 & 57.2 & - & - & 25.8 & 46.5 & 24.4 & 13.7 & 32.2 \\
	VITA~\pub{NIPS22}~\cite{heo2022vita}& Swin-L & 63.0 & 86.9 & 67.9 & 56.3 & 68.1 & 57.5 & 80.6 & 61.0 & 47.7 & 62.6 & 27.7 & 51.9 & 24.9 & 14.9 & 33.0 \\
	GenVIS$^{\ddag}$~\pub{CVPR23}~\cite{heo2023generalized}& Swin-L & 63.8 & 85.7 & 68.5 & 56.3 & 68.4 & 60.1 & 80.9 & 66.5 & \highest{49.1} & 64.7 & 45.4 & 69.2 & 47.8 & 18.9 & 49.0 \\
	MDQE$^{\ddag}$~\pub{CVPR23}~\cite{li2023mdqe} & Swin-L & 59.9 & - & - & - & - & 56.2 & 80.0 & 61.1 & 44.9 & 59.1 & 41.0 & 67.9 & 42.7 & 18.3 & 45.2 \\
	NOVIS$^{\ddag}$~\pub{ICML23}~\cite{meinhardt2023novis} & Swin-L & \second{65.7} & 87.8 & 72.2 & 56.3 & \second{70.3} & 59.8 & 82.0 & 66.5 & 47.9 & 64.6 & 43.0 & 66.9 & 44.5 & 18.9 & 46.3 \\
	Tube-Link$^{\ddag}$~\pub{ICCV23}~\cite{li2023tube} &  Swin-L & 64.6 & 86.6 & 71.3 & 55.9 & 69.1 & 58.4 & 79.4 & 64.3 & 47.5 & 63.6 & - & - & - & - & - \\
	RefineVIS$^{\ddag}$~\pub{ArXiv23}~\cite{abrantes2023refinevis} & Swin-L & 64.4 & \second{88.3} & 72.2 & 55.8 & 68.4 & \second{61.2} & \second{83.7} & \second{69.2} & 47.9 & 64.8 & 46.0 & 70.4 & 48.4 & \second{19.1} & 51.2 \\
        \rowcolor{gray!35}
	DVIS~\pub{ICCV23}~\cite{zhang2023dvis} & Swin-L & 64.9 & 88.0 & \second{72.7} & \second{56.5} & \second{70.3} & 60.1 & 83.0 & 68.4 & 47.7 & \second{65.7} & \second{48.6} & \second{74.7} & \second{50.5} & 18.8 & \second{53.8} \\
        \rowcolor{gray!35}
	DVIS++ & VIT-L & \highest{68.3} & \highest{90.3} & \highest{76.1} & \highest{57.8} & \highest{73.4} & \highest{63.9} & \highest{86.7} & \highest{71.5} & \second{48.8} & \highest{69.5} & \highest{53.4} & \highest{78.9} & \highest{58.5} & \highest{21.1} & \highest{58.7} \\
\bottomrule[1.5pt]
 \end{tabular}
\end{table*}
\subsection{Datasets and Metrics}
To validate the effectiveness of the proposed methods, we conducted extensive experiments on six benchmarks, as described below.

\noindent\textbf{Youtube-VIS 2019.} YouTube-VIS 2019 \cite{yang2019video} is a large-scale dataset for VIS that comprises 2,238/302/343 videos for training/validation/testing, respectively. The videos in YouTube-VIS 2019 are relatively short, and the object motion is relatively simple. This dataset encompasses 40 categories, out of which 22 categories overlap with the COCO dataset \cite{lin2014microsoft}, while the remaining 18 categories are not included in the COCO dataset. The performance of VIS methods is evaluated using the AP (Average Precision) metric. The calculation process is similar to that in image segmentation, but with one key difference: it utilizes the IOU (Intersection over Union) of video segmentation results instead of image segmentation results. Further details can be found in \cite{yang2019video}.

\noindent\textbf{Youtube-VIS 2021.} YouTube-VIS 2021 \cite{yang2019video} was created by expanding the videos and refining the annotations from YouTube-VIS 2019. It consists of 2,985/421/453 videos for training/validation/testing, respectively. Additionally, it includes 40 object categories, which differ from the 2019 version. Out of these categories, 24 overlap with the COCO \cite{lin2014microsoft} dataset, while the remaining 16 categories are not included in the COCO dataset.

\noindent\textbf{Youtube-VIS 2022.} The YouTube-VIS 2022 \cite{yang2019video} dataset utilizes the same training set as YouTube-VIS 2021 but incorporates extra-long videos in the validation and test sets. Additionally, YouTube-VIS 2022 employs different evaluation strategies, separately measuring AP on the long videos (AP$_l$) and the short videos (AP$_s$). The final performance is evaluated by taking the average of AP$_l$ and AP$_s$.

\noindent\textbf{OVIS.} OVIS \cite{qi2022occluded} is a new and highly challenging VIS dataset comprising 25 object categories. Among these categories, 16 categories overlap with the COCO \cite{lin2014microsoft} dataset, while the remaining 9 categories are unique to OVIS. The dataset includes 607/140/154 videos for training/validation/testing, which are longer and contain more instance annotations compared to the YouTube-VIS series. OVIS also features a significant number of videos with objects exhibiting severe occlusion, complex motion trajectories, and rapid deformation, thus making it more representative of real-world scenarios. Therefore, OVIS serves as an ideal benchmark for evaluating the performance of various VIS methods. Additionally, OVIS computes AP for objects with light, medium, and heavy occlusion, denoted as AP$_l$, AP$_m$, and AP$_h$, respectively.

\noindent\textbf{VIPSeg.} VIPSeg \cite{miao2022large} is a large-scale dataset for panoptic segmentation in the wild, which showcases a diverse array of real-world scenarios and encompasses 124 categories. These categories consist of 58 ``thing" classes and 66 ``stuff" classes. The dataset comprises 3,536 videos and 84,750 frames, with 2,806 videos allocated for training, 343 videos for validation, and 387 videos for testing. VPS methods are evaluated using the VPQ (Video Panoptic Quality) metric~\cite{miao2022large}.



\noindent\textbf{VSPW.} VSPW \cite{miao2021vspw} is a large-scale video semantic segmentation (VSS) dataset that shares the same videos and categories as VIPSeg. The performance of VSS methods is evaluated using mIoU and VC (Video Consistency)~\cite{miao2021vspw}.

\subsection{Comparison with the State-of-the-art Methods}
\subsubsection{Performance on Video Instance Segmentation}
\textbf{YouTube-VIS 2019 \& 2021.} YouTube-VIS 2019 and YouTube-VIS 2021 comprise videos with shorter durations and simpler scenes. Firstly, we compare our approach DVIS++ with SOTA VIS methods on these two datasets, and the results are presented in Table~\ref{tab:ytvis}. When using ResNet-50 \cite{he2016deep} as the backbone, DVIS++ achieves an AP of 55.5 and 50.0 in the online mode (without temporal refiner) on YouTube-VIS 2019 and YouTube-VIS 2021, respectively. This surpasses all contemporary online VIS methods. Additionally, by incorporating temporal information, DVIS++ achieves an AP of 56.7 and 52.0 in the offline mode (with temporal refiner) on YouTube-VIS 2019 and YouTube-VIS 2021, respectively. These results significantly outperform the previous SOTA methods (NOVIS \cite{meinhardt2023novis} and RefineVIS \cite{abrantes2023refinevis}) on these two datasets, with improvements of 3.9 AP (56.7 \textit{vs.} 52.8) and 1.8 AP (52.0 \textit{vs.} 50.2), respectively.

When using a frozen pre-trained VIT-L~\cite{dosovitskiy2020image}, DVIS++ achieves an AP of 67.7 and 62.3 on YouTube-VIS 2019 and YouTube-VIS 2021, respectively, in online mode, surpassing all previous SOTA methods, both online and offline. When inferring in offline mode, DVIS++ demonstrates even stronger performance by effectively utilizing spatio-temporal features, achieving an AP of 68.3 and 63.9 on YouTube-VIS 2019 and YouTube-VIS 2021, respectively. DVIS++ outperforms the previous SOTA methods UNINEXT~\cite{yan2023universal} and TCOVIS~\cite{li2023tcovis} with improvements of 1.4 AP (68.3 compared to 66.9) and 2.7 AP (63.9 compared to 61.2), respectively, despite UNINEXT using a larger VIT-H~\cite{dosovitskiy2020image} backbone.

In addition, recent offline methods, such as GenVIS~\cite{heo2023generalized}, MDQE~\cite{li2023mdqe}, NOVIS~\cite{meinhardt2023novis}, and RefineVIS~\cite{abrantes2023refinevis}, have only achieved comparable accuracy to online methods in semi-offline mode (processing the video clip by clip). They exhibit significant performance degradation in pure offline mode (processing the entire video at once). For instance, \cite{heo2023generalized} demonstrates GenVIS experiences approximately a 3 AP performance degradation when inferring in pure offline mode on YouTube-VIS 2021 compared to semi-offline mode. In contrast, our proposed DVIS and DVIS++ demonstrate a significant performance improvement when operating in pure offline mode, surpassing both semi-offline mode and online mode. This indicates that DVIS and DVIS++ effectively leverage spatio-temporal information.

\begin{table}[t!]
\centering
\caption{Results on the validation set of YouTube-VIS 2022. AP$^{\rm L}$ refers to the AP on the long video set.}
\label{tab:ytvis22}
\setlength{\tabcolsep}{4.0pt}
\begin{tabular}{l l | c c c c c}
\toprule[1.5pt]
	\multirow{2}{*}{Method} & \multirow{2}{*}{Backbone} &  \multicolumn{5}{c}{YouTube-VIS 2022}\\
	~ & ~ & AP$^{\rm L}$ &  AP$_{\rm 50}^{\rm L}$ & AP$_{\rm 75}^{\rm L}$ &  AR$_{\rm 1}^{\rm L}$ & AR$_{\rm 10}^{\rm L}$ \\
	\midrule[1pt]
	\multicolumn{7}{c}{\textit{Online}} \\
	\midrule[1pt]
	MinVIS~\pub{NIPS22}~\cite{huang2022minvis} & ResNet-50 & 23.3 & 47.9 & 19.3 & 20.2 & 28.0 \\
        \rowcolor{gray!35}
	DVIS~\pub{ICCV23}~\cite{zhang2023dvis} & ResNet-50 & \second{31.6} & \second{52.5} & \second{37.0} & \second{30.1} & \second{36.3} \\
        \rowcolor{gray!35}
	DVIS++ & ResNet-50 & \highest{37.2} & \highest{57.4} & \highest{40.7} & \highest{31.8} & \highest{44.6} \\
	\hline
	MinVIS~\pub{NIPS22}~\cite{huang2022minvis} & Swin-L & 33.1 & \second{54.8} & 33.7 & 29.5 & 36.6 \\	
        \rowcolor{gray!35}
	DVIS~\pub{ICCV23}~\cite{zhang2023dvis} & Swin-L & \highest{39.9} & \highest{58.2} & \highest{42.6} & \highest{33.5} & \highest{44.9} \\ 
        \rowcolor{gray!35}
	DVIS++ & VIT-L & \second{37.5} & 53.7 & \second{39.4} & \second{32.4} & \second{43.5} \\ 
	\bottomrule[1.5pt]
	\multicolumn{7}{c}{\textit{Offline}} \\
	\midrule[1pt]
	VITA~\pub{NIPS22}~\cite{heo2022vita} & ResNet-50 & 32.6 & 53.9 & 39.3 & 30.3 & \second{42.6} \\
	GenVIS~\pub{CVPR23}~\cite{heo2023generalized} & ResNet-50 & \second{37.2} & 58.5 & \highest{42.9} & \second{33.2} & 40.4 \\
    \rowcolor{gray!35}
     DVIS~\pub{ICCV23}~\cite{zhang2023dvis} & ResNet-50 & 36.8 & \second{59.4} & 41.4 & 32.2 & 42.2 \\
     \rowcolor{gray!35}
	DVIS++ & ResNet-50 & \highest{40.9} & \highest{65.2} & \highest{42.9} & \highest{35.4} & \highest{46.6} \\
	\hline
	VITA~\pub{NIPS22}~\cite{heo2022vita} & Swin-L & 41.1 & 63.0 & 44.0 & 39.3 & 44.3 \\
	GenVIS~\pub{CVPR23}~\cite{heo2023generalized} & Swin-L & 44.3 & \second{69.9} & 44.9 & \second{39.9} & 48.4 \\
        \rowcolor{gray!35}
	DVIS~\pub{ICCV23}~\cite{zhang2023dvis} & Swin-L & \second{45.9} & 69.0 & \second{48.8} & 37.2 & \second{51.8} \\
        \rowcolor{gray!35}
	DVIS++ & VIT-L & \highest{50.9} & \highest{75.7} & \highest{52.8} & \highest{40.6} & \highest{55.8} \\ 
	
\bottomrule[1.5pt]
\end{tabular}
\end{table}

\vspace{2mm}\noindent\textbf{OVIS.} OVIS is a highly challenging VIS dataset, featuring lengthy videos (\textit{e.g.}, 500 frames) and a multitude of occlusion scenes, which brings it closer to real-world scenarios. As a result, OVIS serves as a more suitable benchmark for evaluating the performance of VIS methods in practical applications. Table~\ref{tab:ytvis} also presents a performance comparison of DVIS++ with other SOTA methods. When employing ResNet-50 as the backbone, DVIS++ achieves an AP of 37.2 in online mode and 41.2 in offline mode, surpassing all existing VIS methods. Notably, DVIS++ with ResNet-50 even outperforms the cutting-edge method MinVIS \cite{huang2022minvis} with the Swin-L backbone \cite{liu2021swin} (41.2 \textit{vs.} 39.4). It is worth highlighting that DVIS++ exhibits even more pronounced advantages in complex scenes by effectively modeling the spatio-temporal representation of objects. Specifically, it surpasses the previous SOTA method RefineVIS \cite{abrantes2023refinevis} by 1.8 AP on the simple YouTube-VIS 2021 dataset and 7.5 AP on the complex OVIS dataset. Moreover, when employing a frozen pre-trained VIT-L backbone, DVIS++ achieves an AP of 49.6 in online mode and 53.4 in offline mode, outperforming the previous SOTA method GenVIS \cite{heo2023generalized} by 4.4 AP (49.6 \textit{vs.} 45.2) and 8.0 AP (53.4 \textit{vs.} 45.4), respectively.

In addition, only our proposed DVIS and DVIS++ demonstrate significant performance improvement between offline and online modes (48.6 \textit{vs.} 45.9 and 53.4 \textit{vs.} 49.6) in complex scenarios. Meanwhile, the contemporary methods GenVIS and RefineVIS exhibit similar or even worse performance in offline mode compared to that in the online mode (45.4 \textit{vs.} 45.2 and 46.0 \textit{vs.} 46.1). This proves the importance of our decoupling strategy in effectively utilizing spatio-temporal features.

\begin{table}[t!]
\centering
\caption{Results on the validation set of VSPW. mVC$_{k}$ means that a clip with $k$ frames is used.}
\label{tab:vspw}
\setlength{\tabcolsep}{2.5pt}
\begin{tabular}{l l | c c c}
\toprule[1.5pt]
	\multirow{2}{*}{Method} & \multirow{2}{*}{Backbone} &  \multicolumn{3}{c}{VSPW}\\
	~ & ~ & mVC$_{8}$ &  mVC$_{16}$ & mIOU\\
	\midrule[1pt]
	Mask2Former~\pub{CVPR22}~\cite{cheng2022masked} & ResNet-50 & 87.5 & 82.5 & 38.4 \\
	Video-kMax~\pub{ArXiv23}~\cite{shin2023video} & ResNet-50 & 86.0 & 81.4 & 44.3 \\
	Tube-Link~\pub{ICCV23}~\cite{li2023tube} & ResNet-50 & 89.2 & 85.4 & 43.4 \\
	MPVSS~\pub{NIPS23}~\cite{weng2023mask} & ResNet-50 & 84.1 & 77.2 & 37.5 \\
        \rowcolor{gray!35}
	DVIS(online)~\pub{ICCV23}~\cite{zhang2023dvis} & ResNet-50 & 92.0 & 90.9 & 46.6\\
        \rowcolor{gray!35}
	DVIS(offline)~\pub{ICCV23}~\cite{zhang2023dvis} & ResNet-50 & \second{93.2} & \second{92.3} & \second{47.2} \\
        \rowcolor{gray!35}
	DVIS++(online) & ResNet-50 &  92.3 & 91.1 & 46.9\\
        \rowcolor{gray!35}
     DVIS++(offline) & ResNet-50 & \highest{93.4} & \highest{92.4} & \highest{48.6} \\
	\midrule[1pt]
	DeepLabv3+~\pub{ECCV18}~\cite{deeplabv3plus2018} & ResNet-101 & 83.5 & 78.4 & 35.7 \\
	TCB~\pub{CVPR21}~\cite{miao2021vspw} & ResNet-101 & 86.9 & 82.1 & 37.5 \\
	Video K-Net~\pub{CVPR22}~\cite{li2022video} & ResNet-101 & 87.2 & 82.3 & 38.0 \\
	MRCFA~\pub{ECCV22}~\cite{sun2022mining} & MiT-B2 & 90.9 & 87.4 & 49.9 \\
	CFFM~\pub{CVPR22}~\cite{sun2022coarse} & MiT-B5 & 90.8 & 87.1 & 49.3 \\
	
	Video K-Net+~\pub{CVPR22}~\cite{li2022video} & Swin-B & 90.1 & 87.8 & 57.2 \\
	
	Video-kMax~\pub{ArXiv23}\cite{shin2023video} & ConvNeXt-L & 91.8 & 88.6 & \second{63.6} \\
	TubeFormer~\pub{CVPR22}~\cite{kim2022tubeformer} & Axial-ResNet-50-B4 & 92.1 & 88.0 & 63.2 \\
	MPVSS~\pub{NIPS23}~\cite{weng2023mask} & Swin-L  & 89.6 & 85.8 & 53.9 \\
	
        \rowcolor{gray!35}
	DVIS(online)~\pub{ICCV23}~\cite{zhang2023dvis} & Swin-L & 95.0 & 94.3 & 61.3 \\
        \rowcolor{gray!35}
	DVIS(offline)~\pub{ICCV23}~\cite{zhang2023dvis} & Swin-L & \second{95.1} & \second{94.4} & 63.3 \\
    \rowcolor{gray!35}
     DVIS++(online) & VIT-L & 95.0 & 94.2 & 62.8 \\ 
     \rowcolor{gray!35}
	DVIS++(offline) & VIT-L & \highest{95.7} & \highest{95.1} & \highest{63.8}\\ 
\bottomrule[1.5pt]
\end{tabular}
\end{table}

\vspace{2mm}\noindent\textbf{YouTube-VIS 2022.} YouTube-VIS 2022 has added a significant number of long videos on top of YouTube-VIS 2021. It has separately evaluated the performance of VIS methods using the AP$^{\rm L}$ metric on these long videos. Table~\ref{tab:ytvis22} shows the performance comparison between DVIS++ and other SOTA methods. When using ResNet-50 as the backbone, DVIS++ achieves AP$^{\rm L}$ scores of 37.2 and 40.9, surpassing MinVIS~\cite{huang2022minvis} with 13.9 AP$^{\rm L}$ (37.2 \textit{vs.} 23.3) and GenVIS~\cite{heo2023generalized} with 3.7 AP$^{\rm L}$ (40.9 \textit{vs.} 37.2), respectively. When using a frozen pre-trained VIT-L as the backbone, DVIS++ achieves an AP$^{\rm L}$ score of 50.9 in offline mode, significantly outperforming VITA~\cite{heo2022vita} and GenVIS~\cite{heo2023generalized} with 9.8 AP$^{\rm L}$ (50.9 \textit{vs.} 41.1) and 6.6 AP$^{\rm L}$ (50.9 \textit{vs.} 44.3), respectively. In the case of long videos, DVIS++ demonstrates a significant performance improvement in offline mode compared to online mode(40.9 \textit{vs.} 37.2 with ResNet-50 and 50.9 \textit{vs.} 37.5 with VIT-L), indicating the crucial importance of effectively utilizing temporal information for processing long videos.

\subsubsection{Performance on Video Semantic Segmentation}
\textbf{VSPW.} VSPW is a challenging large-scale video semantic segmentation dataset. We compare DVIS++ with other VSS methods on the validation set of VSPW, as shown in Table~\ref{tab:vspw}. When using ResNet-50 as the backbone, DVIS++ achieves 92.3 mVC$_8$, 91.1 mVC$_{16}$, and 46.9 mIOU in online mode, surpassing all other methods in terms of video segmentation quality and consistency. DVIS++ outperformes Tube-Link~\cite{li2023tube} by 3.1 VC$_8$, 5.7 VC$_{16}$, and 3.5 mIOU, as well as Video-kMax~\cite{shin2023video} by 7.3 VC$_8$, 9.7 VC$_{16}$, and 2.6 mIOU. When running in offline mode, DVIS++ achieves 93.4 VC$_8$, 92.4 VC$_{16}$, and 48.6 mIOU, resulting in improvements of 1.1 VC$_8$, 1.3 VC$_{16}$, and 1.7 mIOU compared to online mode. When using a frozen pre-trained VIT-L as the backbone, DVIS++ achieves 95.0 VC$_8$, 94.2 VC$_{16}$, and 62.8 mIOU in online mode, and 95.7 VC$_8$, 95.1 VC$_{16}$, and 63.8 mIOU in offline mode, surpassing all competitors without any specific design for VSS. 

\begin{table}[t!]
\centering
\caption{Results on the validation set of VIPSeg. VPQ$^{\rm Th}$ and VPQ$^{\rm St}$ refer to the VPQ (Video Panoptic Quality) on the ``thing" objects and the ``stuff" objects, respectively..}
\label{tab:vipseg}
\setlength{\tabcolsep}{2.5pt}
\begin{tabular}{l l | c c c c}
\toprule[1.5pt]
	\multirow{2}{*}{Method} & \multirow{2}{*}{Backbone} &  \multicolumn{4}{c}{VIPSeg}\\
	~ & ~ & VPQ &  VPQ$^{\rm Th}$ & VPQ$^{\rm St}$ & STQ \\
	\midrule[1pt]
	VPSNet~\pub{CVPR20}~\cite{kim2020video}& ResNet-50 & 14.0 & 14.0 & 14.2 & 20.8 \\
	VPSNet-SiamTrack~\pub{CVPR21}~\cite{woo2021learning}& ResNet-50 &  17.2 & 17.3 & 17.3 & 21.1 \\
	VIP-Deeplab~\pub{CVPR21}~\cite{qiao2021vip}& ResNet-50 & 16.0 & 12.3 & 18.2 & 22.0 \\
	Clip-PanoFCN~\pub{CVPR22}~\cite{miao2022large}& ResNet-50 & 22.9 & 25.0 & 20.8 & 31.5 \\
	Video K-Net~\pub{CVPR22}~\cite{li2022video}& ResNet-50 & 26.1 & - & - & 31.5 \\
	TarVIS~\pub{CVPR23}~\cite{athar2023tarvis}& ResNet-50 & 33.5 & 39.2 & 28.5 & \second{43.1} \\
     Tube-Link~\pub{ICCV23}~\cite{li2023tube}& ResNet-50 & 39.2 & - & - & 39.5 \\
     Video-kMax~\pub{ArXiv23}\cite{shin2023video}& ResNet-50 & 38.2 & - & - & 39.9 \\
     \rowcolor{gray!35}
	DVIS(online)~\pub{ICCV23}~\cite{zhang2023dvis} & ResNet-50 & 39.4 & 38.6 & 40.1 & 36.3 \\
    \rowcolor{gray!35}
     DVIS(offline)~\pub{ICCV23}~\cite{zhang2023dvis} & ResNet-50 & \second{43.2} & \second{43.6} & \second{42.8} & 42.8 \\
        \rowcolor{gray!35}
	DVIS++(online) & ResNet-50 & 41.9 & 41.0 & 42.7 & 38.5 \\
    \rowcolor{gray!35}
     DVIS++(offline) & ResNet-50 & \highest{44.2} & \highest{44.5} & \highest{43.9} & \highest{43.6} \\
	\hline
	TarVIS~\pub{CVPR23}~\cite{athar2023tarvis} & Swin-L & 48.0 & 58.2 & 39.0 & 52.9 \\
        \rowcolor{gray!35}
	DVIS(online)~\pub{ICCV23}~\cite{zhang2023dvis} & Swin-L & 54.7 & 54.8 & 54.6 & 47.7\\ 
        \rowcolor{gray!35}
	DVIS(offline)~\pub{ICCV23}~\cite{zhang2023dvis} & Swin-L & \second{57.6} & \second{59.9} & \highest{55.5} & \second{55.3} \\ 
    \rowcolor{gray!35}
     DVIS++(online) & VIT-L & 56.0 & 58.0 & 54.3 & 49.8 \\ 
        \rowcolor{gray!35}
	DVIS++(offline) & VIT-L & \highest{58.0} & \highest{61.2} &\second{55.2} & \highest{56.0} \\ 
	
\bottomrule[1.5pt]
\end{tabular}
\end{table}
\begin{table*}[t!]
\centering
 \caption{Open-vocabulary VIS performance on the validation sets of YouTube-VIS 2019, 2021, and OVIS. The supervised methods are trained using the combination training sets of the video datasets and tested on the validation sets. $\dag$ indicates using the same association method as MinVIS~\cite{huang2022minvis}.}
 \label{tab:ov-vis}
\setlength{\tabcolsep}{1.2pt}
\begin{tabular}{l l c | c c c c c| c c c c c | c c c c c}
\toprule[1.5pt]
	\multirow{2}{*}{Method} & \multirow{2}{*}{Backbone} & \multirow{2}{*}{Training} & \multicolumn{5}{c|}{Youtube-VIS 2019} &  \multicolumn{5}{c|}{Youtube-VIS 2021} & \multicolumn{5}{c}{OVIS} \\
	~ & ~ & ~ & AP &  AP$_{\rm 50}$ & AP$_{\rm 75}$ &  AR$_{\rm 1}$ & AR$_{\rm 10}$ & AP &  AP$_{\rm 50}$ & AP$_{\rm 75}$ &  AR$_{\rm 1}$ & AR$_{\rm 10}$ & AP &  AP$_{\rm 50}$ & AP$_{\rm 75}$ &  AR$_{\rm 1}$ & AR$_{\rm 10}$ \\
	\midrule[1pt]
	\multicolumn{18}{c}{\textit{Supervised}} \\
	\midrule[1pt]
	FC-CLIP$\dag$~\pub{NIPS23}~\cite{yu2023convolutions} &  ConvNext-L & - & 57.3 & 80.1 & 63.6 & 49.7 & 61.8 & 52.8 & 73.6 & 58.9 & 43.2 & 57.6 & 31.5 & 55.5 & 29.9 & 16.6 & 34.9 \\
        \rowcolor{gray!35}
	OV-DVIS++(online) &  ConvNext-L & - & \second{60.1} & \second{83.6} & \second{65.6} & \highest{54.8} & \second{68.6} & \second{56.0} & \second{78.9} & \highest{62.5} & \highest{46.4} & \second{63.6} & \second{38.9} & \second{65.0} & \second{40.6} & \highest{17.4} & \second{45.6} \\
        \rowcolor{gray!35}
	OV-DVIS++(offline) &  ConvNext-L & - & \highest{61.1} & \highest{84.3} & \highest{66.4} & \highest{54.8} & \highest{70.3} & \highest{56.7} & \highest{80.6} & \second{62.3} & \second{46.3} & \highest{65.4} & \highest{40.6} & \highest{67.8} & \highest{42.3} & \highest{17.4} & \highest{48.9} \\
	\midrule[1pt]
	\multicolumn{18}{c}{\textit{Zero-Shot}} \\
	\midrule[1pt]
	Detic-SORT~\pub{ECCV22}~\cite{zhou2022detecting} \pub{ICIP16}~\cite{bewley2016simple} & ResNet-50 & LVIS & 14.6 & - & - & - & - & 12.7 & - & - & - & - & 6.7 & - & - & - & - \\
	Detic-OWTB~\pub{ECCV22}~\cite{zhou2022detecting} \pub{CVPR22}~\cite{liu2022opening}& ResNet-50 & LVIS & 17.9 & - & - & - & - & 16.7 & - & - & - & - & 9.0 & - & - & - & - \\
	MindVLT~\pub{ICCV23}~\cite{wang2023towards} & ResNet-50 & LVIS & 23.1 & - & - & - & - & 20.9 & - & - & - & - & 11.4 & - & - & - & - \\
	FC-CLIP$\dag$~\pub{NIPS23}~\cite{yu2023convolutions} & ResNet-50 & COCO & 28.9 & 43.9 & 29.9 & 32.4 & 41.2 & 25.5 & 40.0 & 26.8 & 28.0 & 37.0 & 11.8 & 25.1 & 10.5 & 8.5 & 16.4  \\
        \rowcolor{gray!35}
	OV-DVIS++(online) & ResNet-50 & COCO & \highest{34.5} & \second{50.6} & \highest{39.2} & \second{39.5} & \second{49.5} & \second{30.9} & \second{46.7} & \highest{34.8} & \second{34.4} & \second{45.8} & \highest{14.8} & \highest{31.2} & \highest{13.1} & \highest{10.5} & \second{24.7} \\
        \rowcolor{gray!35}
	OV-DVIS++(offline) & ResNet-50 & COCO & \second{34.4} & \highest{51.8} & \highest{39.2} & \highest{40.0} & \highest{50.6} & \highest{31.0} & \highest{48.5} & \second{34.3} & \highest{34.6} & \highest{46.8} & \second{13.0} & \second{28.4} & \second{10.9} & \second{9.8} & \highest{25.1} \\
	\hline 
	Detic-SORT~\pub{ECCV22}~\cite{zhou2022detecting} \pub{ICIP16}~\cite{bewley2016simple} & Swin-B & LVIS & 23.8 & - & - & - & - & 21.6 & - & - & - & - & 11.7 & - & - & - & - \\
	Detic-OWTB~\pub{ECCV22}~\cite{zhou2022detecting} \pub{CVPR22}~\cite{liu2022opening} & Swin-B & LVIS & 30.0 & - & - & - & - & 27.1 & - & - & - & - & 13.6 & - & - & - & -\\
	MindVLT~\pub{ICCV23}~\cite{wang2023towards} & Swin-B & LVIS & 37.6 & - & - & - & - & 33.9 & - & - & - & - & 17.5 & - & - & - & -\\
	FC-CLIP$\dag$~\pub{NIPS23}~\cite{yu2023convolutions} & ConvNext-L & COCO & 44.1 & 63.6 & 48.8 & 43.2 & 57.4 & 42.2 & 62.4 & 46.6 & 38.5 & \highest{55.1} & 20.9 & 37.6 & 20.8 & 12.5 & 24.2 \\
        \rowcolor{gray!35}
	OV-DVIS++(online) & ConvNext-L & COCO & \highest{48.8} & \second{69.9} & \second{54.9} & \highest{47.2} & \second{59.6} & \highest{44.5} & \highest{65.8} & \highest{49.3} & \highest{39.1} & \highest{55.1}  & \highest{24.0} & \highest{46.2} & \highest{22.3} & \highest{13.2} & \second{30.9} \\
        \rowcolor{gray!35}
	OV-DVIS++(offline) & ConvNext-L & COCO & \second{48.7} & \highest{70.2} & \highest{55.2} & \second{47.1} & \highest{59.9} & \second{44.2} & \second{65.6} & \second{48.4} & 38.4 & \highest{55.1} & \second{21.6} & \second{44.4} & 19.1 & 12.1 & \highest{31.0}\\
\bottomrule[1.5pt]
 \end{tabular} \vspace{-2mm}
\end{table*}

\begin{table*}[t!]
\centering
 \caption{Open-vocabulary VSS and VPS performance on the VSPW and VIPSeg validation sets.}
 \label{tab:ov-vss-vps}
\setlength{\tabcolsep}{6.5pt}
\begin{tabular}{l l c | c c c | c c c c}
\toprule[1.5pt]
	\multirow{2}{*}{Method} & \multirow{2}{*}{Backbone} & \multirow{2}{*}{Training} & \multicolumn{3}{c|}{VSPW} &  \multicolumn{4}{c}{VIPSeg} \\
	~ & ~ & ~ & mVC$_{8}$ &  mVC$_{16}$ & mIOU & VPQ &  VPQ$^{\rm Th}$ & VPQ$^{\rm St}$ & STQ \\
	\midrule[1pt]
	\multicolumn{10}{c}{\textit{Supervised}} \\
	\midrule[1pt]
	FC-CLIP$\dag$~\pub{NIPS23}~\cite{yu2023convolutions} &  ConvNext-L & - & 91.8 & 90.7 & 47.9 & 49.5 & 52.1 & 47.3 & 42.2 \\
        \rowcolor{gray!35}
	OV-DVIS++(online) &  ConvNext-L & - & \second{94.8} & \second{94.0} & \second{53.3} & \second{49.7} & \second{52.2} & \second{47.5} & \second{46.7} \\
        \rowcolor{gray!35}
	OV-DVIS++(offline) &  ConvNext-L & - & \highest{95.0} & \highest{94.3} & \highest{56.4} & \highest{51.7} & \highest{54.6} & \highest{49.2} & \highest{51.0} \\
	\midrule[1pt]
	\multicolumn{10}{c}{\textit{Zero-Shot}} \\
	\midrule[1pt]
	FC-CLIP$\dag$~\pub{NIPS23}~\cite{yu2023convolutions} & ResNet-50 & COCO & 84.9 & 82.7 & 24.3 & 22.3 & 25.5 & 19.1  & 19.7\\
        \rowcolor{gray!35}
	OV-DVIS++(online) & ResNet-50 & COCO & \highest{92.7} & \highest{91.5} & \second{27.6} & \highest{24.4} & \highest{26.8} & \highest{22.4} & \second{22.0} \\
        \rowcolor{gray!35}
	OV-DVIS++(offline) & ResNet-50 & COCO & \second{92.4} & \second{91.3} & \highest{28.4} & \second{23.8} & \second{26.4} & \second{21.4} & \highest{24.4} \\
	\hline 
	FC-CLIP$\dag$~\pub{NIPS23}~\cite{yu2023convolutions} & ConvNext-L & COCO & 89.9 & 88.4 & 28.9 & 27.9 & 30.9 & 25.0 & 24.2 \\
        \rowcolor{gray!35}
	OV-DVIS++(online) & ConvNext-L & COCO & \highest{94.2} & \highest{93.3} & \highest{34.3} & \second{28.9} & \second{31.3} & \second{26.8} & \second{28.4} \\
        \rowcolor{gray!35}
	OV-DVIS++(offline) & ConvNext-L & COCO & \second{93.9} & \second{93.0} & \second{34.1} & \highest{30.4} & \highest{31.9} & \highest{29.1} & \highest{32.2} \\
\bottomrule[1.5pt]
 \end{tabular} \vspace{-2mm}
\end{table*}

\subsubsection{Performance on Video Panoptic Segmentation}
\textbf{VIPSeg.} VIPSeg is a large-scale video panoptic segmentation dataset with a wide range of real scenes and categories. The comparative results of DVIS++ and other methods are shown in Table~\ref{tab:vipseg}. When using ResNet-50 as the backbone, DVIS++ achieves a VPQ of 41.9 in online mode, surpassing DVIS~\cite{zhang2023dvis} 2.5 VPQ and Tube-Link~\cite{li2023tube} 2.7 VPQ. In offline mode, DVIS++ achieves a VPQ of 44.2 and an STQ of 43.6, surpassing all contemporaneous methods. When a frozen pre-trained VIT-L is used as the backbone, DVIS++ achieves a VPQ of 56.0 in online mode and 58.0 in offline mode, surpassing TarVIS~\cite{athar2023tarvis} 10.0 VPQ (58.0 \textit{vs.} 48.0).

The experimental results demonstrate that DVIS and DVIS++ are powerful universal video segmentation baselines.

\begin{table}[t!]
\centering
\caption{Ablation of the main components. All models utilize ResNet-50 as the backbone and are evaluated on the OVIS dataset. The baseline $\mathcal{M}0$ is MinVIS~\cite{zhang2023dvis}. RT and TR represent the Referring Tracker and Temporal Refiner, respectively. CL$_{\rm Seg}$, CL$_{\rm RT}$, and CL$_{\rm TR}$ indicate the incorporation of contrastive loss during the training of the segmenter, referring tracker, and temporal refiner, respectively. DTS refers to the Denosing Training Strategy. AP$_{\rm l}$, AP$_{\rm m}$, and AP$_{\rm h}$ refer to AP on objects with light, medium, and heavy occlusion, respectively.}
\label{tab:ablation}
\setlength{\tabcolsep}{2.5pt}
\begin{tabular}{c| c c c c c c | c c c c}
\toprule[1.5pt]
	$\mathcal{M}$ & CL$_{\rm Seg}$ & RT & DTS & CL$_{\rm RT}$ & TR & CL$_{\rm TR}$ & AP & AP$_{\rm l}$ & AP$_{\rm m}$ & AP$_{\rm h}$ \\
	\midrule[1pt]
	0 & & & & & &     & 25.0 & 39.0 & 29.3 & 8.9 \\
	1 & \checkmark & & & & &     & 25.8 & 47.0 & 30.6 & 9.3 \\
	2 & \checkmark &\checkmark & & & &     & 32.8 & 51.2 & 37.6 & 14.1 \\
	3 & \checkmark &\checkmark &\checkmark & & &     & 36.5 & 53.9 & 41.9 & 18.2 \\
	4 & \checkmark &\checkmark &\checkmark &\checkmark & &     & 37.2 & 55.0 & 43.1 & 17.8 \\
	5 & \checkmark &\checkmark &\checkmark &\checkmark &\checkmark &     & 41.2 & 53.0 & 46.0 & 23.9 \\
	6 & \checkmark &\checkmark &\checkmark &\checkmark &\checkmark &\checkmark & 40.6 & 56.3 & 45.3 & 21.7 \\
\bottomrule[1.5pt]
\end{tabular}\vspace{-2mm}
\end{table}
\begin{table}[t!]
\centering
\caption{The ablation of the cross-attention types and the selection of the initial values. $Q_{\rm Seg}$ refers to the object representations outputted by the segmenter, while Matched $Q_{\rm Seg}$ refers to $Q_{\rm Seg}$ after being matched using the Hungarian algorithm. Zero and Learnable indicate using a zero vector and a learnable embedding as the initial value, respectively.}
\label{tab:referring tracker}
\setlength{\tabcolsep}{3.5pt}
\begin{tabular}{c| c c c c c | c c c }
\toprule[1.5pt]
	& AP & AP$_{50}$ & AP$_{75}$ & AR$_{1}$ &  AR$_{10}$ & AP$_{\rm l}$ & AP$_{\rm m}$ & AP$_{\rm h}$\\
	\midrule[1pt]
	\multicolumn{9}{c}{\textit{Cross Attention Type}} \\
	\midrule[1pt]
	Standard & 28.3 & 51.4 & 26.4 & 14.2 & 32.4 & 47.8 & 32.5 & 12.0 \\
	Referring & 32.8 & 55.9 & 32.5 & 15.4 & 38.7 & 51.2 & 37.6 & 14.1\\
	\midrule[1pt]
	\multicolumn{9}{c}{\textit{Initial Value}} \\
	\midrule[1pt]
	$Q_{Seg}$ & 31.9 & 55.2 & 31.0 & 15.1 & 38.0 & 48.2 & 37.0 & 14.2 \\
	Matched $Q_{Seg}$ & 32.8 & 55.9 & 32.5 & 15.4 & 38.7 & 51.2 & 37.6 & 14.1 \\
	Zero & 33.0 & 56.8 & 32.2 & 16.9 & 38.7 & 51.5 & 38.5 & 14.1\\
	Learnable & 33.1 & 57.2 & 32.5 & 15.4 & 38.5 & 49.7 & 38.2 & 15.9 \\
\bottomrule[1.5pt]
\end{tabular}\vspace{-2mm}
\end{table}
\begin{table}[t!]
\centering
\caption{The ablation of the main components in the temporal refiner. w/o denotes without.}
\label{tab:temporal refiner}
\setlength{\tabcolsep}{2.0pt}
\begin{tabular}{c| c c c c c | c c c }
\toprule[1.5pt]
	& AP & AP$_{50}$ & AP$_{75}$ & AR$_{1}$ &  AR$_{10}$ & AP$_{\rm l}$ & AP$_{\rm m}$ & AP$_{\rm h}$\\
	\midrule[1pt]
	Temporal Refiner & 40.2 & 68.1 & 39.9 & 16.8 & 46.2 & 54.2 & 46.0 & 21.9 \\
	w/o Short-Term Conv. & 40.0 & 68.1 & 39.3 & 16.6 & 46.3 & 54.6 & 45.8 & 21.2 \\
	w/o Long-Term Attn. & 37.0 & 63.7 & 35.1 & 16.0 & 43.3 & 53.0 & 43.3 & 17.6 \\
	w/o Sefl-Attn. & 38.5 & 66.9 & 36.4 & 16.4 & 44.3 & 53.3 & 43.9 & 20.1 \\
	w/o Cross-Attn. & 39.4 & 66.9 & 37.8 & 16.7 & 44.6 & 55.2 & 45.5 & 20.8 \\
\bottomrule[1.5pt]
\end{tabular}
\end{table}
\begin{table}[t!]
\centering
\caption{The ablation of the denoising training strategy. $S10$ is identical to $M4$ in Table~\ref{tab:ablation}. P refers to the probability of adding simulated noise to the initial values. Iter represents the number of training iterations. WA, CC, and RS refer to the random weighted averaging, random cropping \& concatenating, and random shuffling noise simulation strategies, respectively.}
\label{tab:denoising strategy}
\setlength{\tabcolsep}{1.6pt}
\begin{tabular}{c| c c | c c c| c c c c c | c c c}
\toprule[1.5pt]
	$S$ & P & Iter & WA & CC & RS & AP & AP$_{50}$ & AP$_{75}$ & AR$_{1}$ & AR$_{10}$ & AP$_{\rm l}$ & AP$_{\rm m}$ & AP$_{\rm h}$ \\
	\midrule[1pt]
	0 & 0.0 & 40k & & & & 33.1 & 56.2 & 32.9 & 15.4 & 38.9 & 51.4 & 37.9 & 13.8 \\
	1 & 1.0 & 40k& \checkmark & & & 35.3 & 60.8 & 34.7 & 15.7 & 40.9 & 51.0 & 41.2 & 16.7 \\
	2 & 1.0 & 40k&  & \checkmark & & 34.7 & 60.6 & 33.5 & 16.5 & 40.7 & 51.3 & 39.1 & 17.1    \\
	3 & 1.0 & 40k &  & & \checkmark & 33.7 & 57.7 & 33.1 & 16.2 & 39.4 & 51.0 & 38.6 & 15.2 \\
	\midrule[1pt]
	4 & 1.0 & 160k& \checkmark & & & 36.7 & 62.3 & 37.6 & 16.2 & 42.0 & 52.5 & 42.1 & 18.0  \\
	5 & 1.0 & 160k&  & & \checkmark & 35.6 & 61.2 & 35.7 & 16.1 & 41.2 & 53.2 & 42.1 & 16.6   \\
	\midrule[1pt]
	6 & 0.8 & 40k & \checkmark & & &  35.3 & 61.7 & 33.8 & 16.7 & 40.6 & 53.2 & 41.0 & 16.7 \\
	7 & 0.5 & 40k & \checkmark & &  & 35.8 & 60.9 & 35.6 & 16.2 & 41.3 & 53.9 & 42.1 & 16.9 \\
	8 & 0.3 & 40k & \checkmark & &  & 34.3 & 60.0 & 33.7 & 16.1 & 40.6 & 49.9 & 39.1 & 15.5 \\
	\midrule[1pt]
	9 & 0.5 & 160k & \checkmark & &  & 36.5 & 61.8 & 36.2 & 15.6 & 42.3 & 53.9 & 41.9 & 18.2 \\
	10 & 0.8 & 160k & \checkmark & &  & 37.2 & 62.8 & 37.3 & 15.8 & 42.9 & 55.0 & 43.1 & 17.8 \\
	
\bottomrule[1.5pt]
\end{tabular}
\end{table}

\subsubsection{\hspace{-3mm}Performance on Open-Vocabulary Video Segmentation}

\noindent\textbf{VIS.} We compare the zero-shot performance of OV-DVIS++ in open-vocabulary instance segmentation with other SOTA methods on the VIS datasets, and the results are presented in Table \ref{tab:ov-vis}. When utilizing ResNet-50 as the backbone and training solely on the COCO dataset, OV-DVIS++ achieves AP scores of 34.5, 30.9, and 14.8 on YouTube-VIS 2019, 2021, and OVIS datasets, respectively. Remarkably, OV-DVIS++ outperforms the previous SOTA method MindVLT~\cite{wang2023towards} by 11.4, 10.0, and 3.4 AP on YouTube-VIS 2019, 2021, and OVIS datasets, despite MindVLT being trained on the LVIS dataset~\cite{gupta2019lvis}, which encompasses more diverse categories than the COCO dataset. Furthermore, when employing ConvNext-L~\cite{liu2022convnet} as the backbone, OV-DVIS++ achieves AP scores of 48.8, 44.5, and 24.0 on YouTube-VIS 2019, 2021, and OVIS datasets, respectively, surpassing all other open-vocabulary video instance segmentation methods.

When running in online mode, OV-DVIS++ outperforms FC-CLIP$\dag$ (a combination of FC-CLIP~\cite{yu2023convolutions} and MinVIS~\cite{huang2022minvis} that we designed) in zero-shot performance by 5.6, 5.4, and 3.0 AP on the YouTube 2019, 2021, and OVIS datasets, respectively. Therefore, the referring tracker demonstrates its stronger tracking capabilities than heuristic algorithms, even with limited training data (training only on the COCO dataset). However, the pseudo-videos generated through affine transformations of images are too simplistic, lacking complex scenarios such as occlusions. Thus, when trained solely on the COCO dataset, OV-DVIS++ does not exhibit performance advantages in offline mode compared to online mode and even experiences performance degradation on the complex OVIS dataset (21.6 \textit{vs.} 24.0). 

When jointly trained on the image and video datasets, OV-DVIS++ achieves 60.1, 56.0, and 38.9 AP in online mode on the YouTube 2019, 2021, and OVIS datasets, respectively. In offline mode, OV-DVIS++ shows a performance improvement of 1.0, 0.7, and 1.7 AP on the YouTube 2019, 2021, and OVIS datasets compared to online mode.

\noindent\textbf{VSS \& VPS.} As there are currently no methods for open-vocabulary video semantic and panoptic segmentation, we solely compare OV-DVIS++ with FC-CLIP$\dag$. The results are presented in Table~\ref{tab:ov-vss-vps}. When utilizing ResNet-50 as the backbone, OV-DVIS++ outperforms FC-CLIP$\dag$ in zero-shot performance by 3.3 mIOU and 2.1 VPQ. When ConvNext-L is employed as the backbone, OV-DVIS++ surpasses FC-CLIP by 5.4 mIOU and 1.0 VPQ. Besides the performance advantages, OV-DVIS++ exhibits significantly higher temporal consistency in segmentation results compared to FC-CLIP (91.3 mVC$_{16}$ \textit{vs.} 82.7 mVC$_{16}$ and 93.0 mVC$_{16}$ \textit{vs.} 88.4 mVC$_{16}$).

\subsection{Ablation Study}
We conduct ablation experiments on the validation set of OVIS to verify the effectiveness of the proposed components. The baseline, MinVIS~\cite{huang2022minvis} ($\mathcal{M}0$ in Table~\ref{tab:ablation}), consists of mask2former and a simple heuristic association algorithm. Table~\ref{tab:ablation} illustrates how DVIS++ is constructed based on this baseline and presents the impact of each component on performance.

\noindent\textbf{Referring Tracker.} The referring tracker is designed to replace heuristic association algorithms by modeling the tracking task as a reference denoising task. As shown in Table~\ref{tab:ablation} ($\mathcal{M}2$ \textit{vs.} $\mathcal{M}1$), this change results in a significant improvement of 7.0 AP, 4.2 AP$_{\rm l}$, 7.0 AP$_{\rm m}$, and 4.8 AP$_{\rm h}$. This fully demonstrates that the referring tracker can learn association capabilities that far exceed those from the heuristic algorithm, especially for occluded objects.

In the referring tracker, the referring cross-attention we designed is the most crucial core component responsible for inter-frame information propagation. We replace referring cross-attention with standard cross-attention and observe a drastic drop in performance, as shown in Table~\ref{tab:referring tracker}.

In addition, the initial value of the referring tracker's input is also important. We attempt various initial value selections, and the results are shown in Table ~\ref{tab:referring tracker}. Firstly, the object representation $Q_{Seg}$ outputted by the segmenter is directly used as the initial value, resulting in an AP of 31.9. When the matched $Q_{seg}$ obtained from the heuristic matching algorithm is used as the initial value, the model achieves an AP of 32.8. This slight improvement comes from the reduction of noise contained in the initial value.

We also attempt to use zero vector and learnable embedding as the initial value. In this case, the input for the referring tracker is the same for all objects, and the denoising task becomes a more challenging reconstruction task. When zero vector is used as the initial value, the referring tracker achieves a performance of 33.0 AP, surpassing the performance achieved by using matched $Q_{Seg}$ as the initial value. When learnable embedding is used as the initial value, the referring tracker achieves a performance of 33.1 AP. It performs better on heavily occluded objects than zero initialization but worse on lightly occluded objects.

Through the attempts above, we discover that modeling more challenging tasks can enhance the tracking performance learned by the referring tracker. This revelation has motivated us to improve the referring tracker's performance by incorporating simulated noise into the initial value, which will be elaborated upon in the following discussion of the denoising training strategy. Considering the synergistic effect of the denoising training strategy, despite using learnable embedding as the initial value yields the optimal outcomes, we ultimately opt for matched $Q_{Seg}$.

\begin{figure*}[t!]
    \centering
\begin{minipage}[c]{1.00\linewidth}
\includegraphics[width=0.163\linewidth]{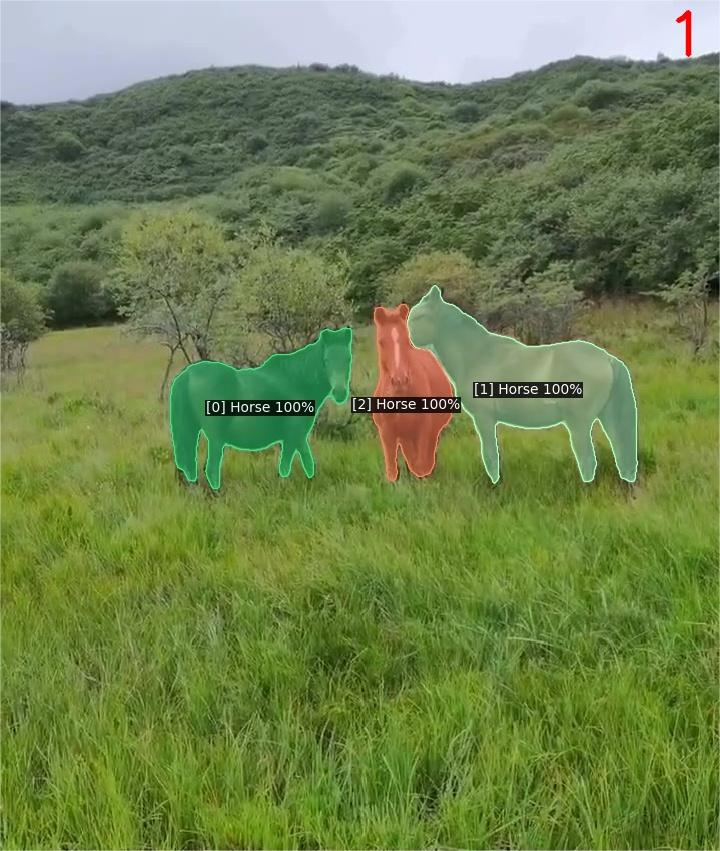}
\includegraphics[width=0.163\linewidth]{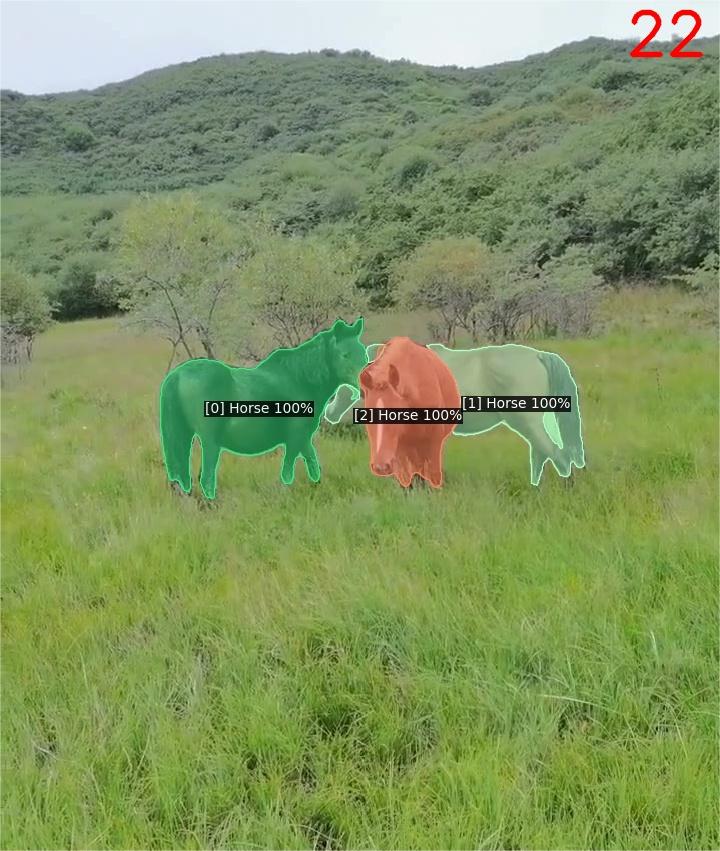}
\includegraphics[width=0.163\linewidth]{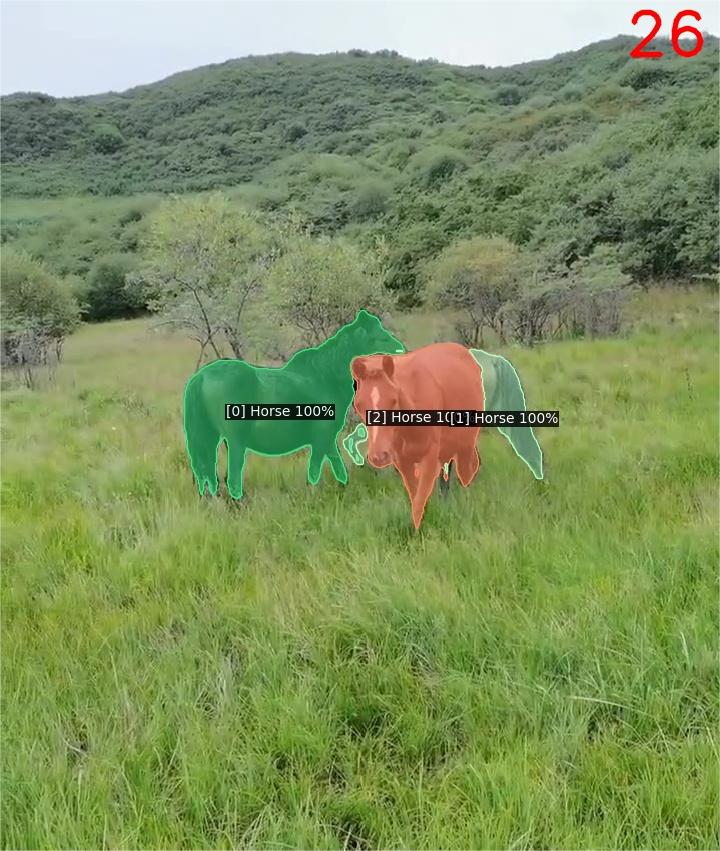}
\includegraphics[width=0.163\linewidth]{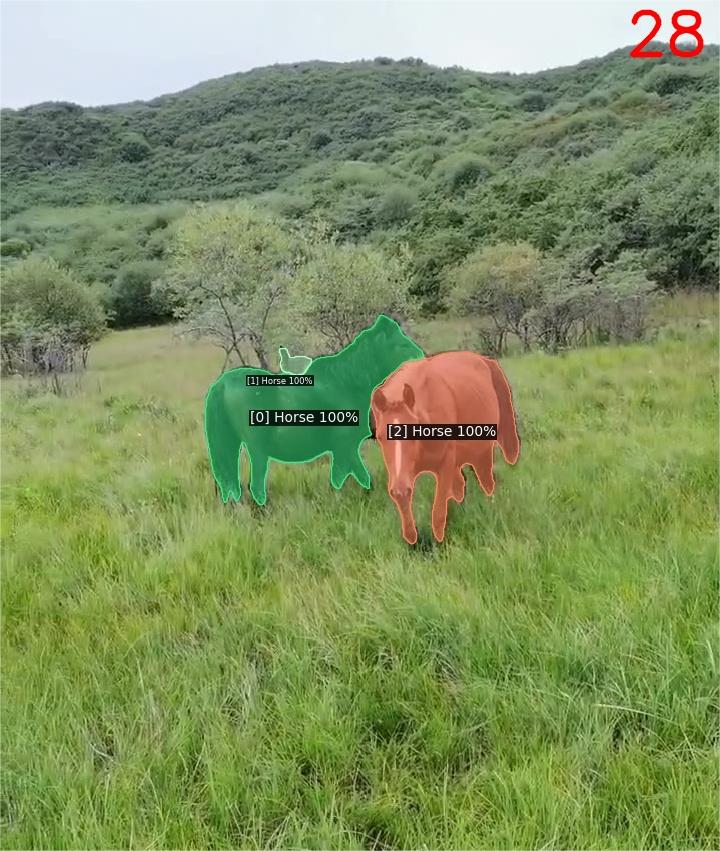}
\includegraphics[width=0.163\linewidth]{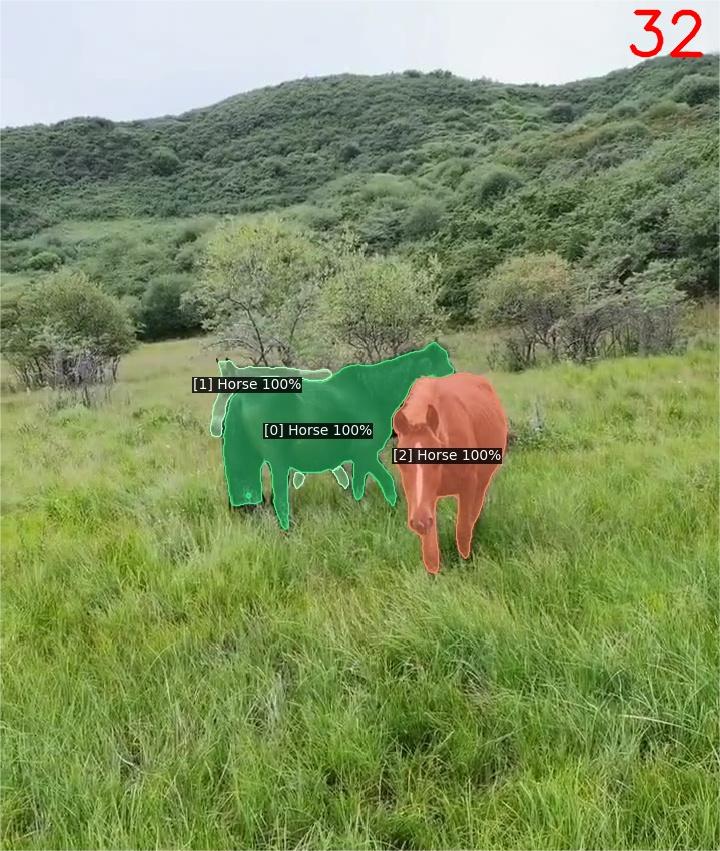}
\includegraphics[width=0.163\linewidth]{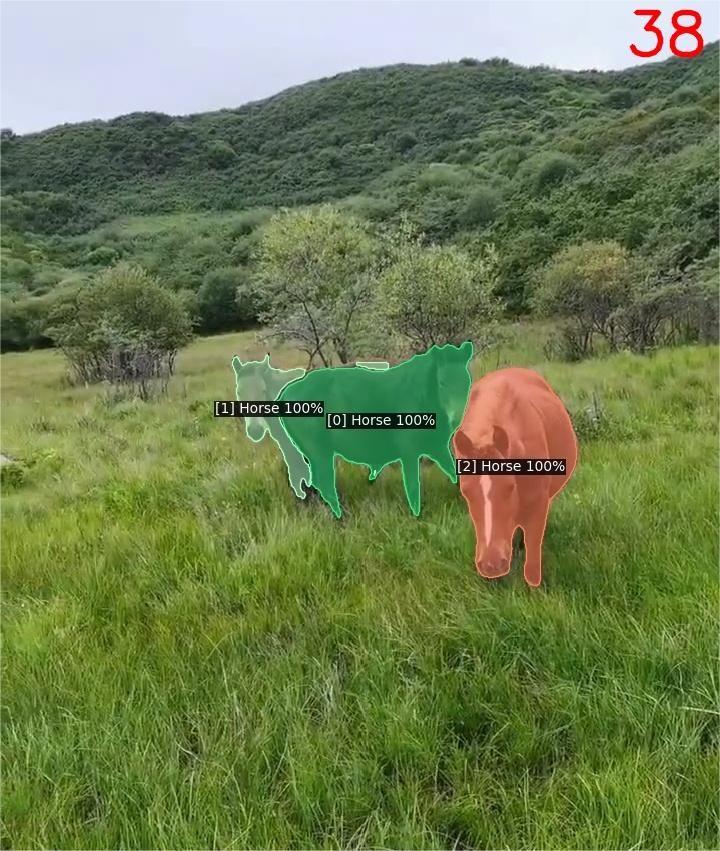}
\end{minipage}\hfill
\begin{minipage}[c]{1.0\linewidth}
\includegraphics[width=0.163\linewidth]{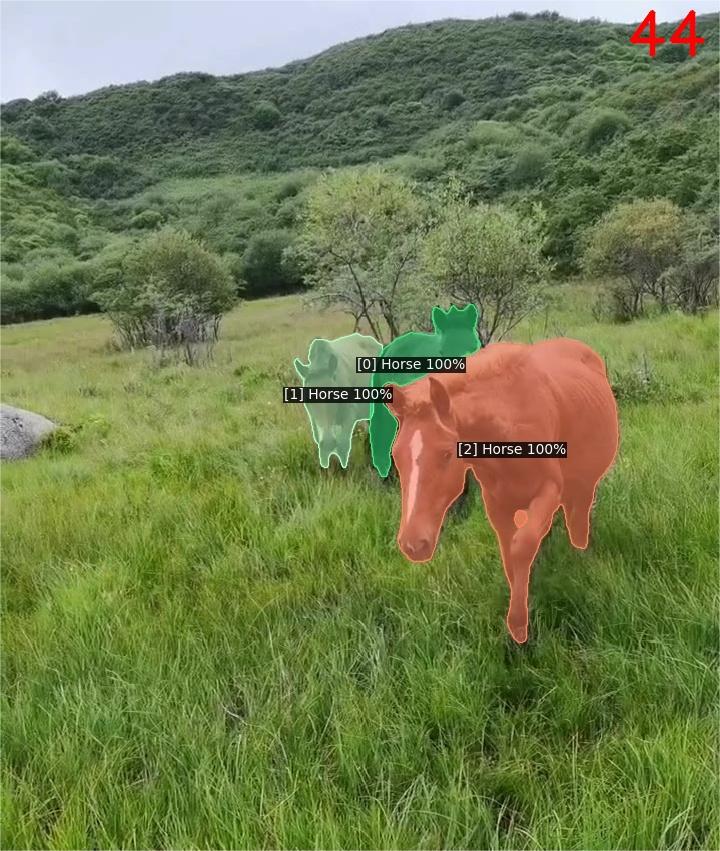}
\includegraphics[width=0.163\linewidth]{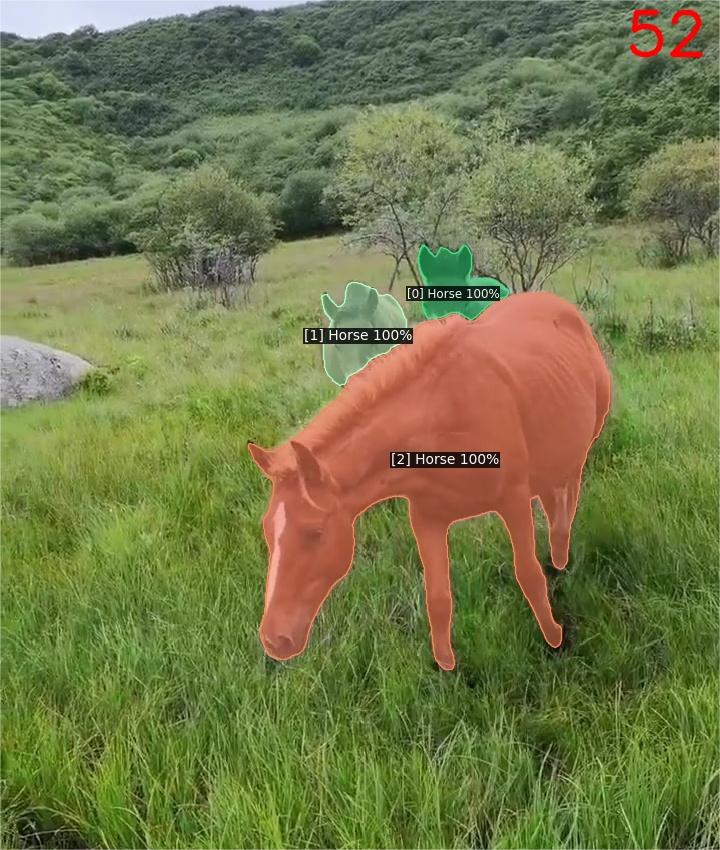}
\includegraphics[width=0.163\linewidth]{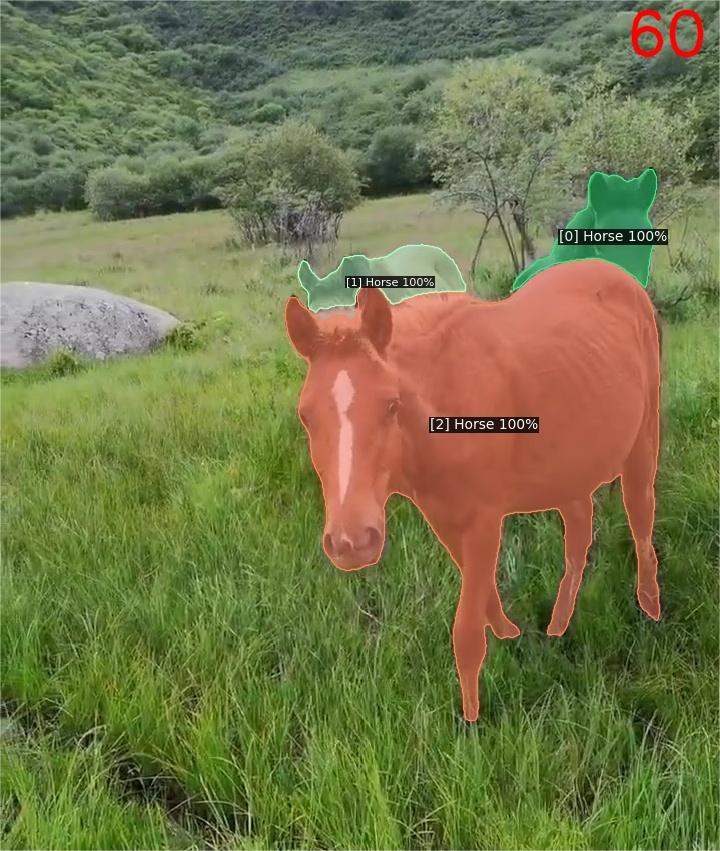}
\includegraphics[width=0.163\linewidth]{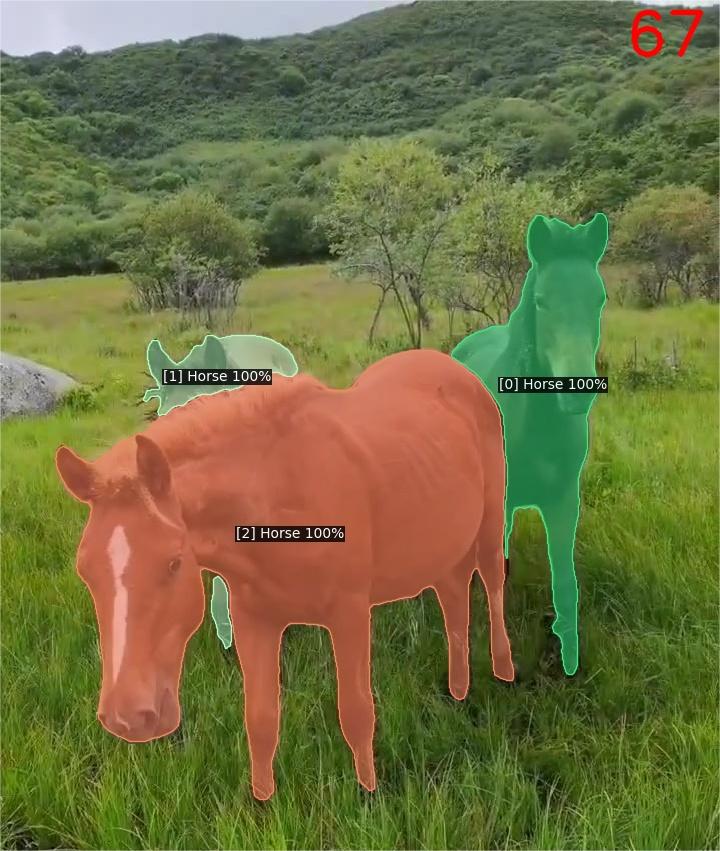}
\includegraphics[width=0.163\linewidth]{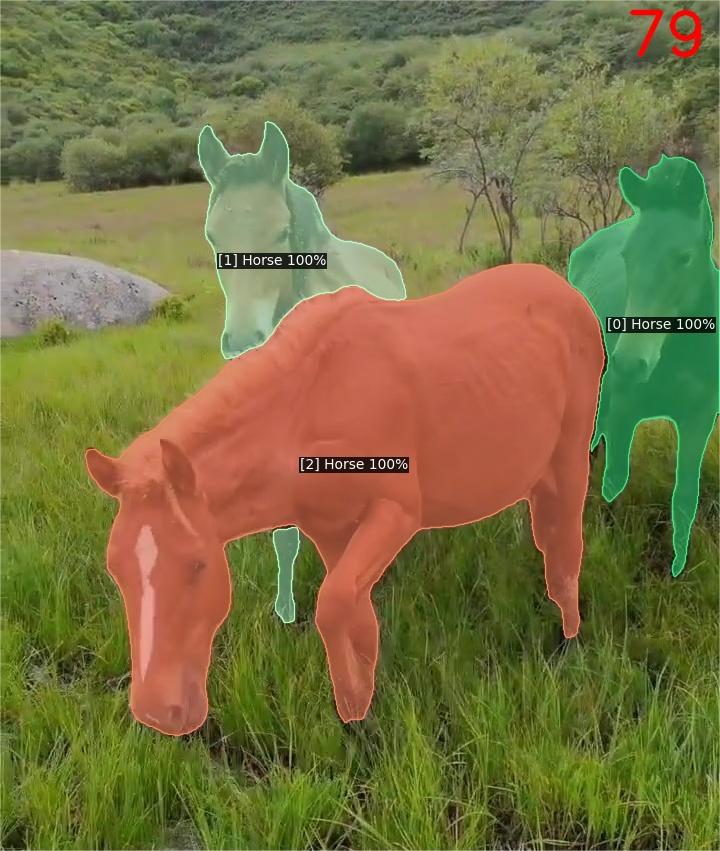}
\includegraphics[width=0.163\linewidth]{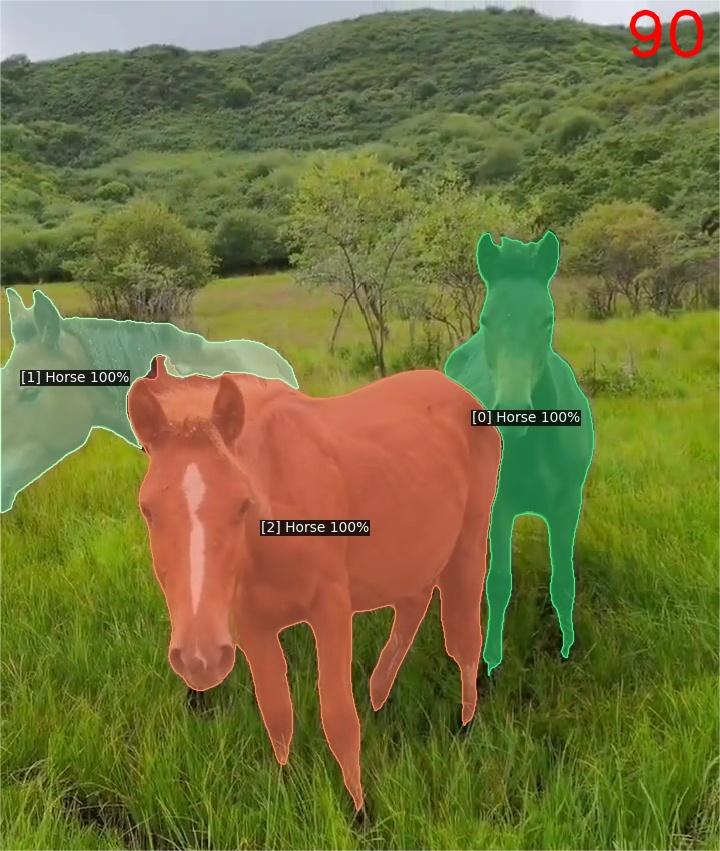}
\end{minipage}\hfill\vspace{1mm}

\begin{minipage}[c]{1.0\linewidth}
\includegraphics[width=0.163\linewidth]{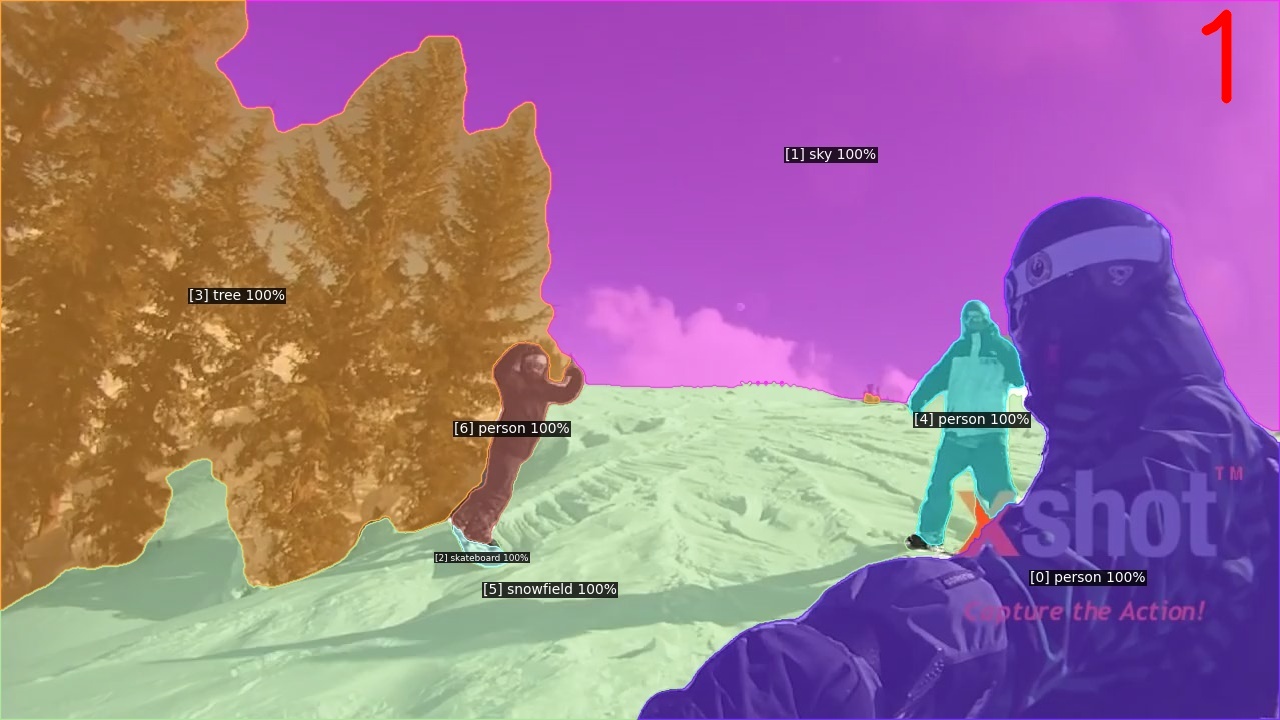}
\includegraphics[width=0.163\linewidth]{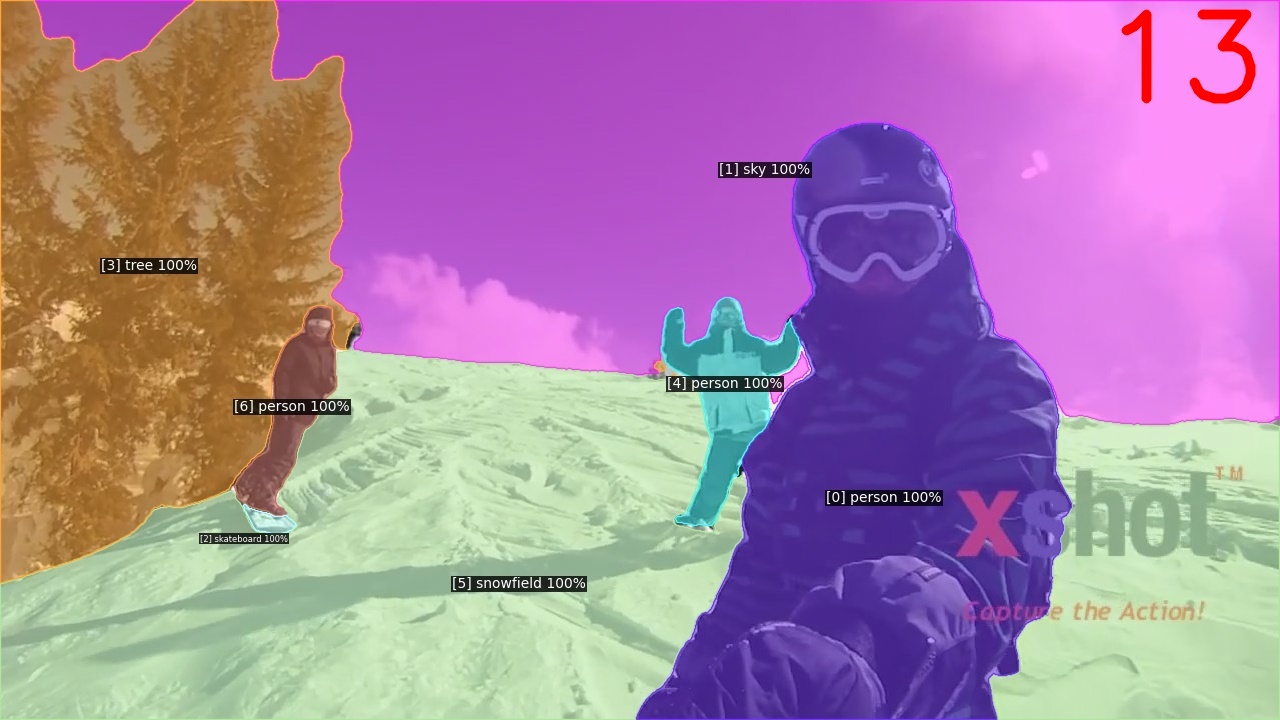}
\includegraphics[width=0.163\linewidth]{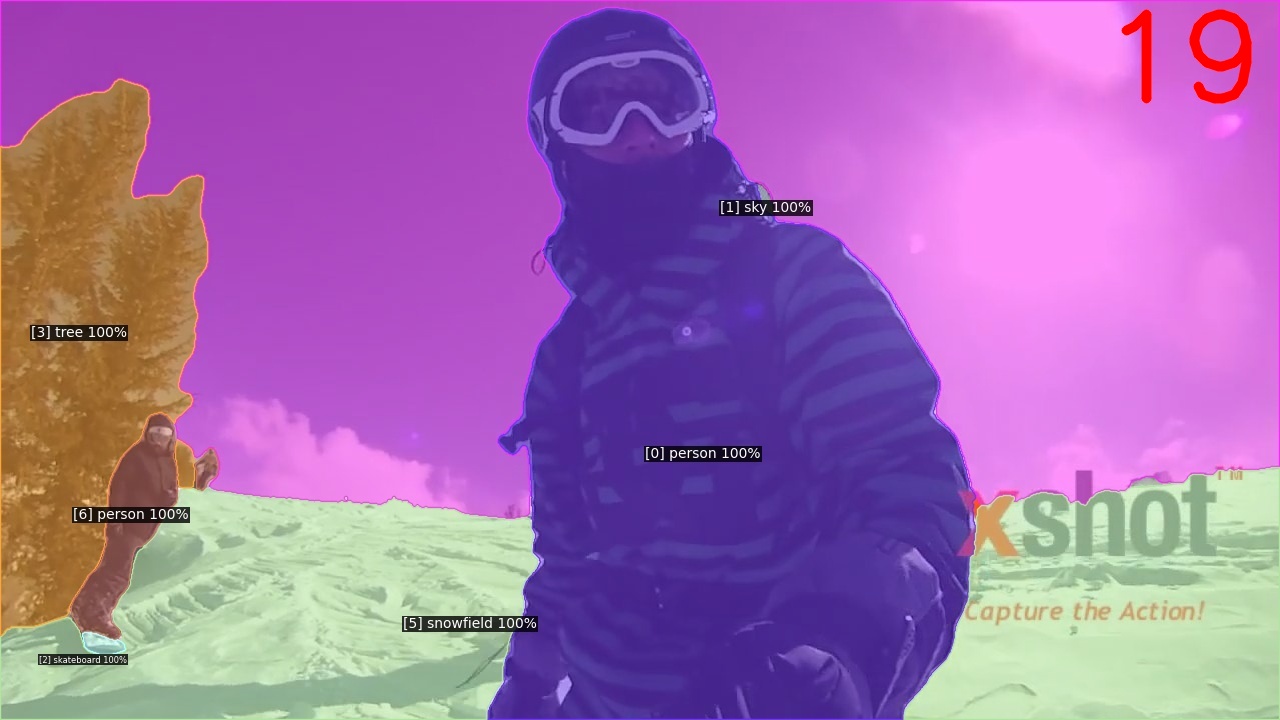}
\includegraphics[width=0.163\linewidth]{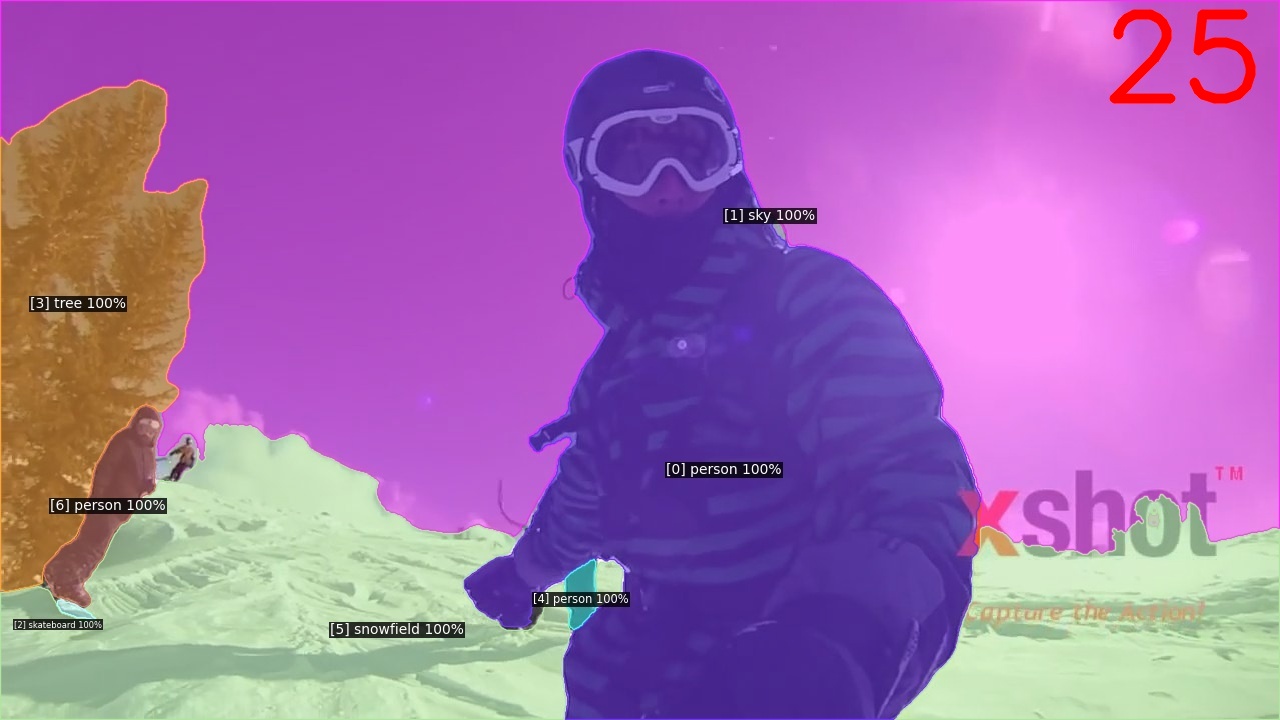}
\includegraphics[width=0.163\linewidth]{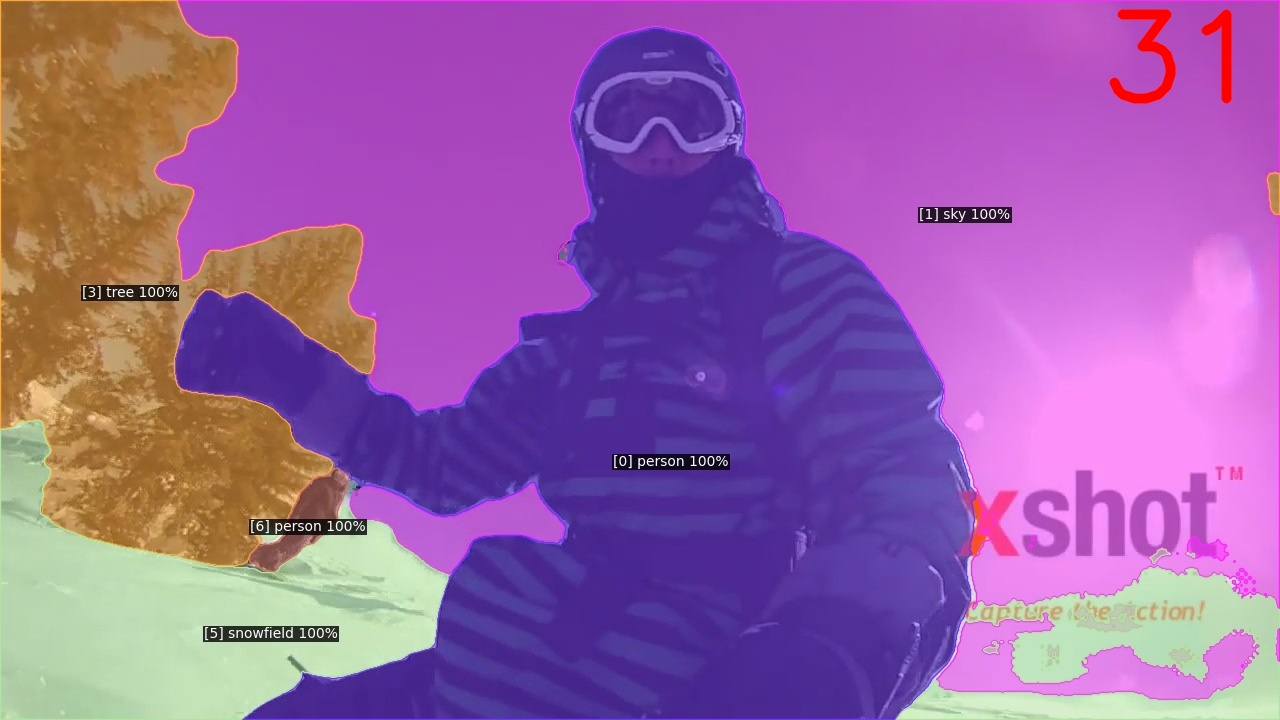}
\includegraphics[width=0.163\linewidth]{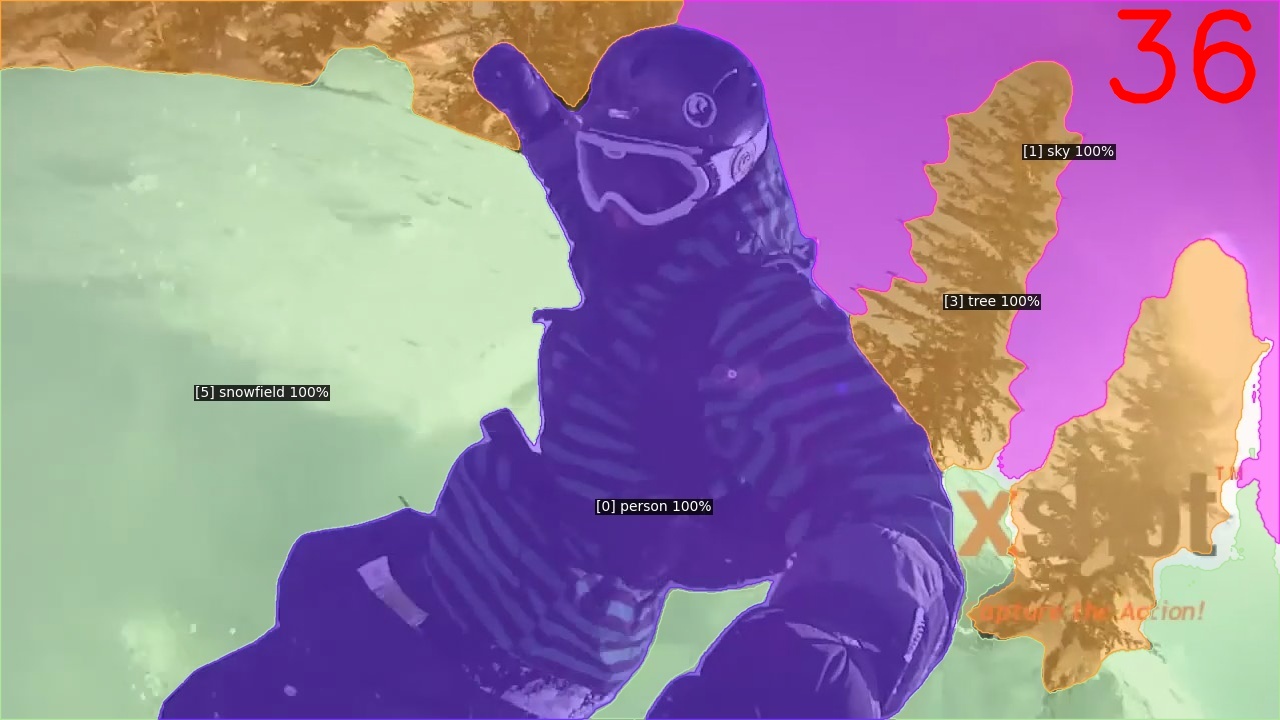}
\end{minipage}\hfill

\begin{minipage}[c]{1.0\linewidth}
\includegraphics[width=0.163\linewidth]{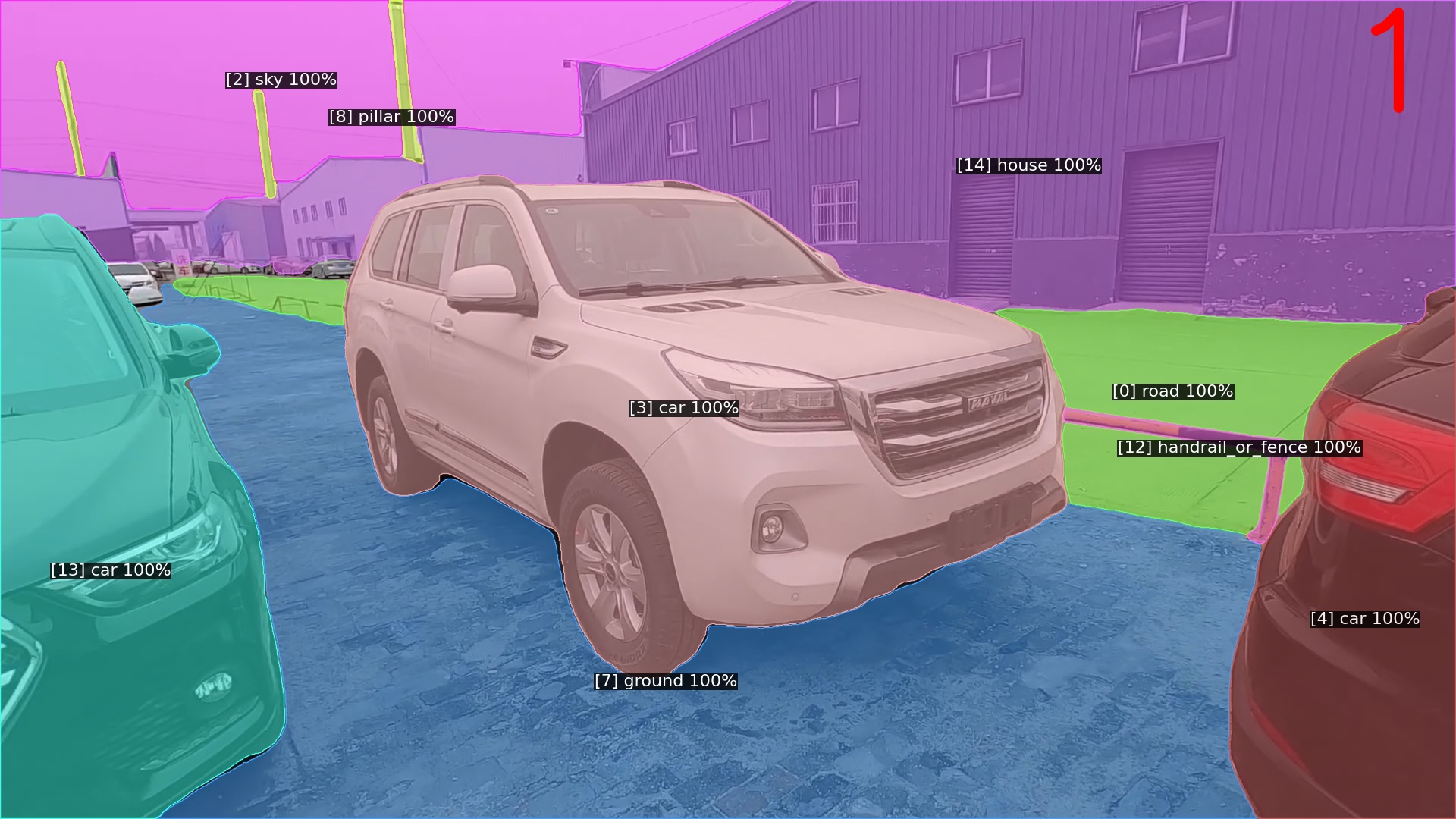}
\includegraphics[width=0.163\linewidth]{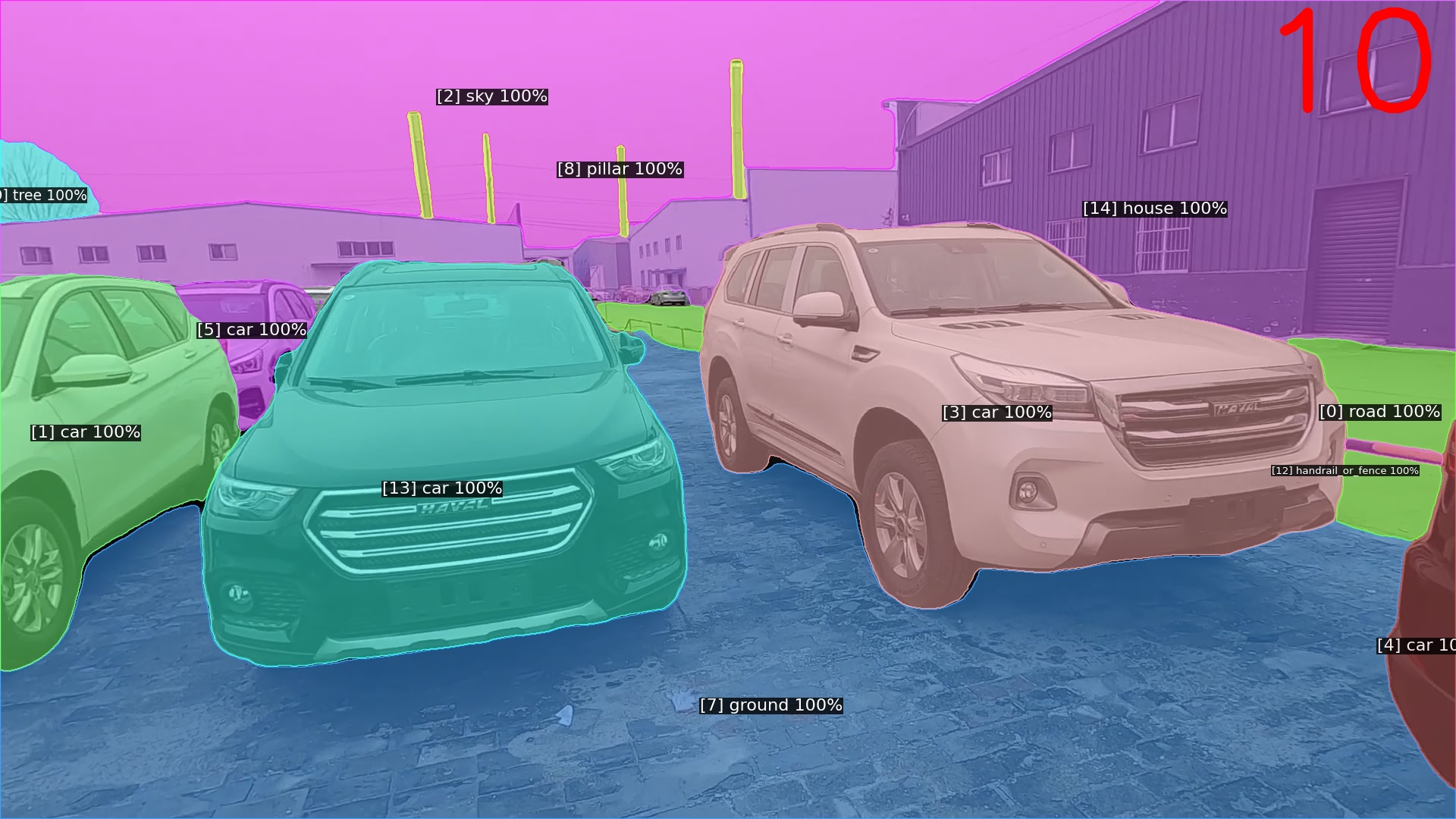}
\includegraphics[width=0.163\linewidth]{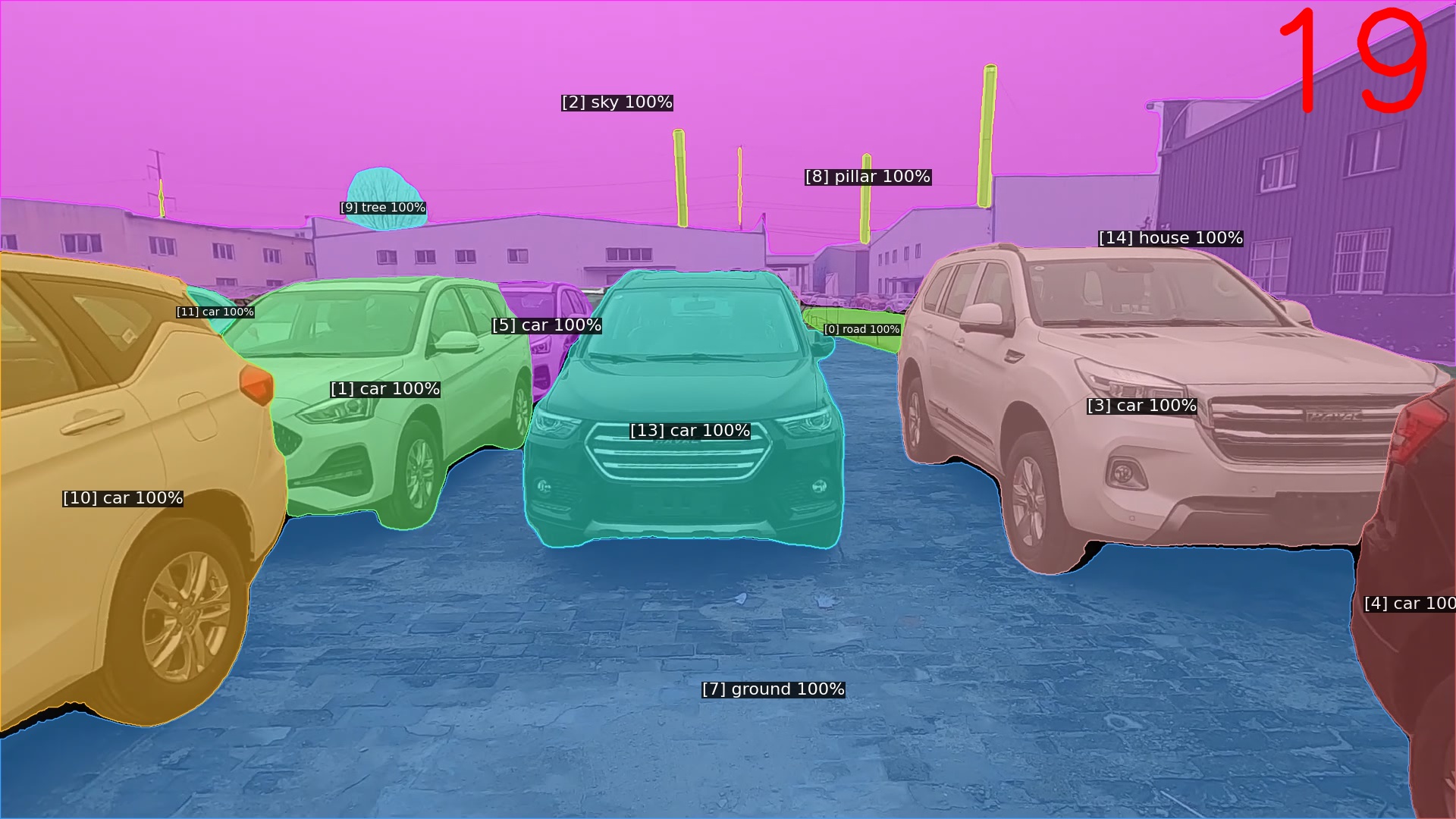}
\includegraphics[width=0.163\linewidth]{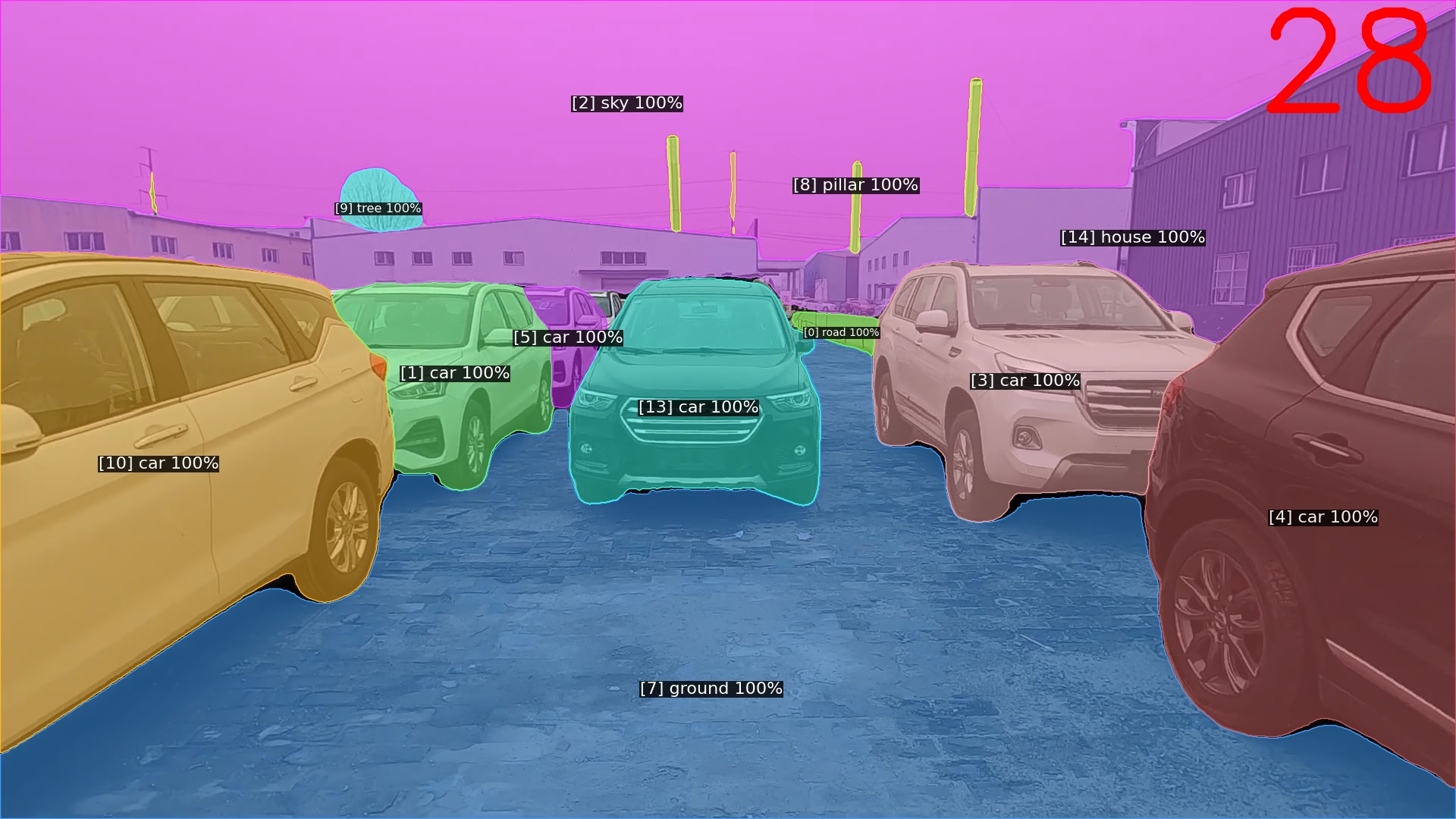}
\includegraphics[width=0.163\linewidth]{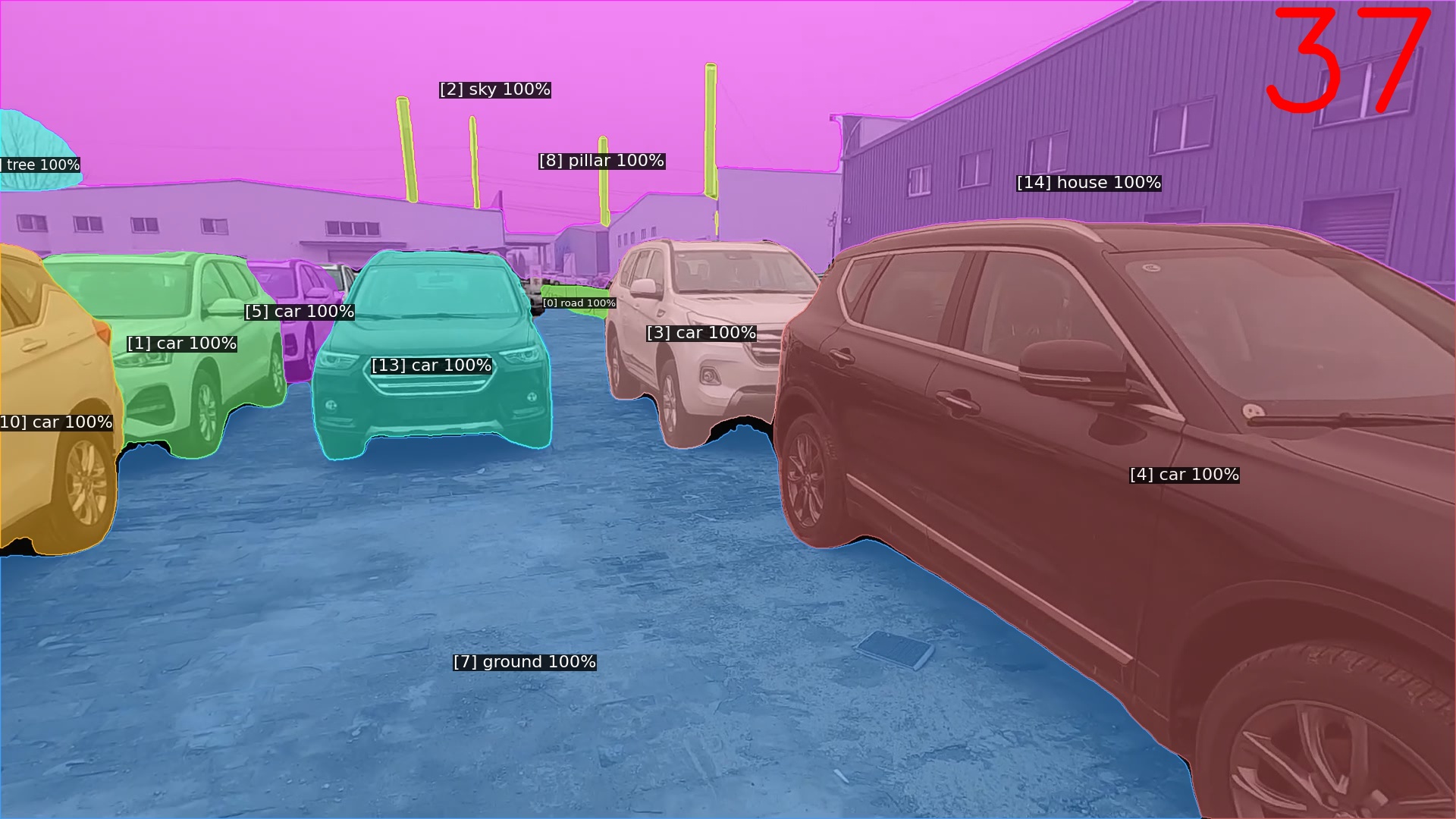}
\includegraphics[width=0.163\linewidth]{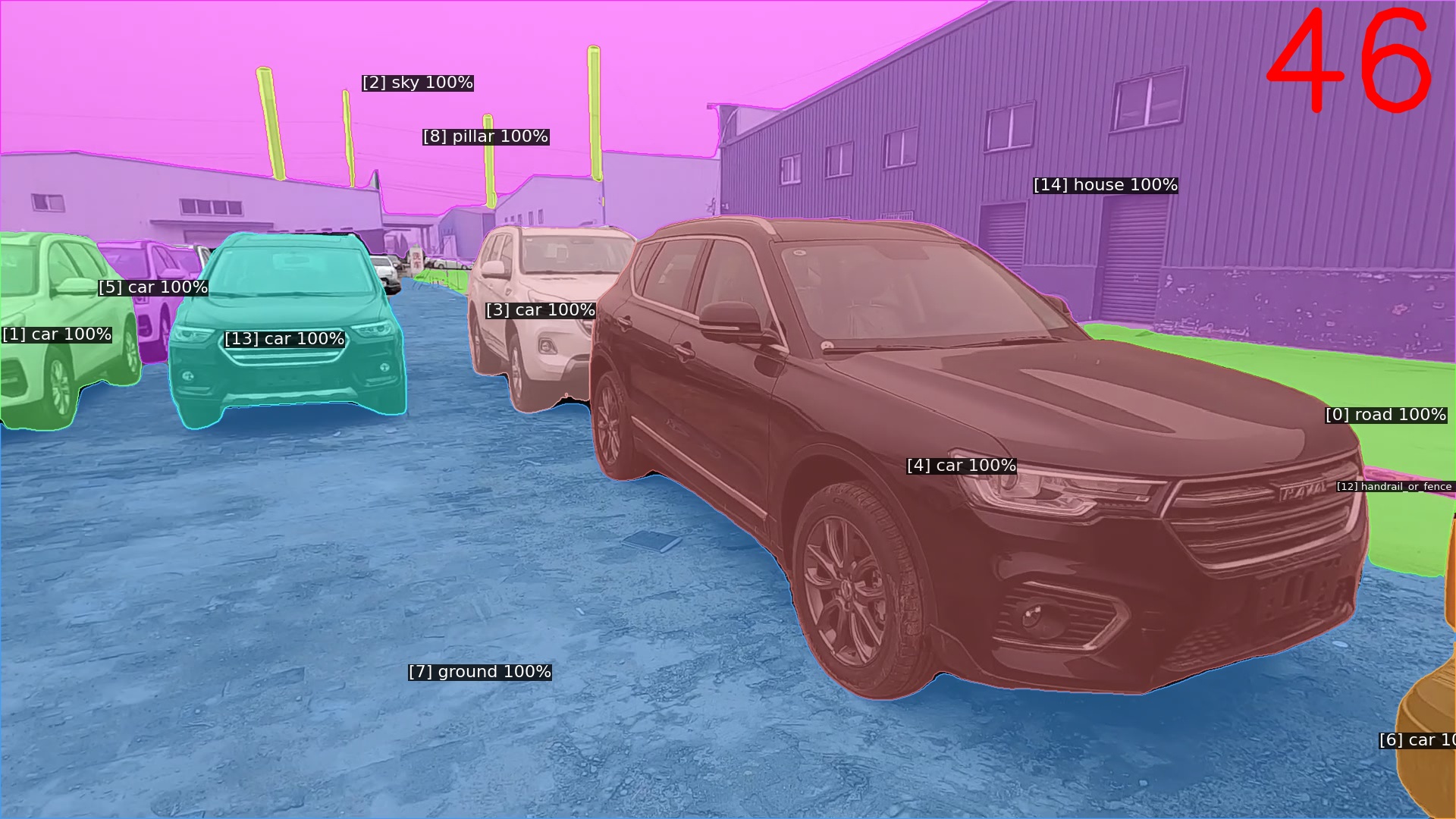}
\end{minipage}\hfill\vspace{1mm}

\begin{minipage}[c]{1.00\linewidth}
\includegraphics[width=0.163\linewidth]{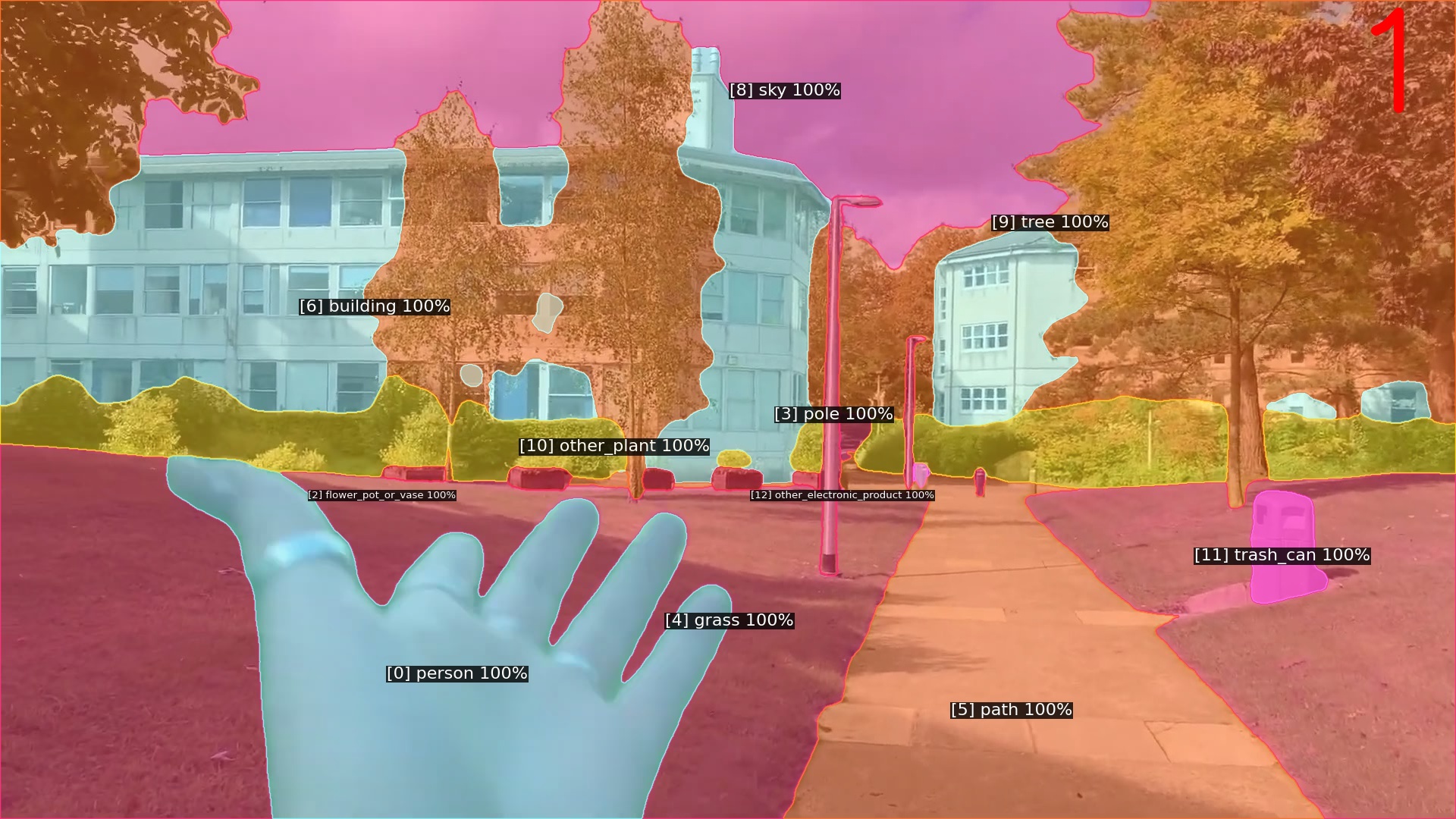}
\includegraphics[width=0.163\linewidth]{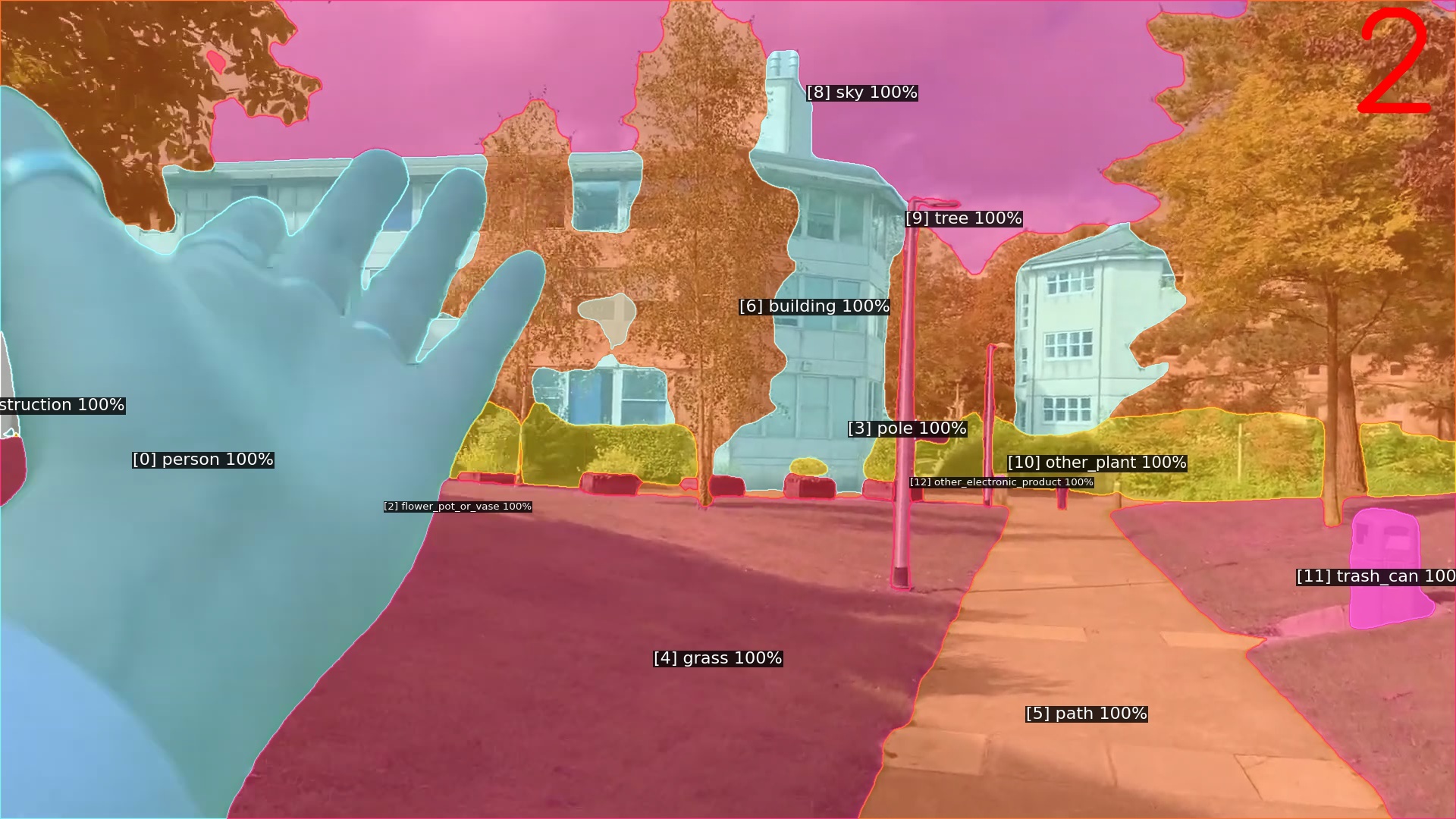}
\includegraphics[width=0.163\linewidth]{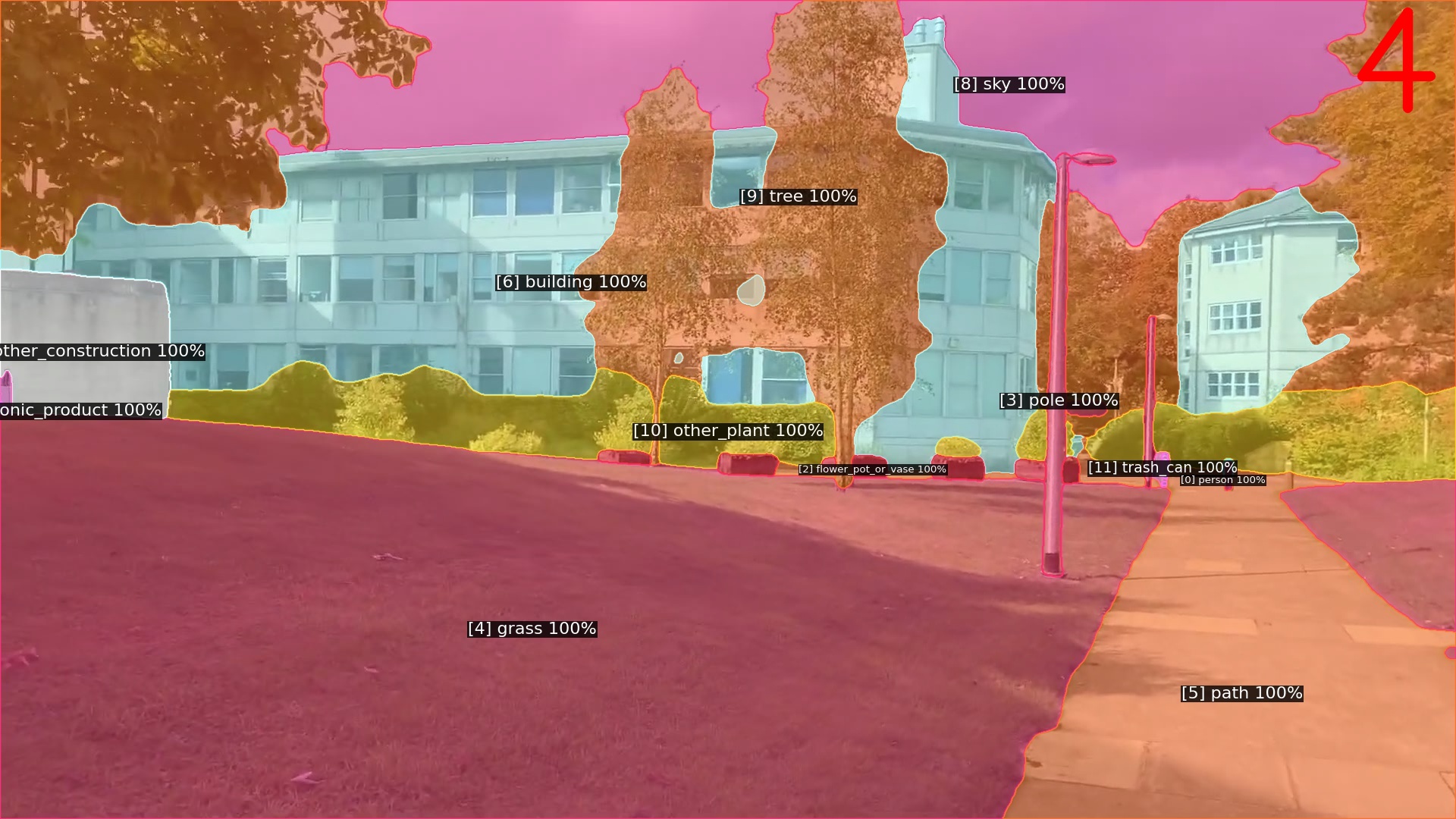}
\includegraphics[width=0.163\linewidth]{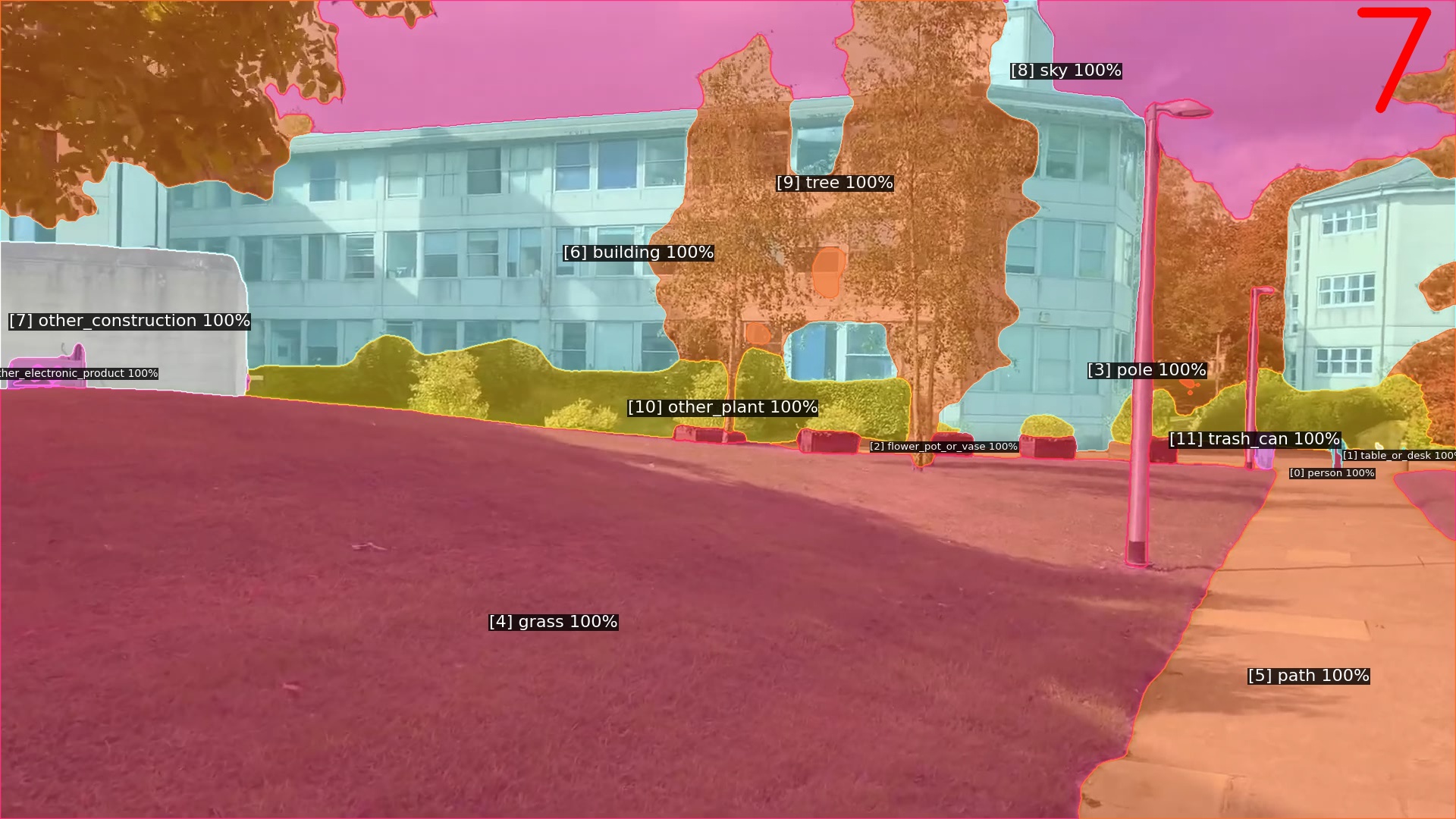}
\includegraphics[width=0.163\linewidth]{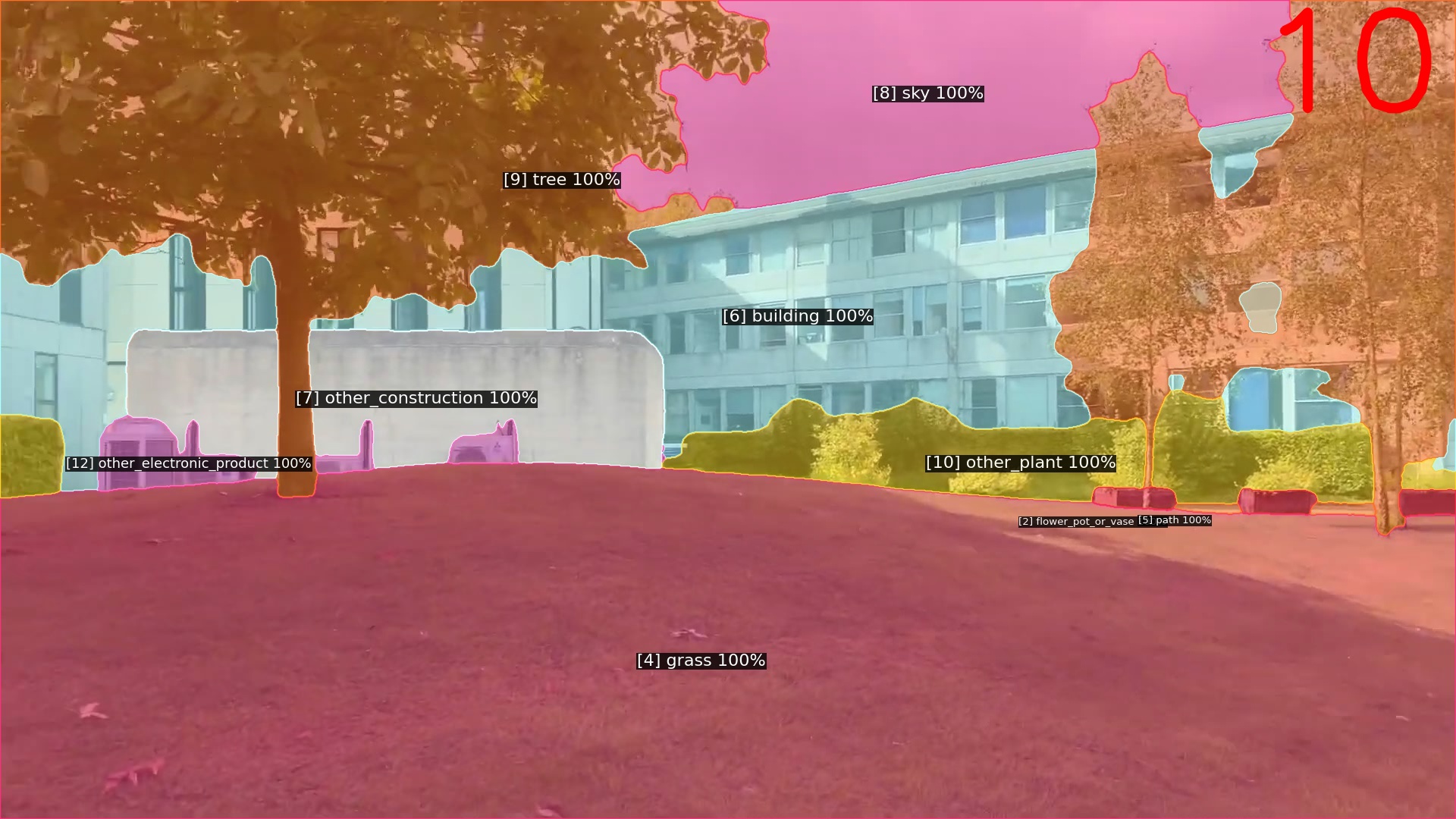}
\includegraphics[width=0.163\linewidth]{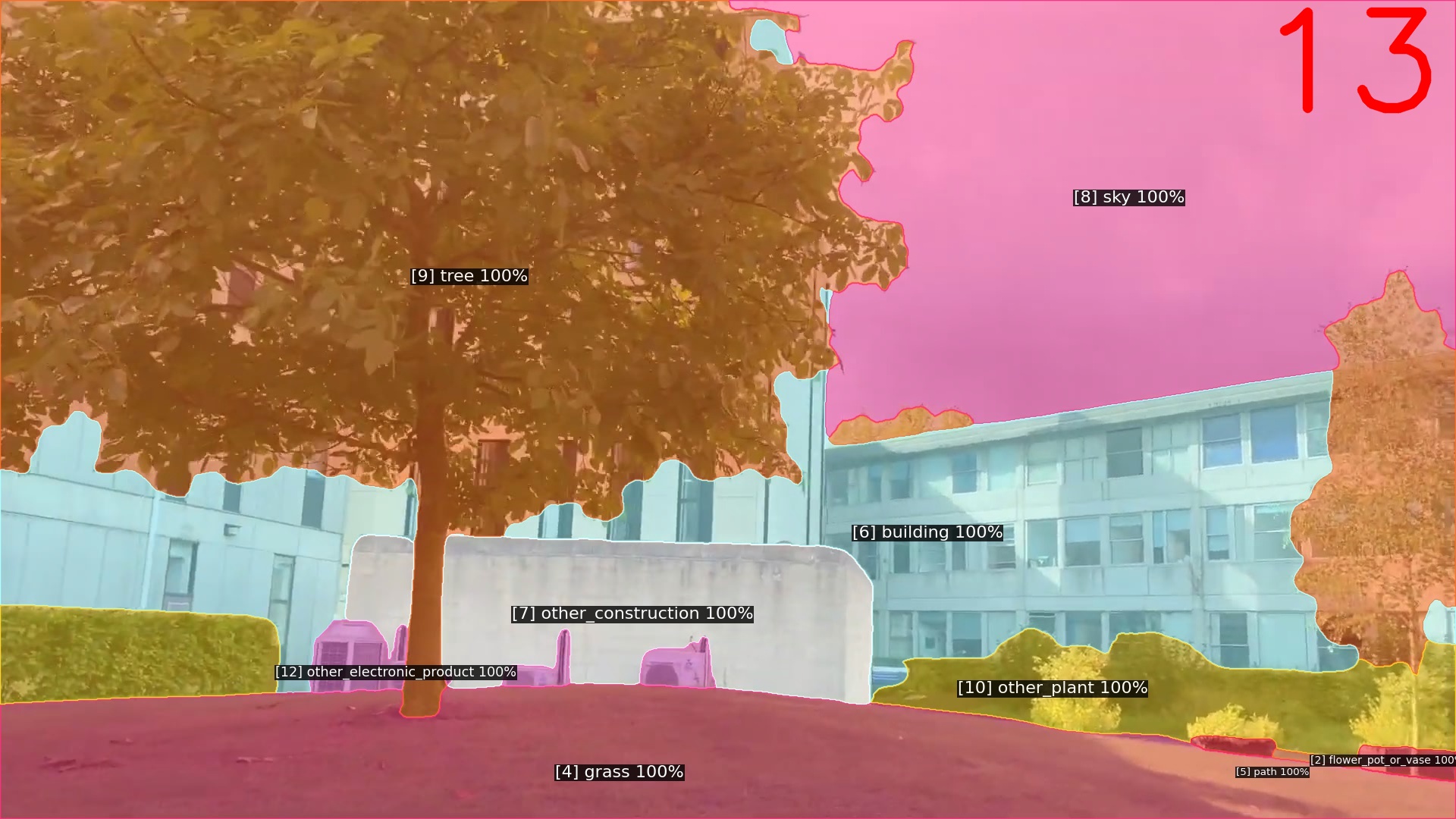}
\end{minipage}\hfill

\begin{minipage}[c]{1.0\linewidth}
\includegraphics[width=0.163\linewidth]{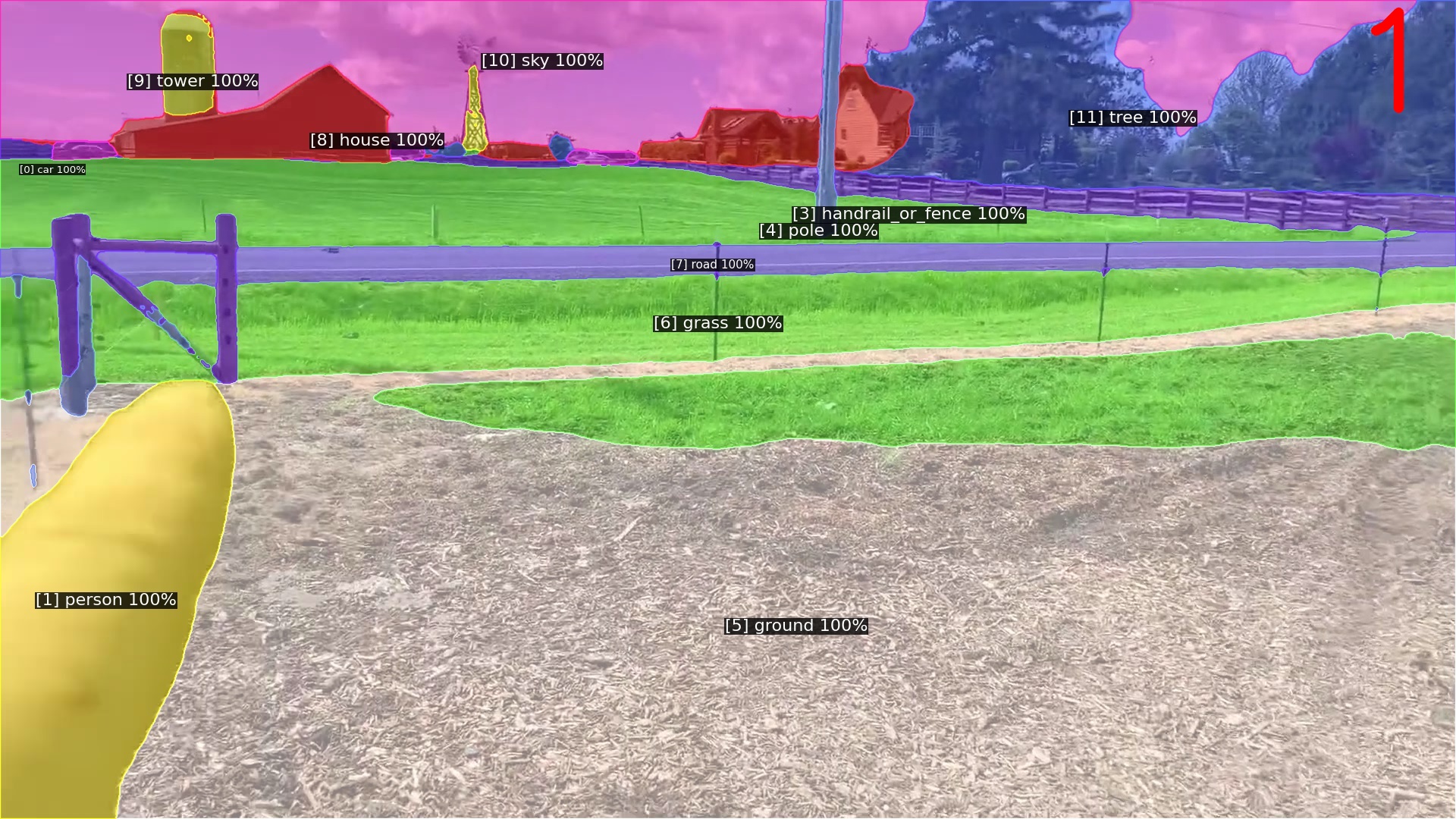}
\includegraphics[width=0.163\linewidth]{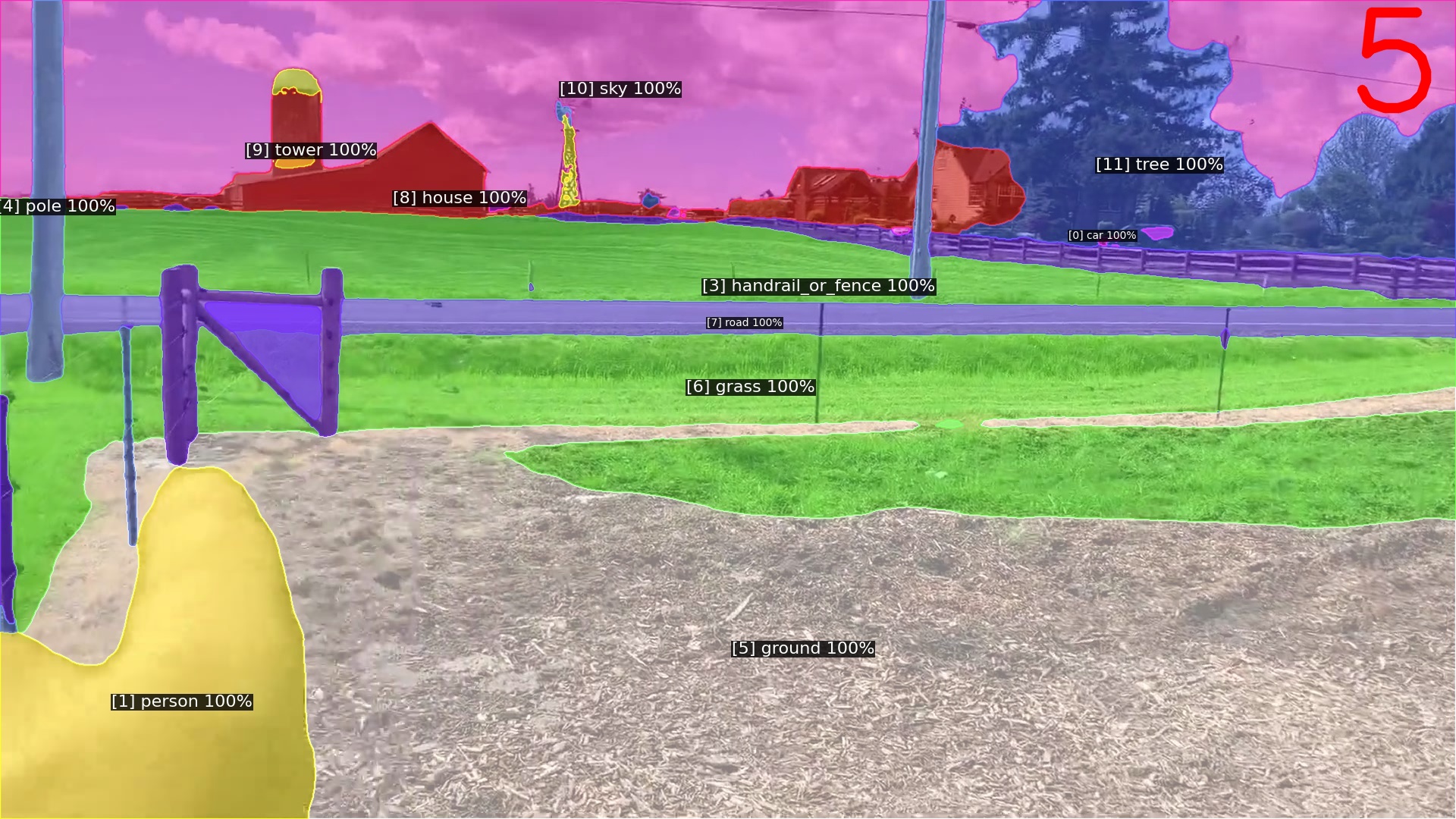}
\includegraphics[width=0.163\linewidth]{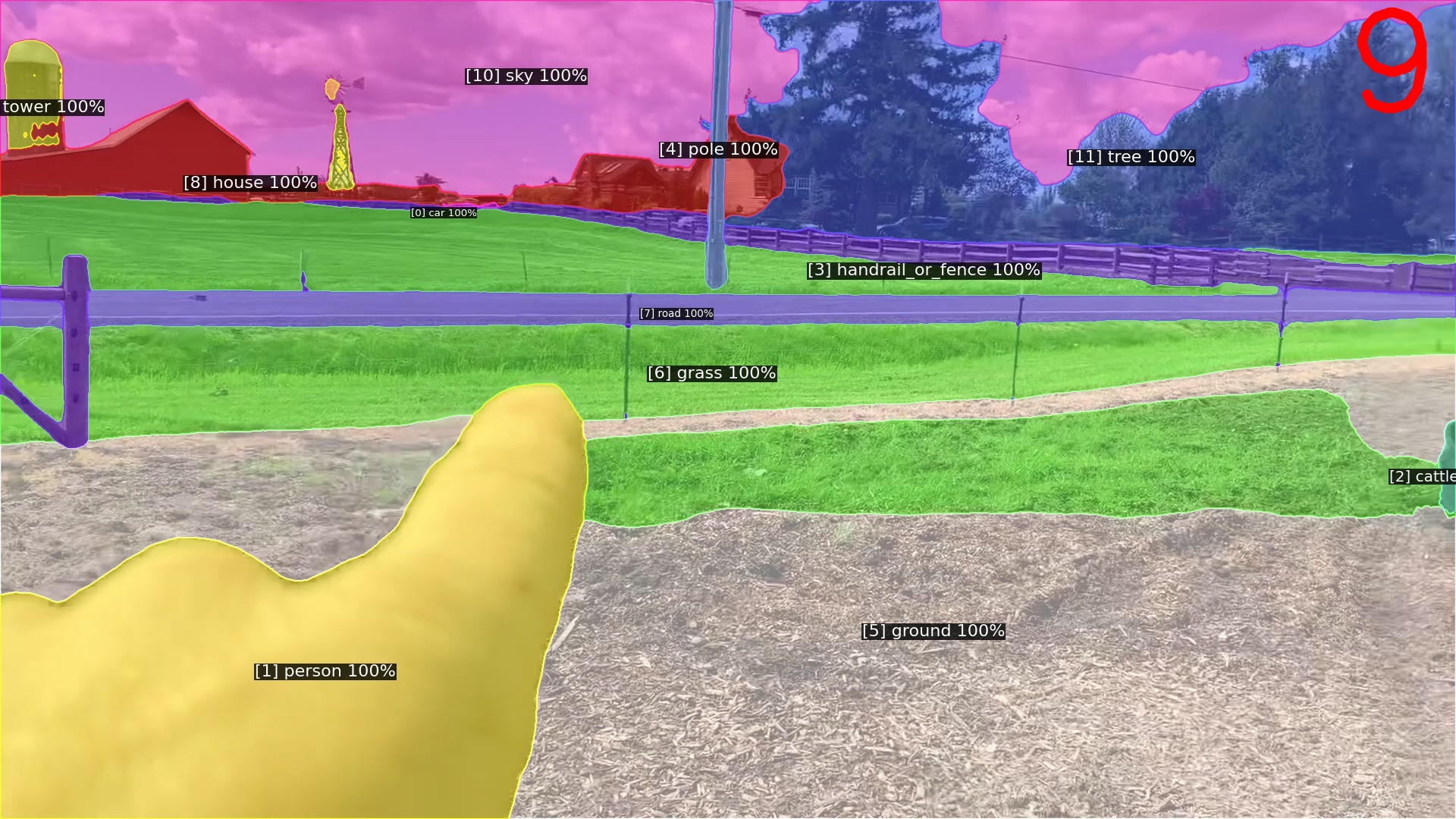}
\includegraphics[width=0.163\linewidth]{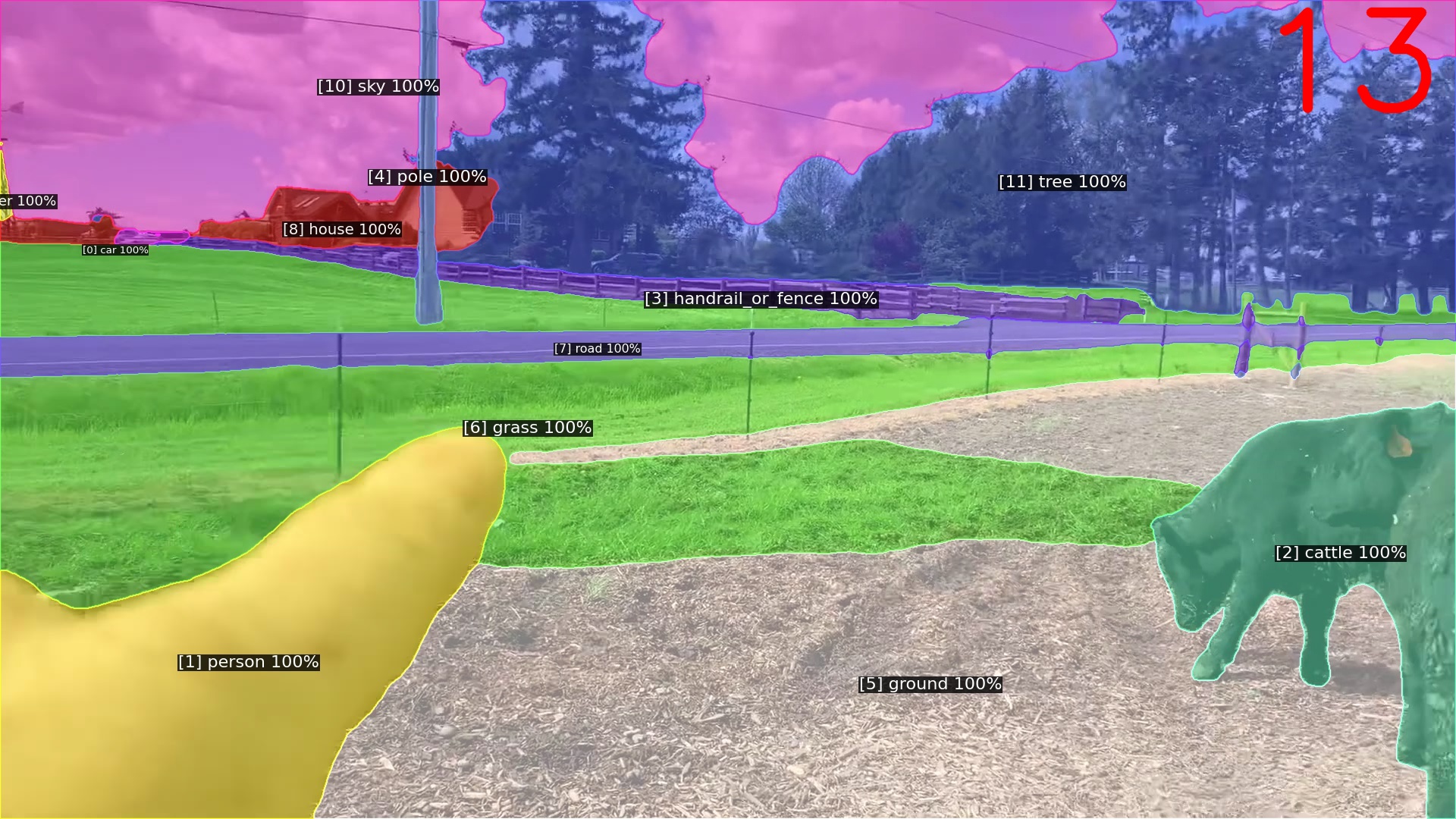}
\includegraphics[width=0.163\linewidth]{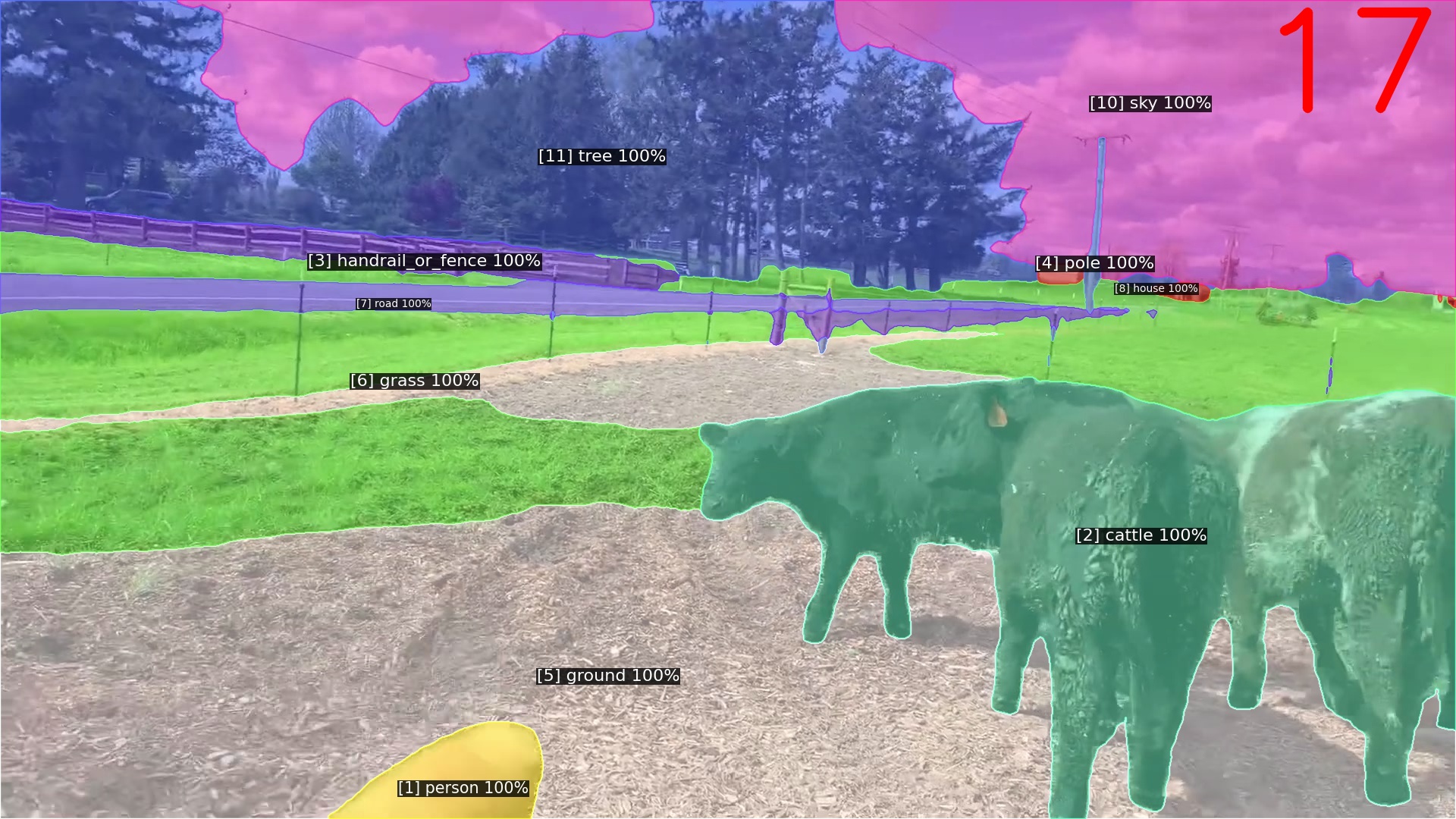}
\includegraphics[width=0.163\linewidth]{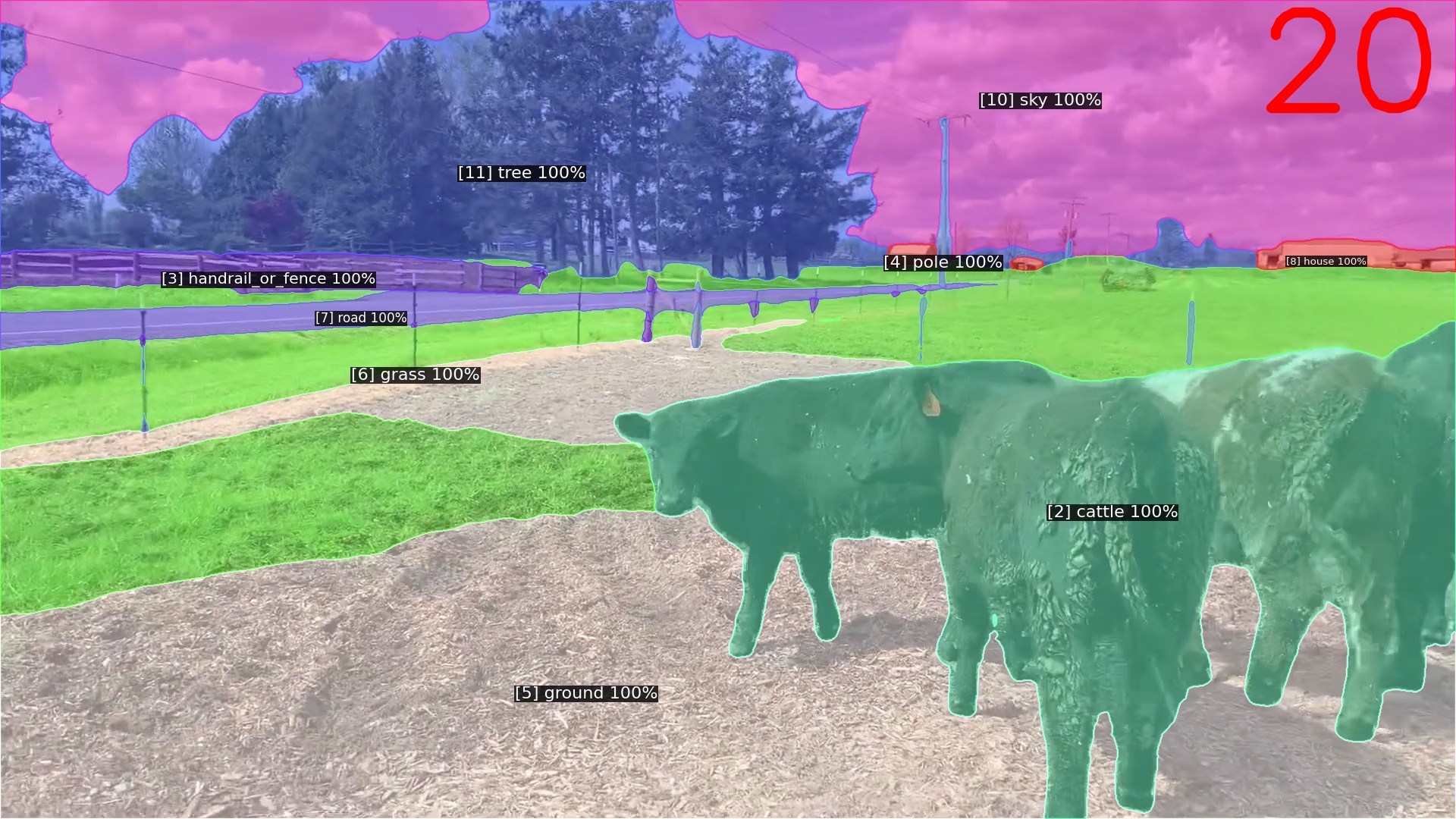}
\end{minipage}\hfill

    \caption{Qualitative results of DVIS++. The top, middle, and bottom rows show VIS, VPS, and VSS results, respectively. The red numbers in the top right corner of the images indicate the frame index within the video.}\vspace{-2mm}
    \label{fig:demos}
\end{figure*}

\noindent\textbf{Temporal Refiner.} The temporal refiner is designed to model the spatio-temporal representations of objects based on pre-aligned object representations from the tracker output. As shown in Table~\ref{tab:ablation} ($\mathcal{M}5$ \textbf{vs.} $\mathcal{M}4$), the temporal refiner brings a significant improvement of 4.0 AP, 4.1 AP$_{\rm m}$, and 4.4 AP$_{\rm h}$. This indicates that modeling spatio-temporal representations of objects is crucial for challenging scenarios in the representative OVIS dataset~\cite{qi2022occluded}.

\begin{figure*}[t!]
    \centering
\begin{minipage}[c]{0.16\linewidth}
New class: carrot

Task: VIS
\end{minipage}\hfill
\begin{minipage}[c]{0.838\linewidth}
\includegraphics[width=0.195\linewidth]{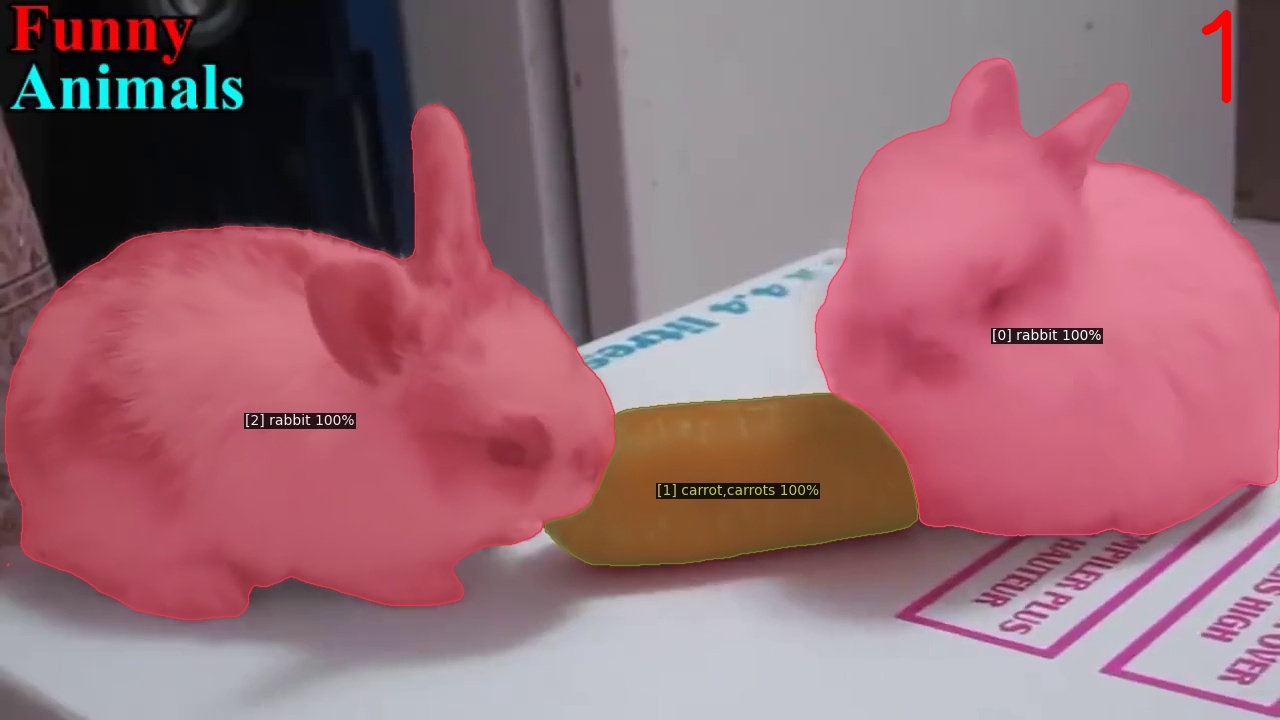}
\includegraphics[width=0.195\linewidth]{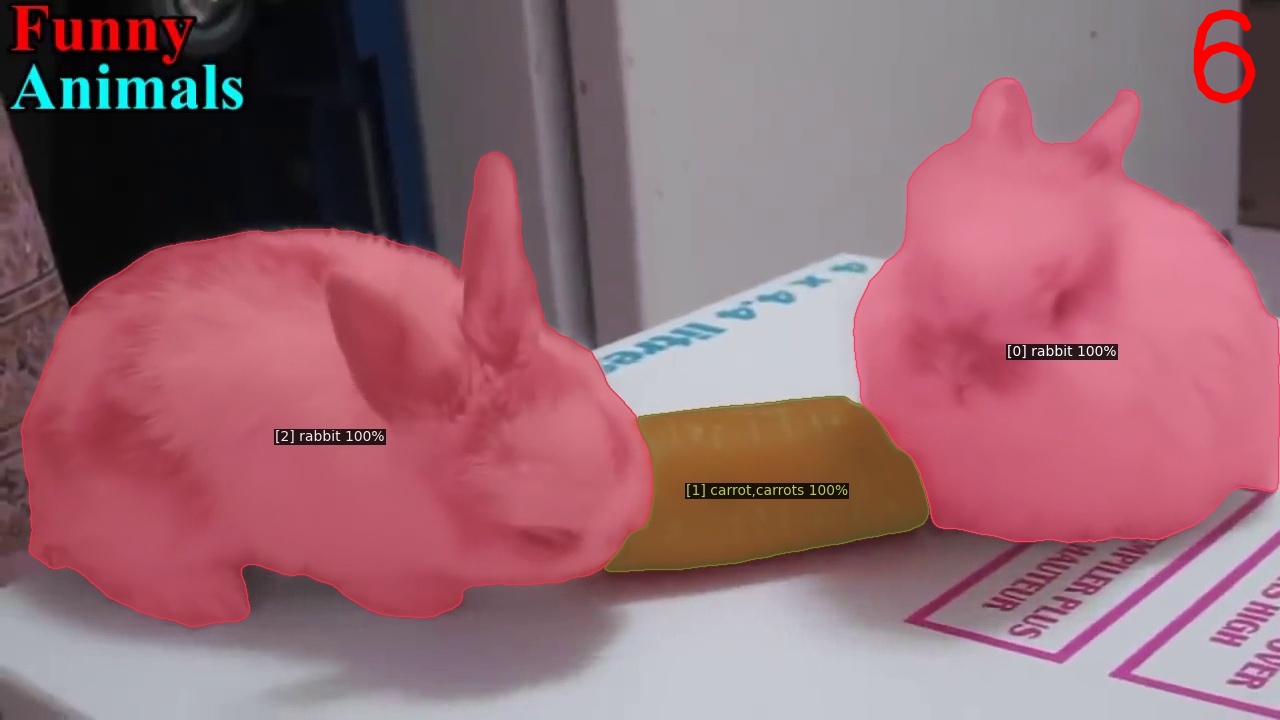}
\includegraphics[width=0.195\linewidth]{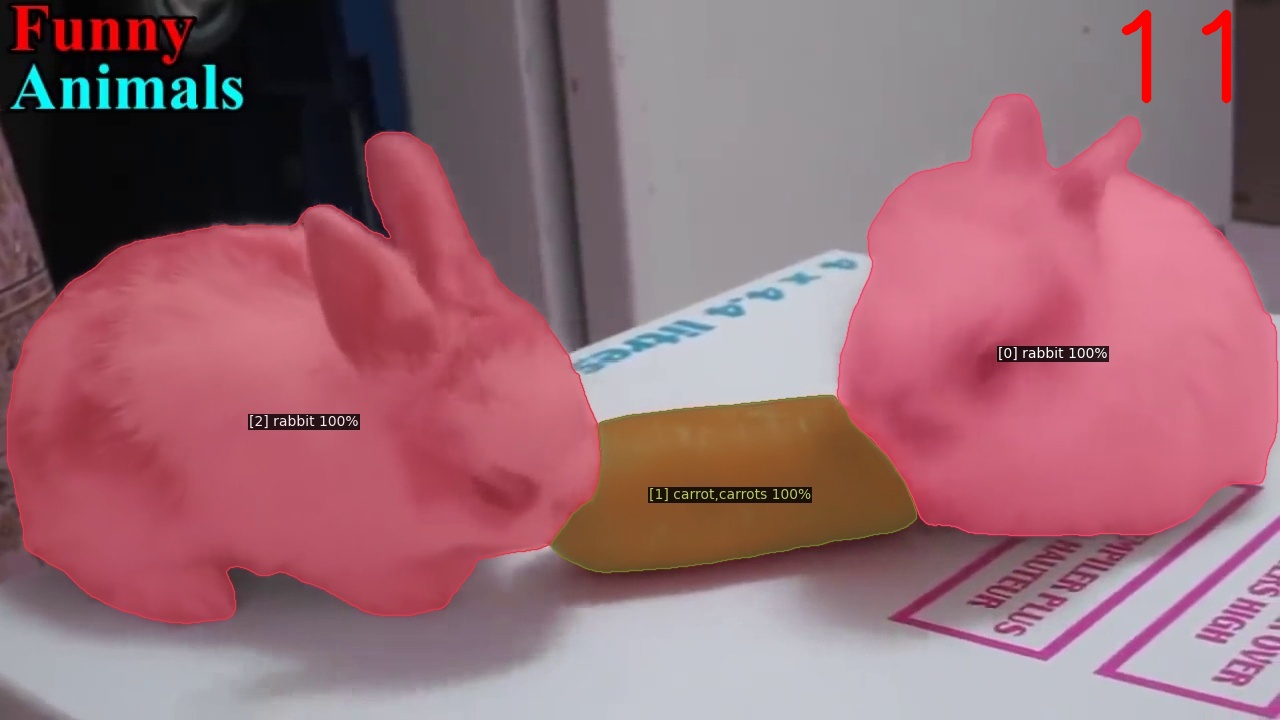}
\includegraphics[width=0.195\linewidth]{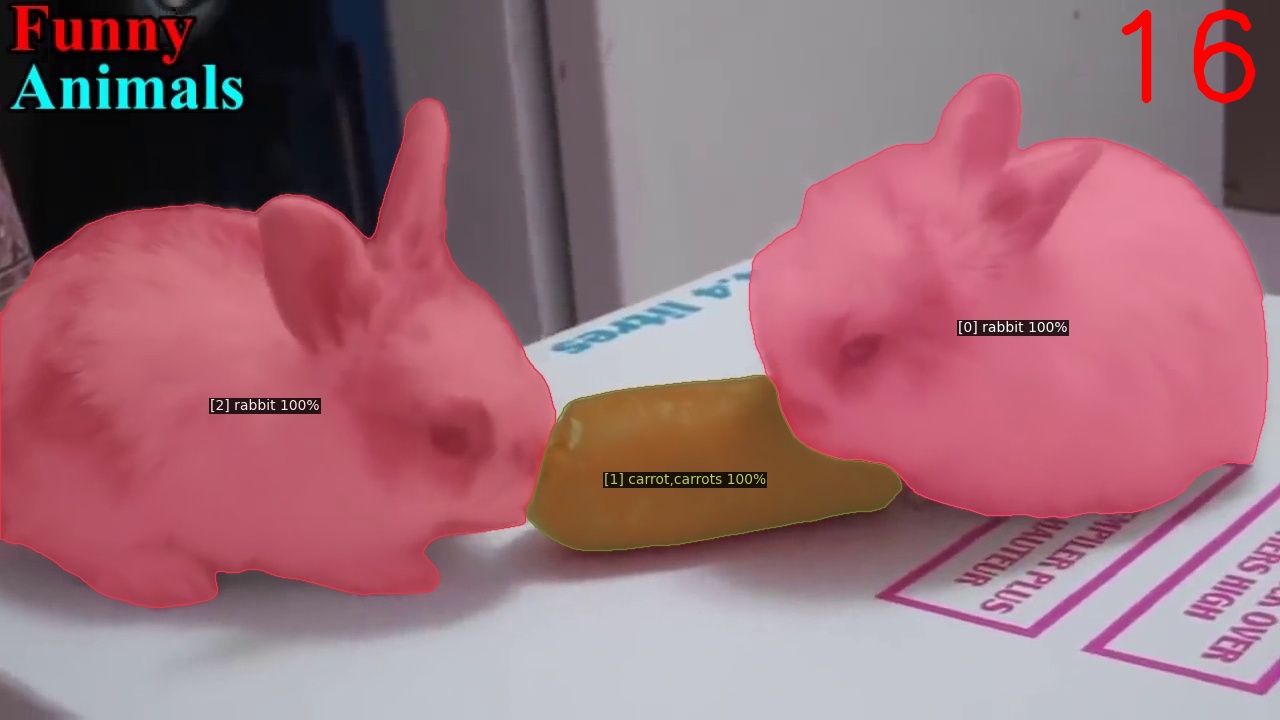}
\includegraphics[width=0.195\linewidth]{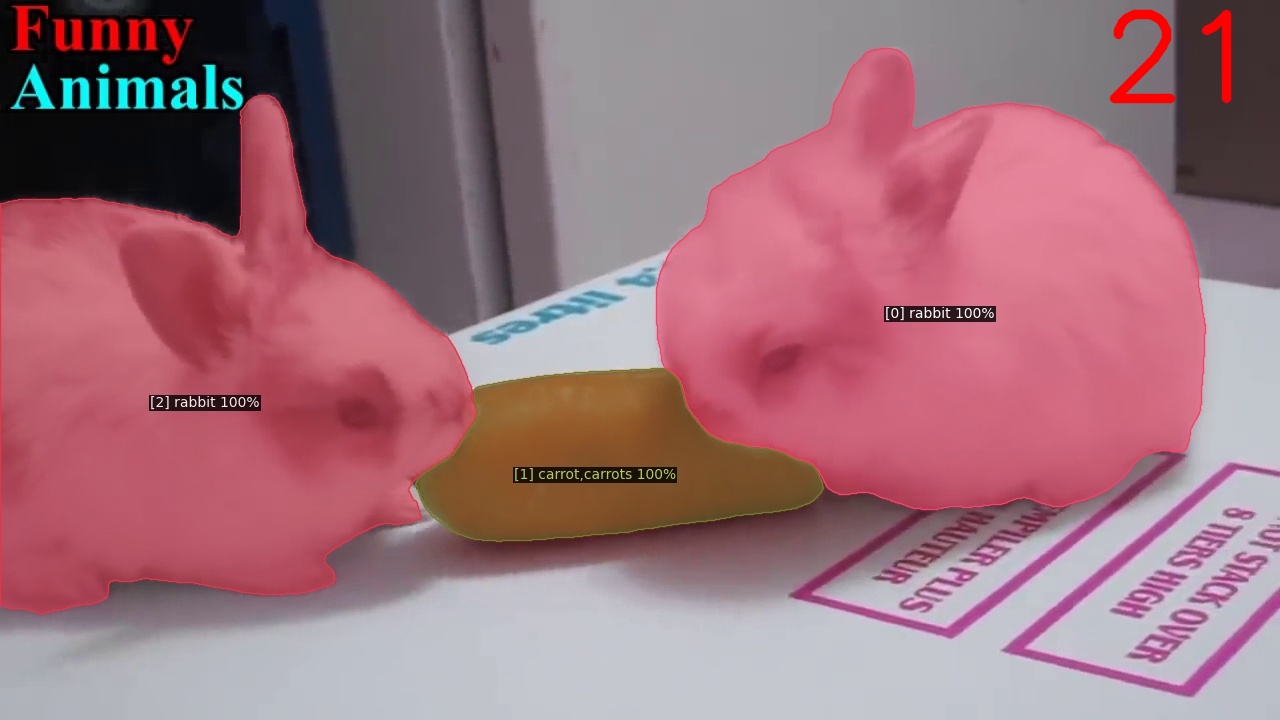}
\end{minipage}\hfill\vspace{1mm}

\begin{minipage}[c]{0.16\linewidth}
New class: hay

Task: VSS
\end{minipage}\hfill
\begin{minipage}[c]{0.838\linewidth}
\includegraphics[width=0.195\linewidth]{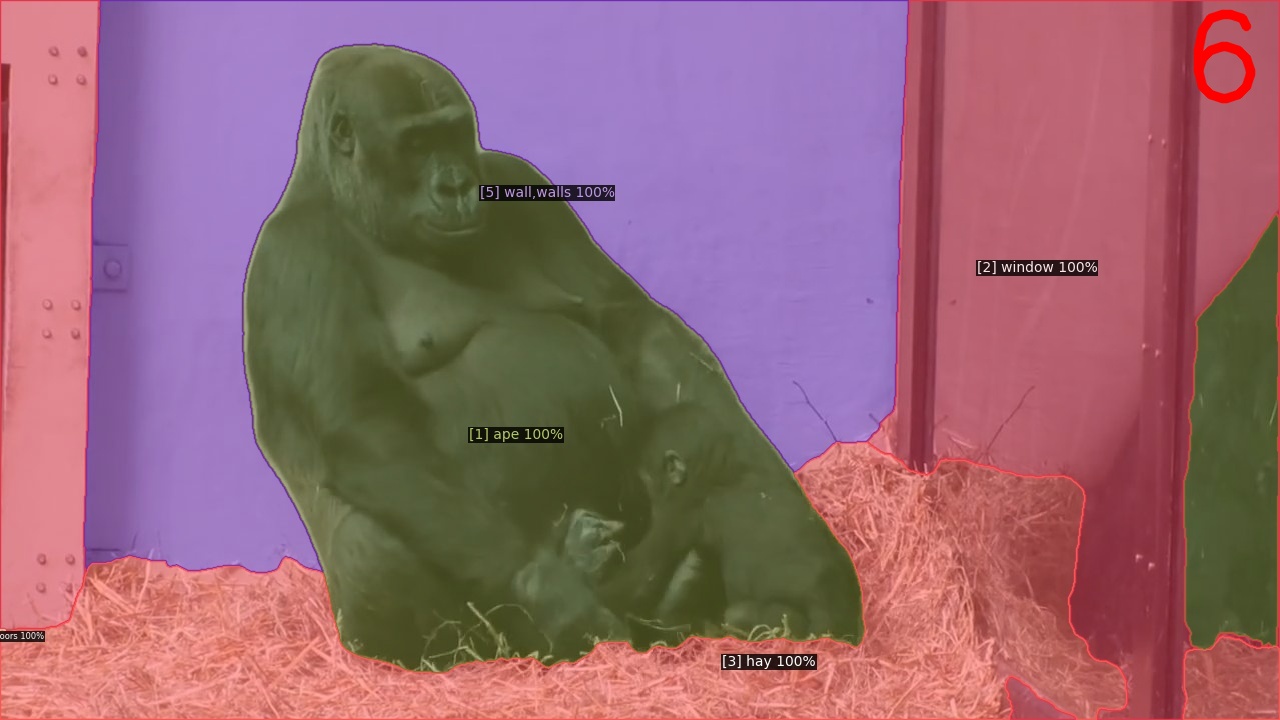}
\includegraphics[width=0.195\linewidth]{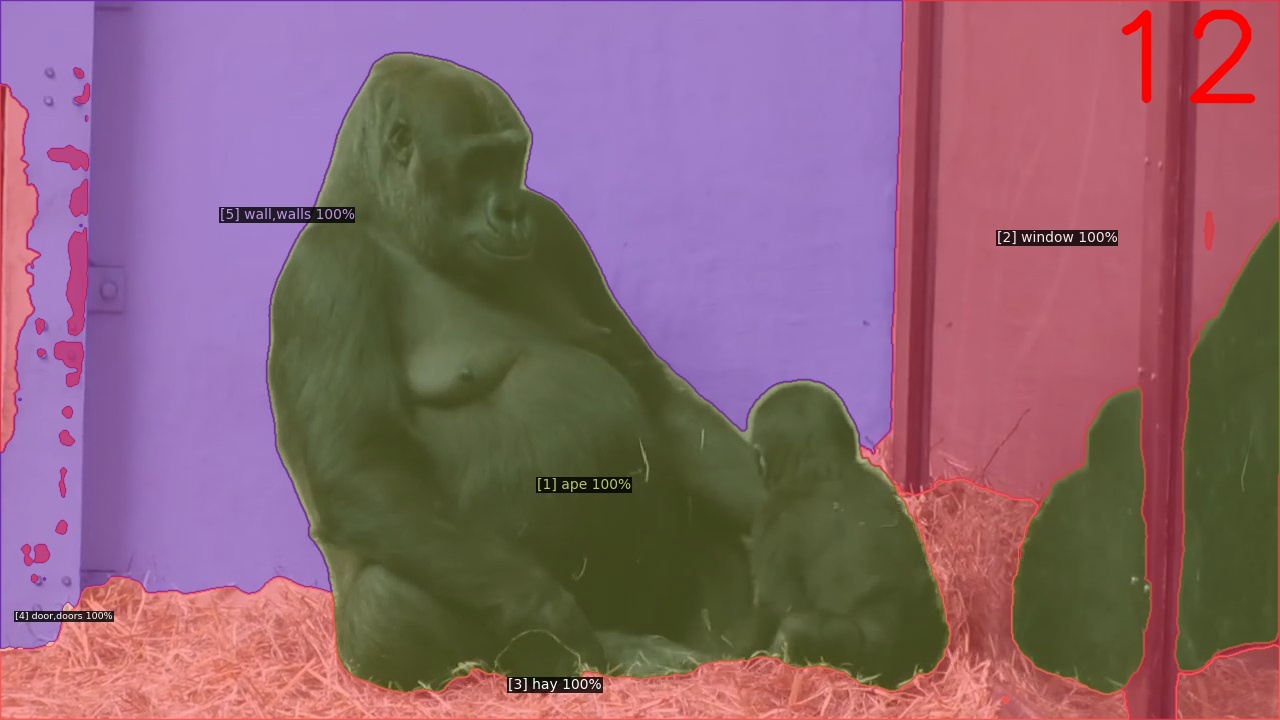}
\includegraphics[width=0.195\linewidth]{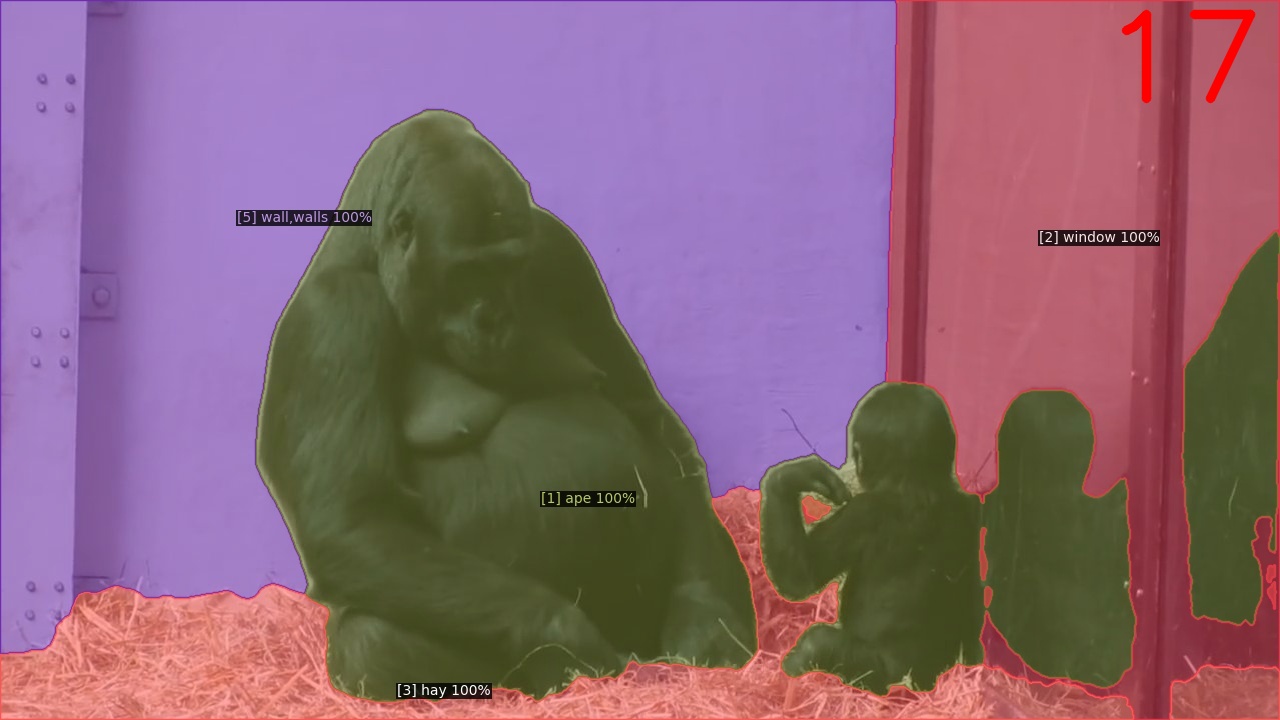}
\includegraphics[width=0.195\linewidth]{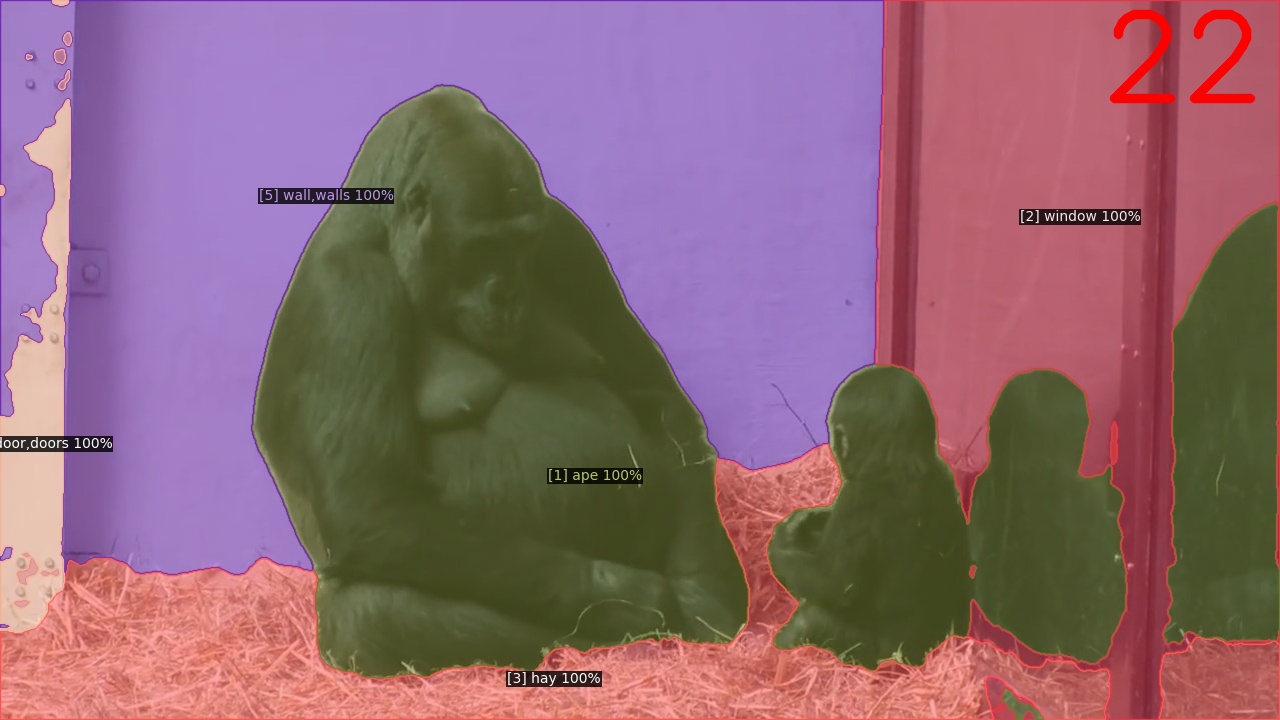}
\includegraphics[width=0.195\linewidth]{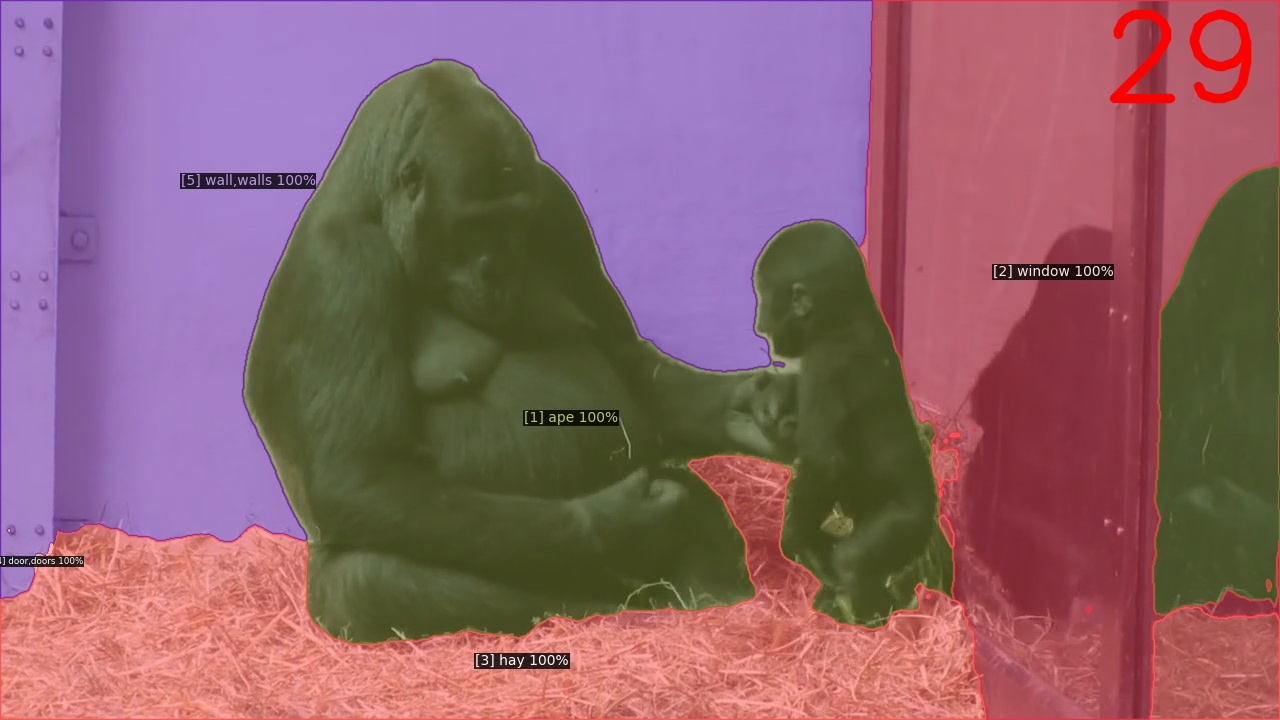}
\end{minipage}\hfill\vspace{1mm}

\begin{minipage}[c]{0.16\linewidth}
New class: lantern

Task: VSS
\end{minipage}\hfill
\begin{minipage}[c]{0.838\linewidth}
\includegraphics[width=0.195\linewidth]{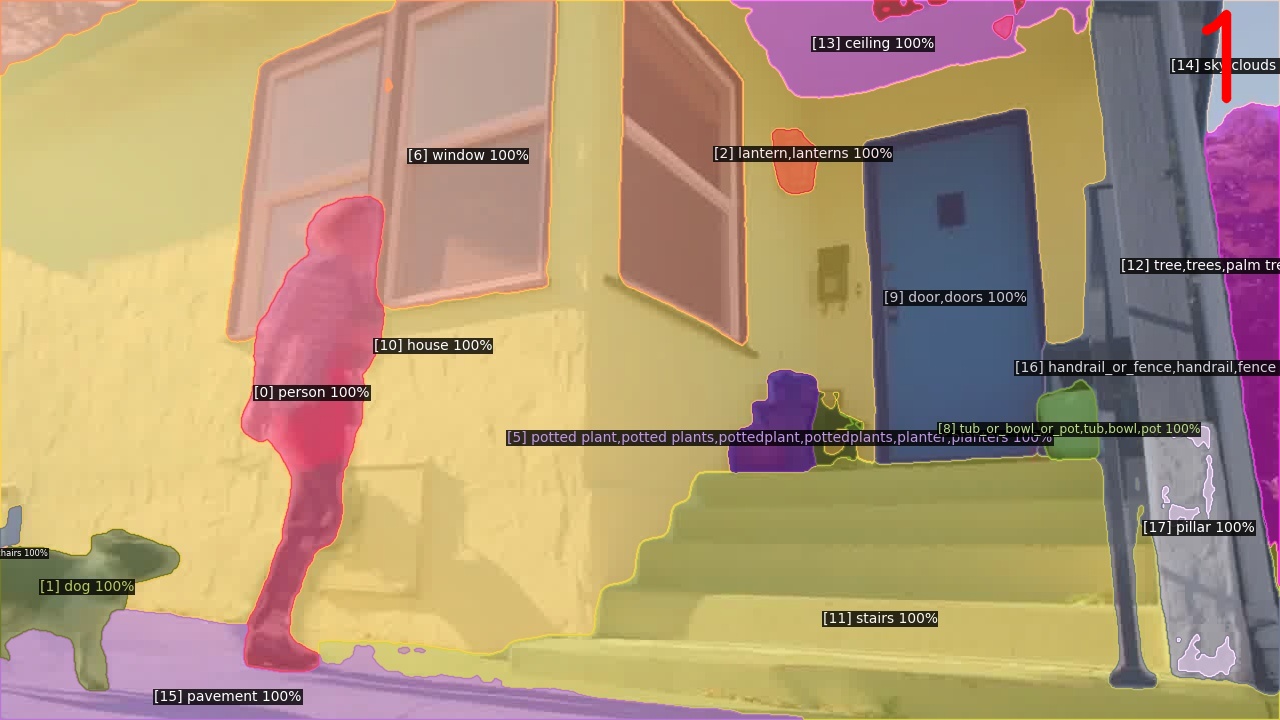}
\includegraphics[width=0.195\linewidth]{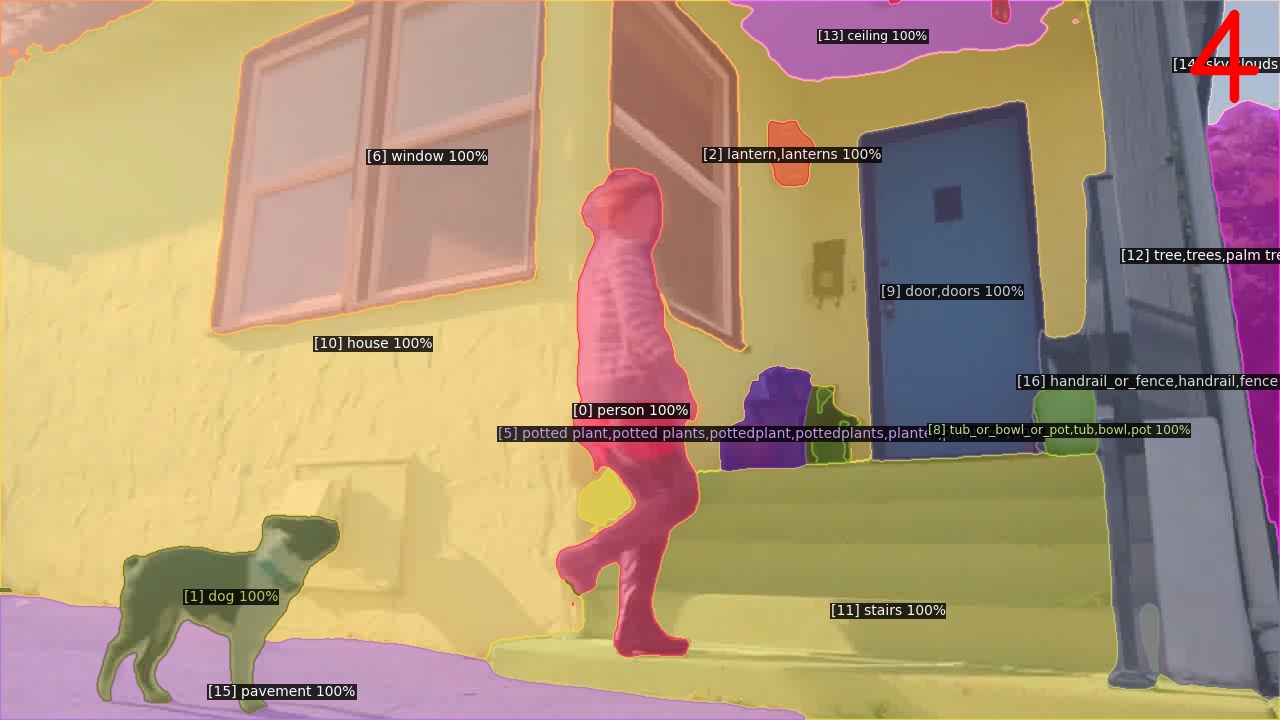}
\includegraphics[width=0.195\linewidth]{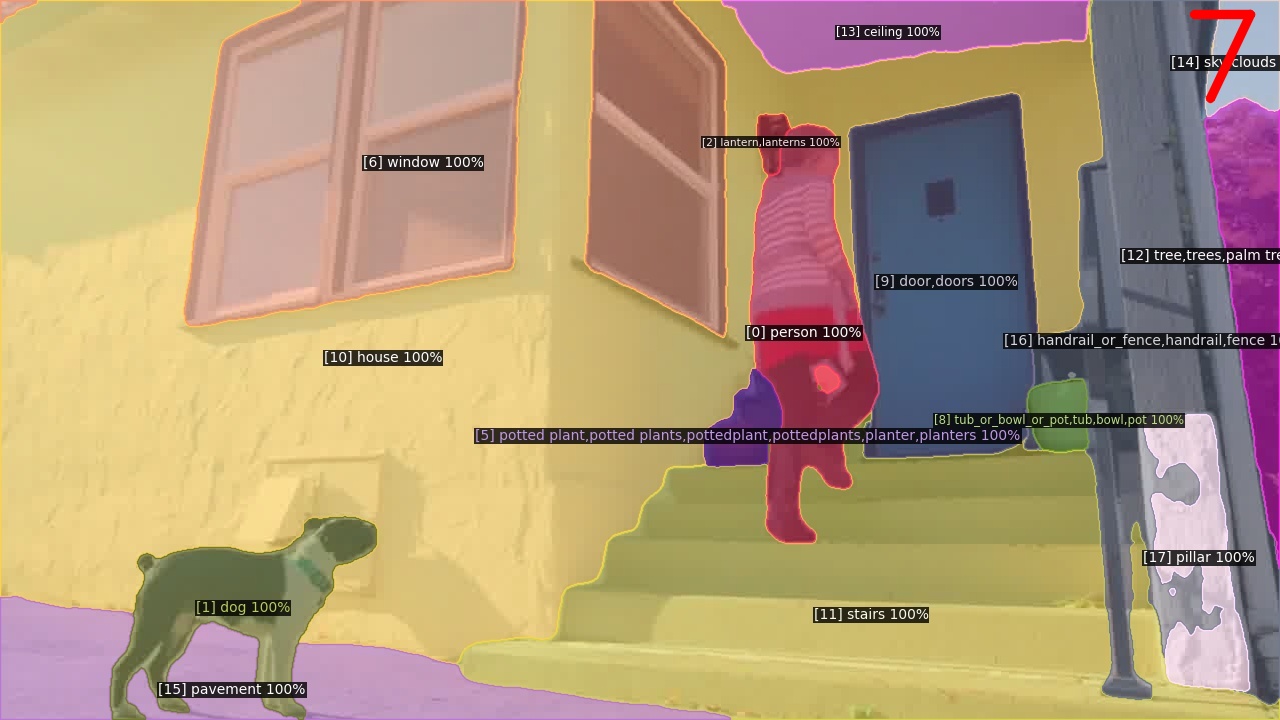}
\includegraphics[width=0.195\linewidth]{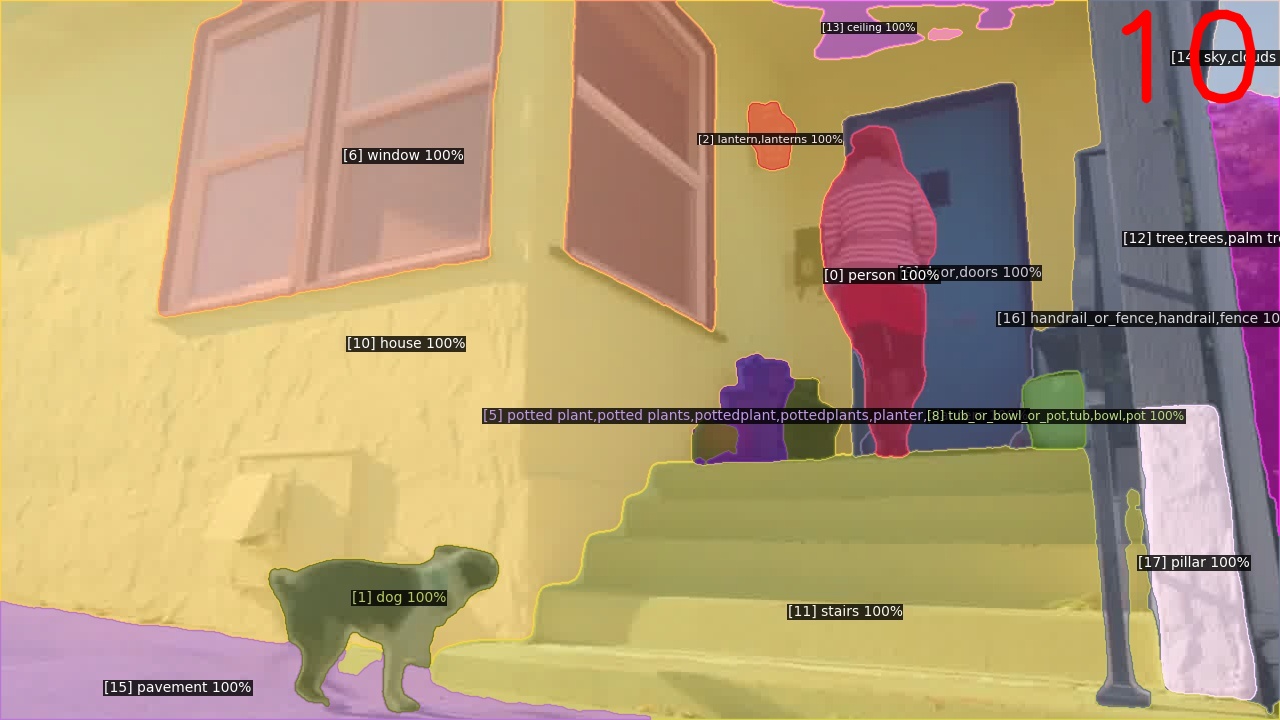}
\includegraphics[width=0.195\linewidth]{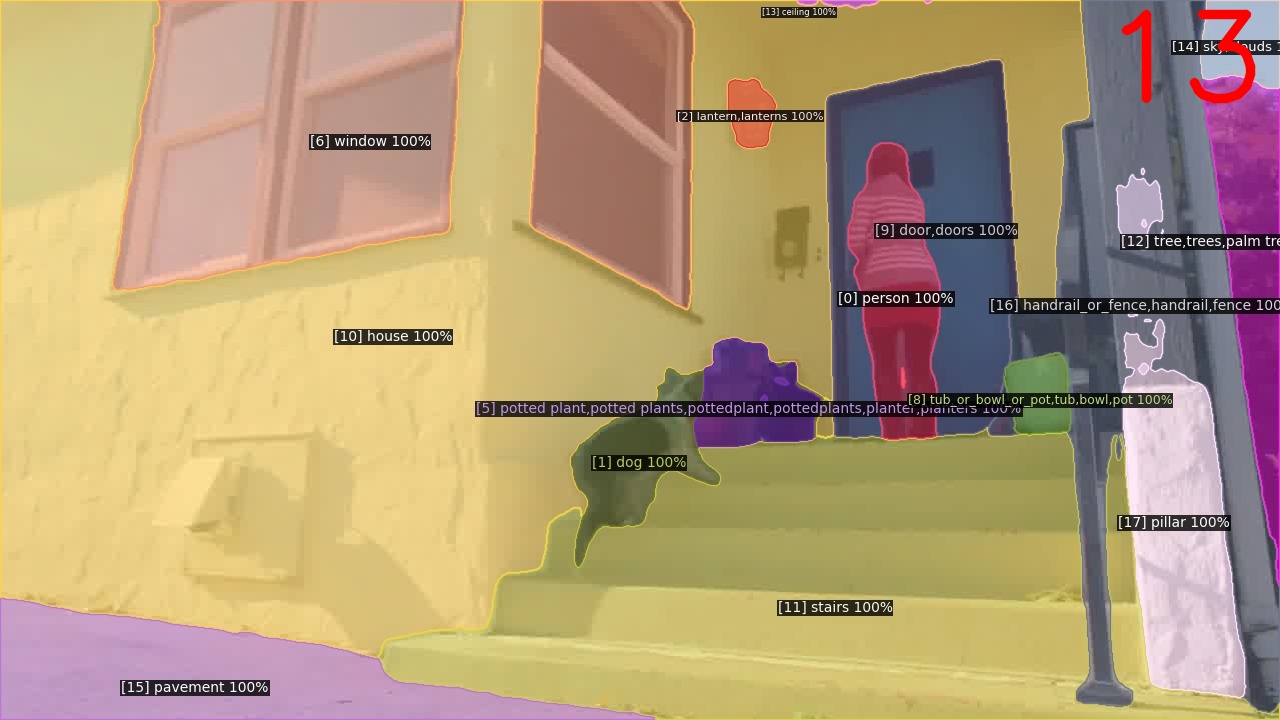}

\end{minipage}\hfill\vspace{1mm}

\begin{minipage}[c]{0.16\linewidth}
New class: kangaroo

Task: VPS
\end{minipage}\hfill
\begin{minipage}[c]{0.838\linewidth}
\includegraphics[width=0.195\linewidth]{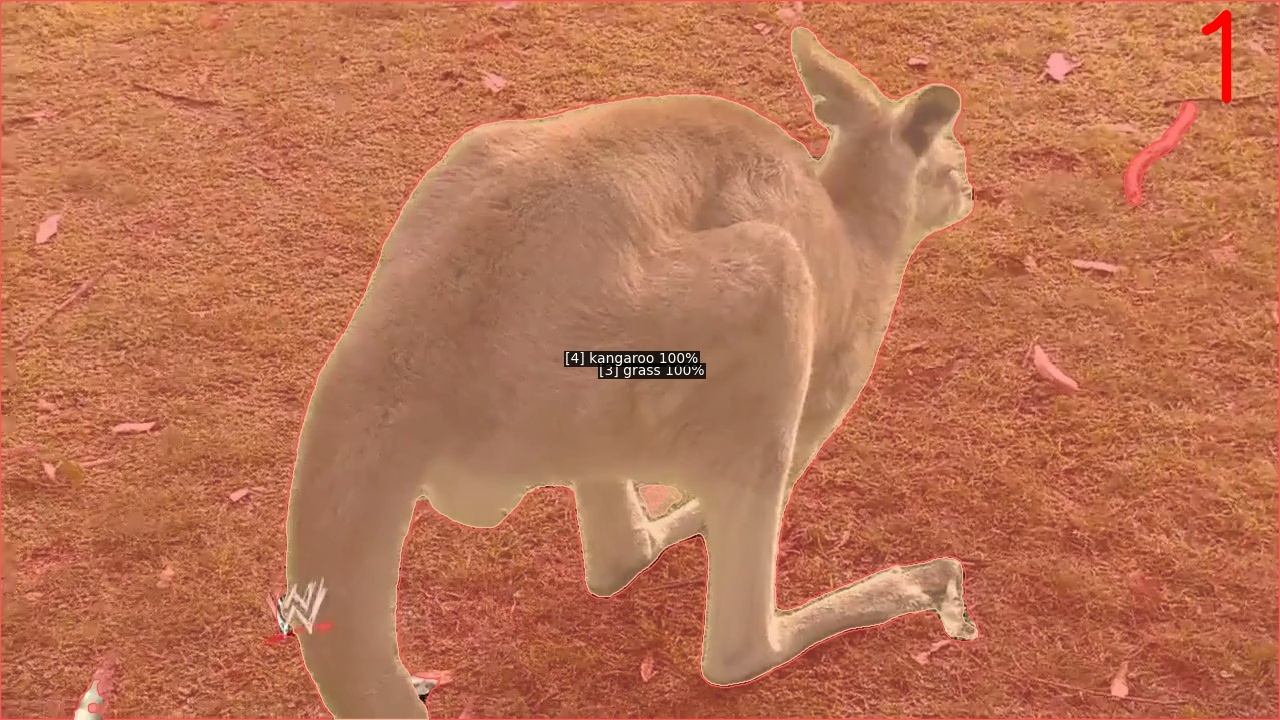}
\includegraphics[width=0.195\linewidth]{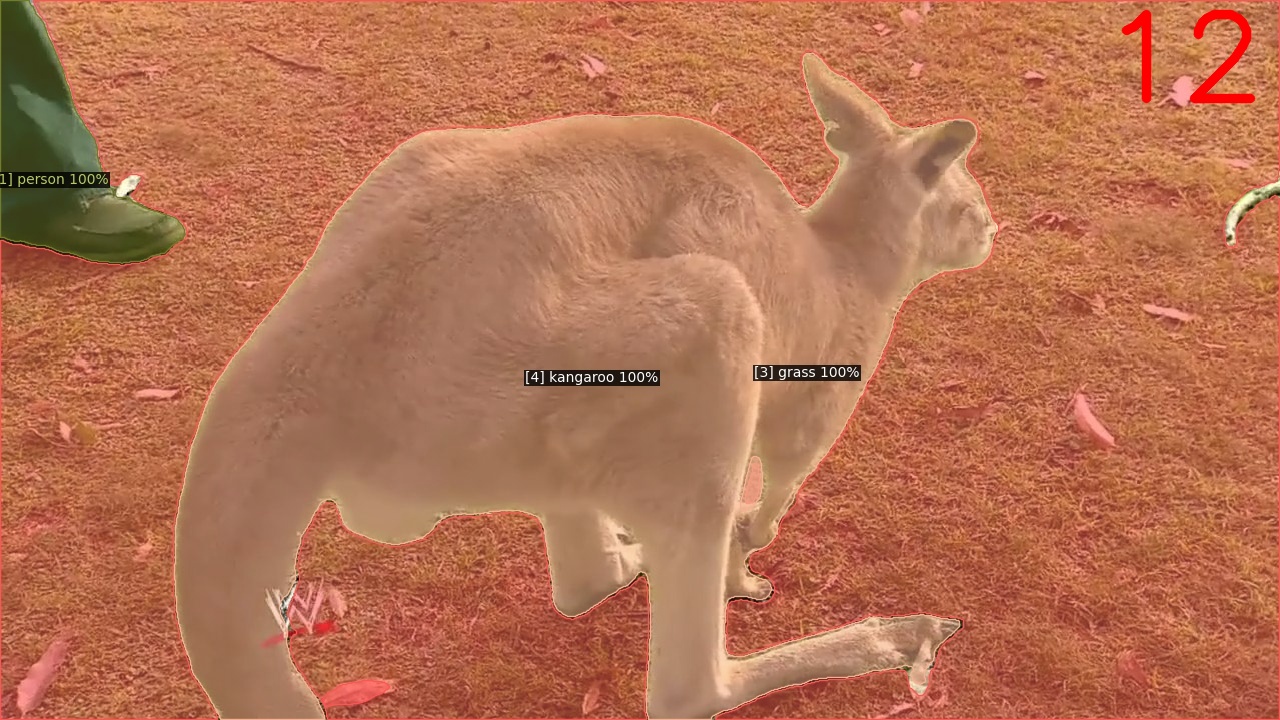}
\includegraphics[width=0.195\linewidth]{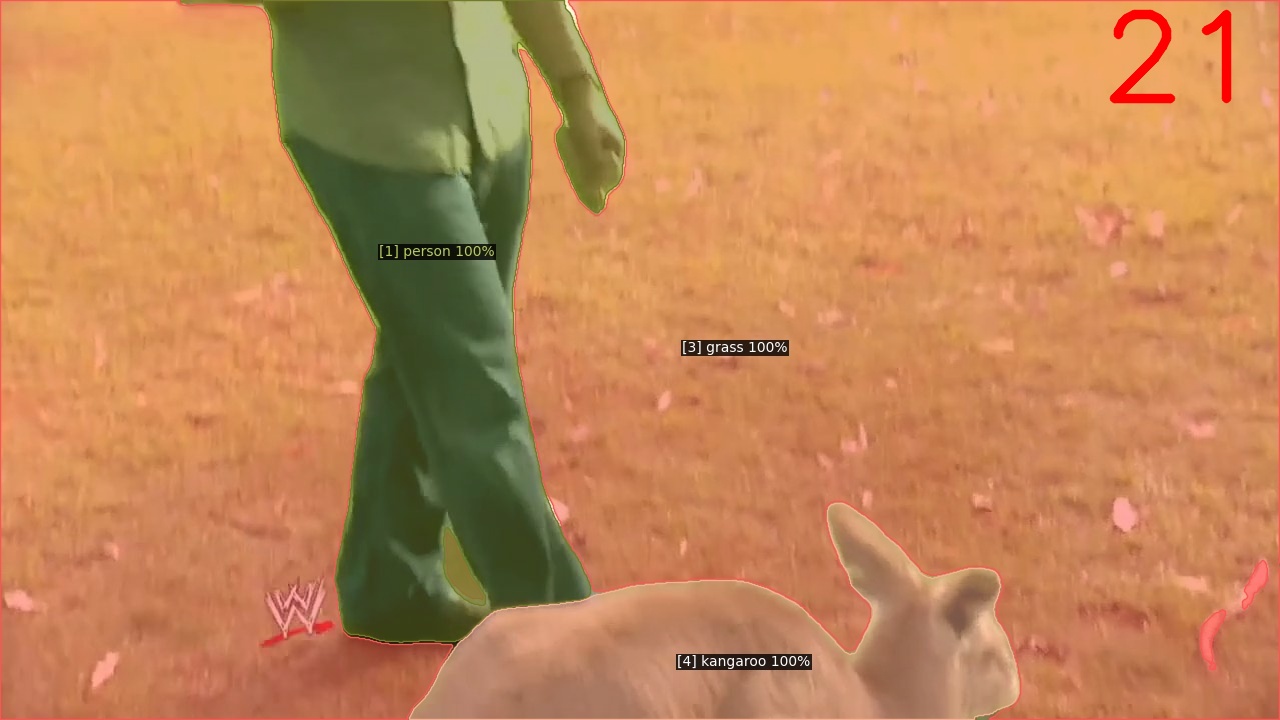}
\includegraphics[width=0.195\linewidth]{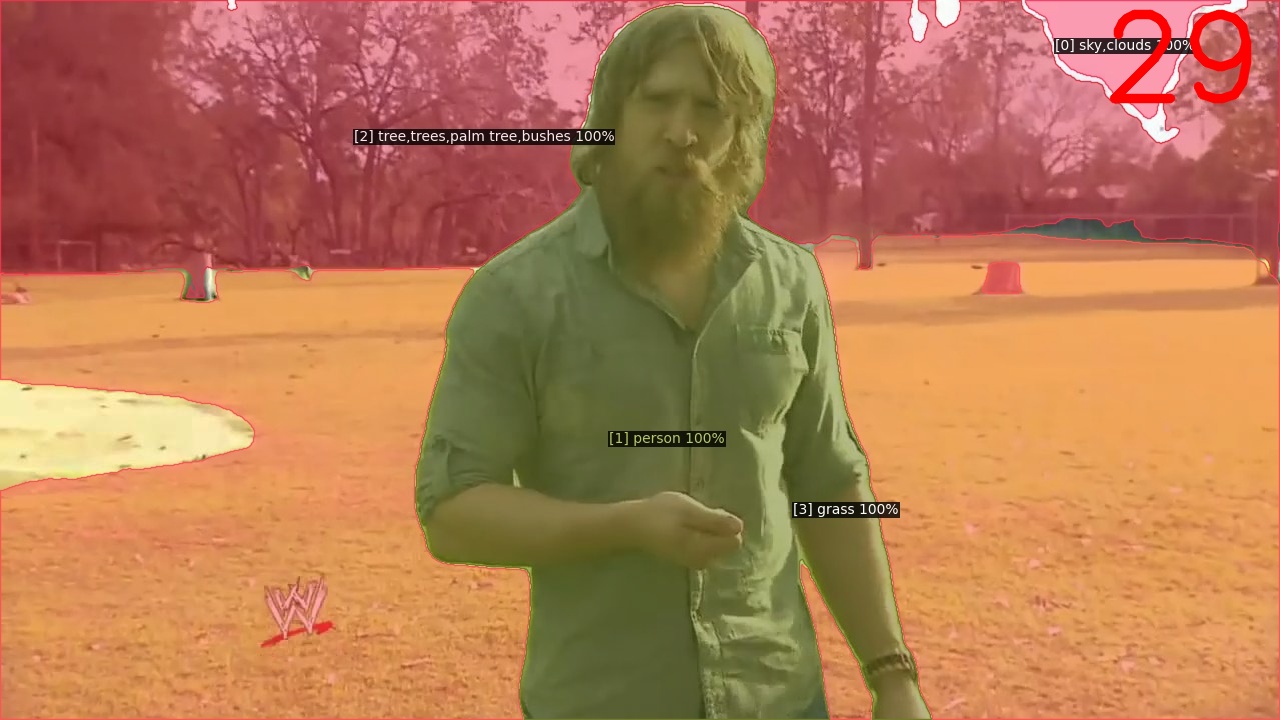}
\includegraphics[width=0.195\linewidth]{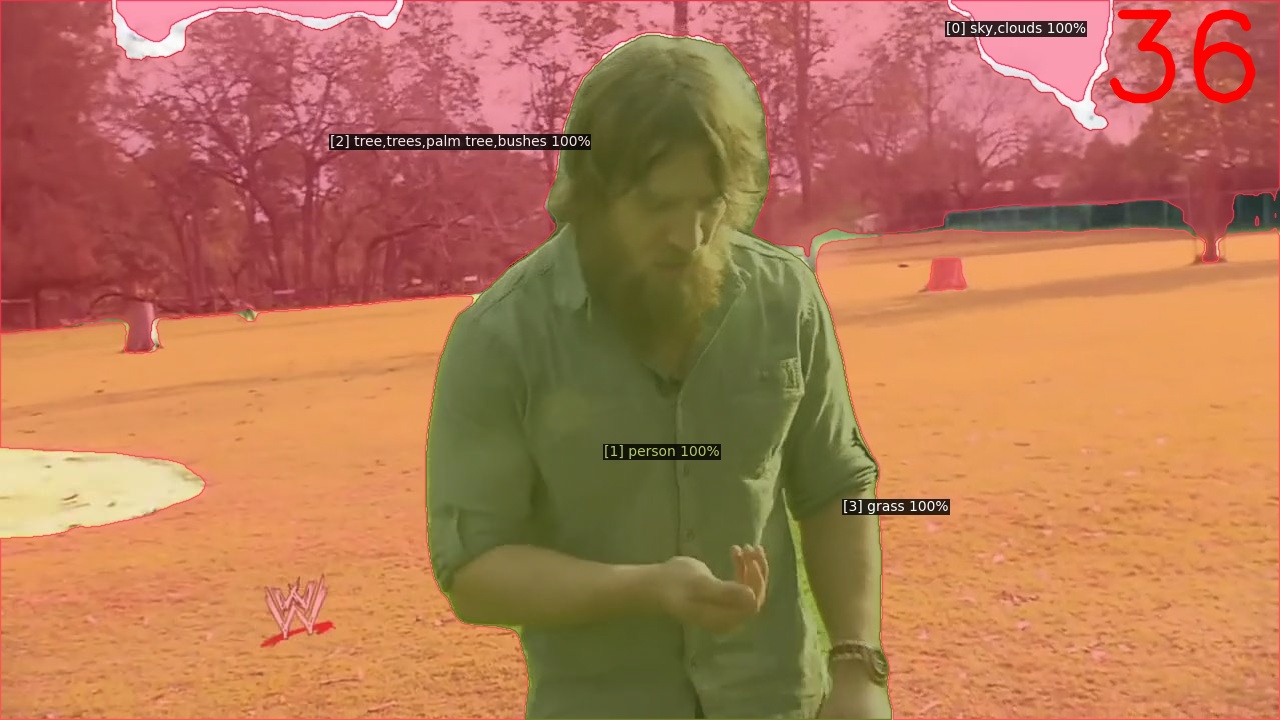}
\end{minipage}\hfill\vspace{1mm}

\begin{minipage}[c]{0.16\linewidth}
New classes: card, goose

Task: VPS
\end{minipage}\hfill
\begin{minipage}[c]{0.838\linewidth}
\includegraphics[width=0.195\linewidth]{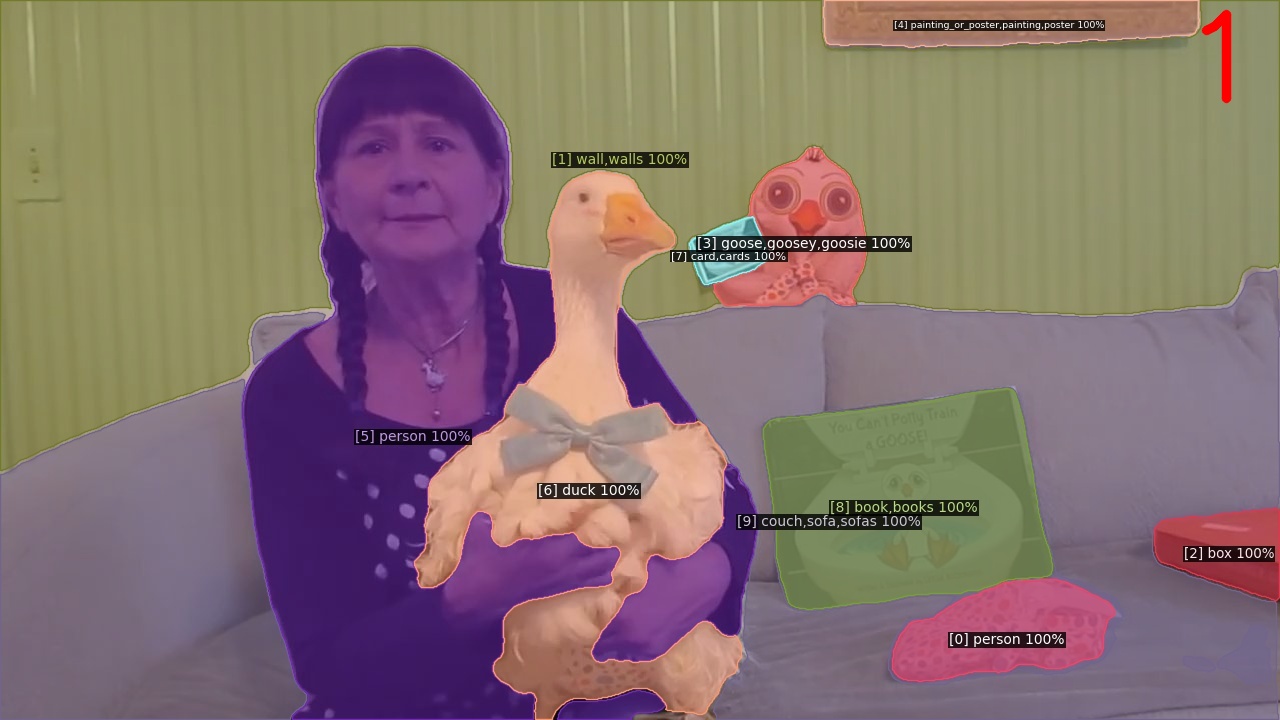}
\includegraphics[width=0.195\linewidth]{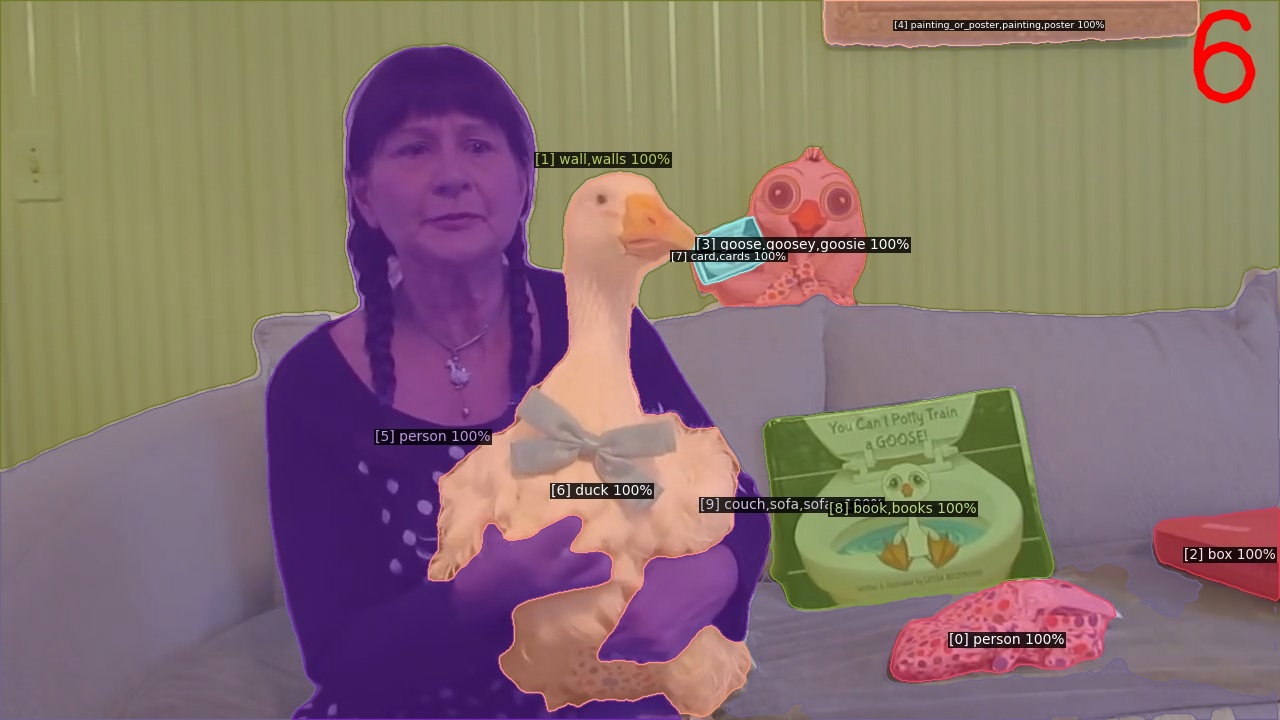}
\includegraphics[width=0.195\linewidth]{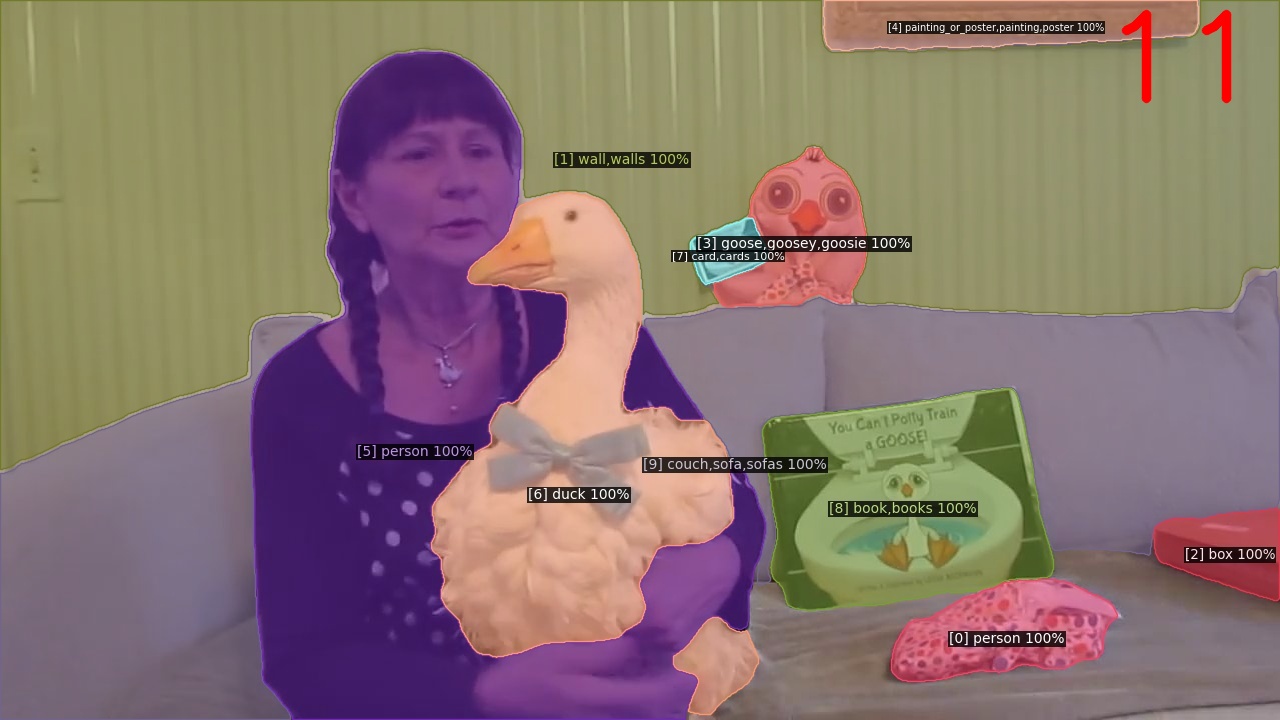}
\includegraphics[width=0.195\linewidth]{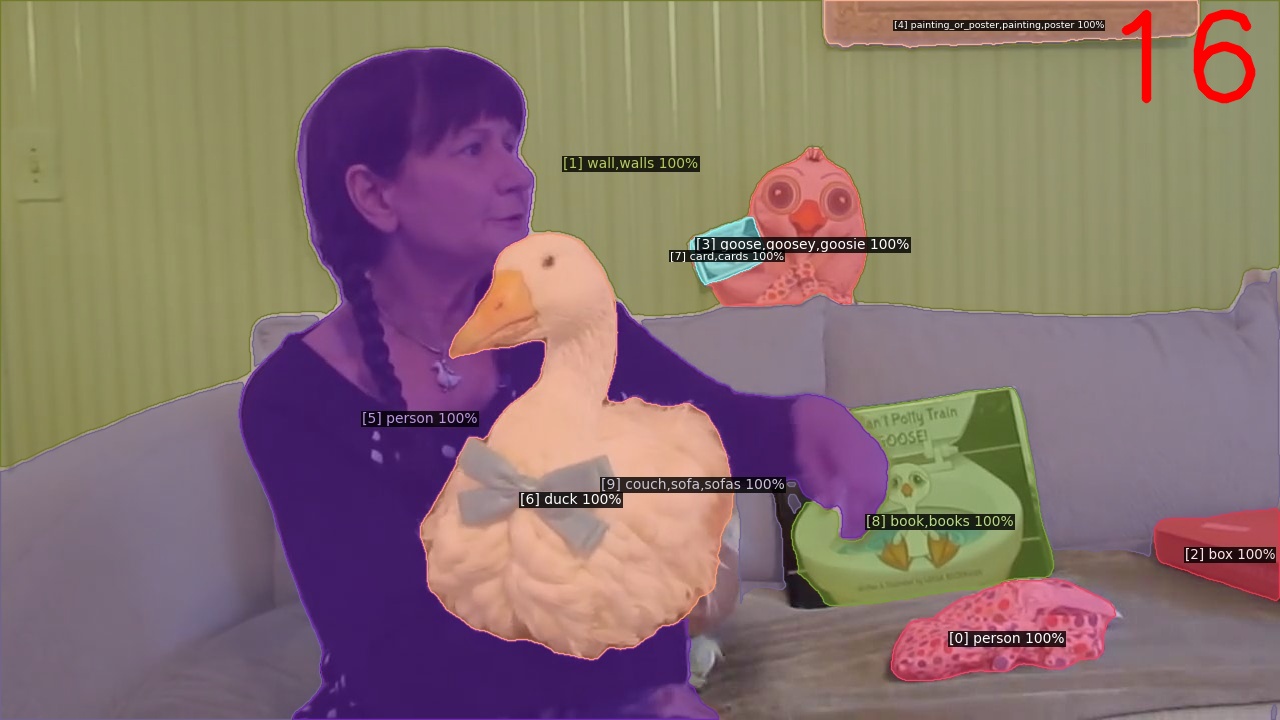}
\includegraphics[width=0.195\linewidth]{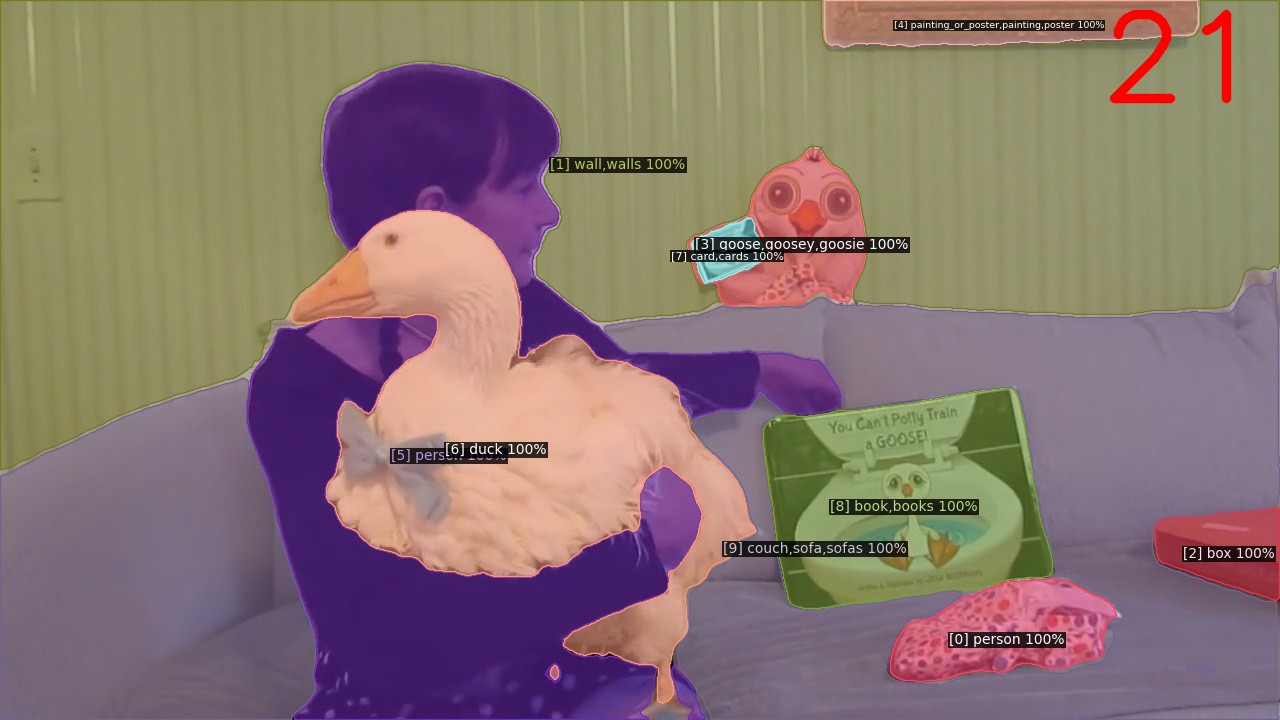}
\end{minipage}\hfill
    \caption{Qualitative open-vocabulary video segmentation results of OV-DVIS++. The model is jointly trained on the COCO dataset and video datasets. During inference, we add additional new class vocabularies that have never appeared in the above datasets, such as ``carrot" and ``hay".}
    \label{fig:demos_ov}
\end{figure*}

We also conduct ablation experiments on the key components of the temporal refiner, and the results are presented in Table~\ref{tab:temporal refiner}. The temporal refiner utilizes self-attention to model long-term temporal relationships. Removing long-term attention leads to a performance degradation of 3.2 AP. To model short-term temporal relationships, the temporal refiner employs 1D convolution. Although long-term attention theoretically covers this aspect, removing 1D convolution result in a performance degradation of 0.2 AP, indicating that 1D convolution is more effective in capturing short-term relationships. Furthermore, the temporal refiner leverages cross-attention to access image information provided by the segmenter, enabling the correction of potential errors such as segmentation and tracking errors. Removing cross-attention leads to a performance degradation of 0.8 AP, highlighting the reliance of the temporal refiner on the original information provided by the segmenter for correcting errors. The network heavily depends on self-attention to suppress confusion between different objects, and removing it results in a performance drop of 1.7 AP.

\noindent\textbf{Denosing Training Strategy.} The denoising training strategy is proposed to enhance the tracking capabilities of the referring tracker. As demonstrated in Table~\ref{tab:ablation}, the implementation of this strategy leads to notable performance improvements across various metrics: 3.7 AP, 2.7 AP$_{\rm l}$, 4.3 AP$_{\rm m}$, and 4.1 AP$_{\rm h}$ ($\mathcal{M}3$ \textit{vs}. $\mathcal{M}2$). Particularly, the denoising training strategy significantly enhances the tracker's performance in challenging scenarios, including objects with moderate to heavy occlusion. As a result, the strategy yields more substantial performance improvements for heavily occluded objects compared to slightly occluded objects (+4.1 AP$_{\rm h}$ \textit{vs.} +2.7 AP$_{\rm l}$).

\begin{figure*}[t!]
    \centering
\begin{minipage}[c]{1.00\linewidth}
\includegraphics[width=0.163\linewidth]{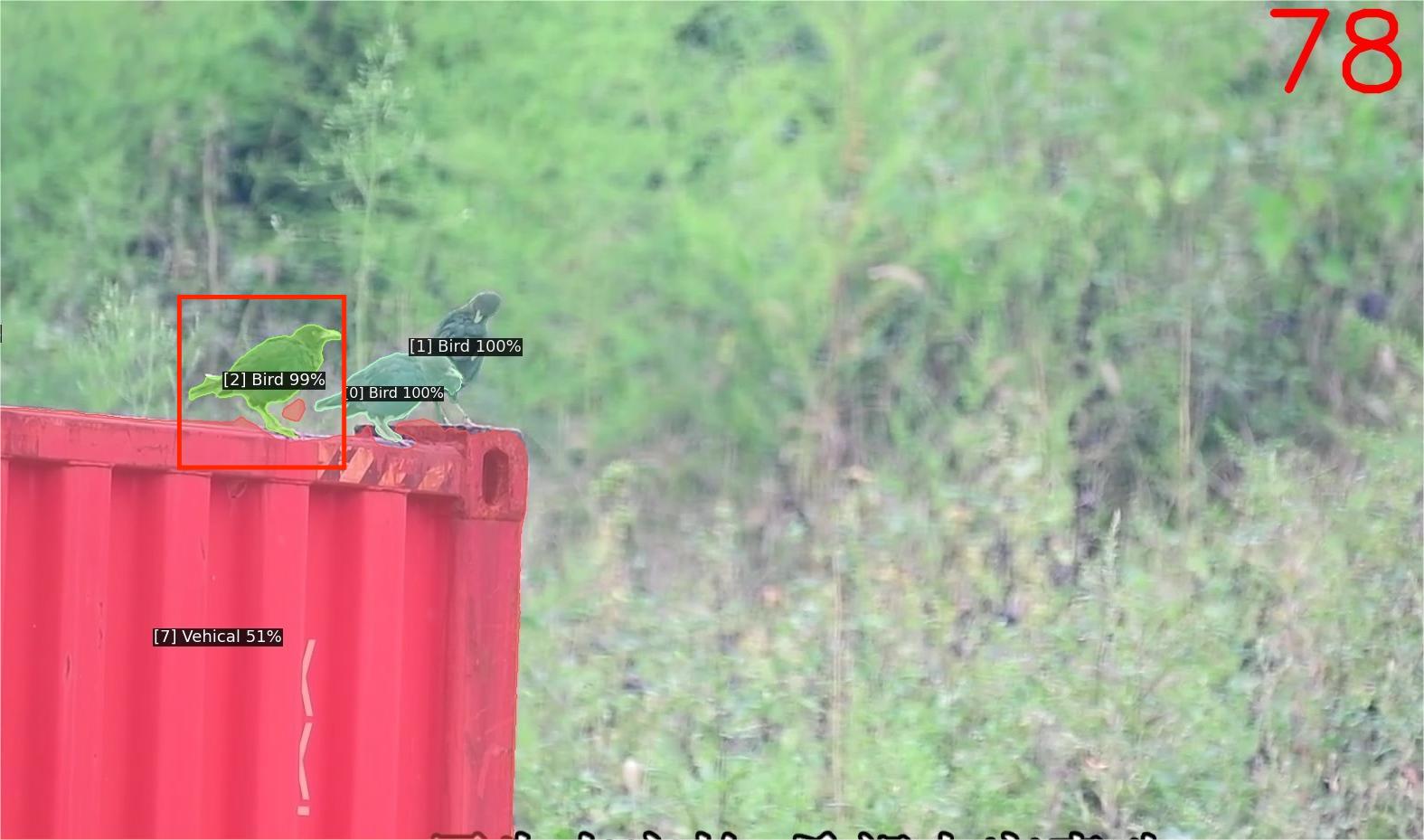}
\includegraphics[width=0.163\linewidth]{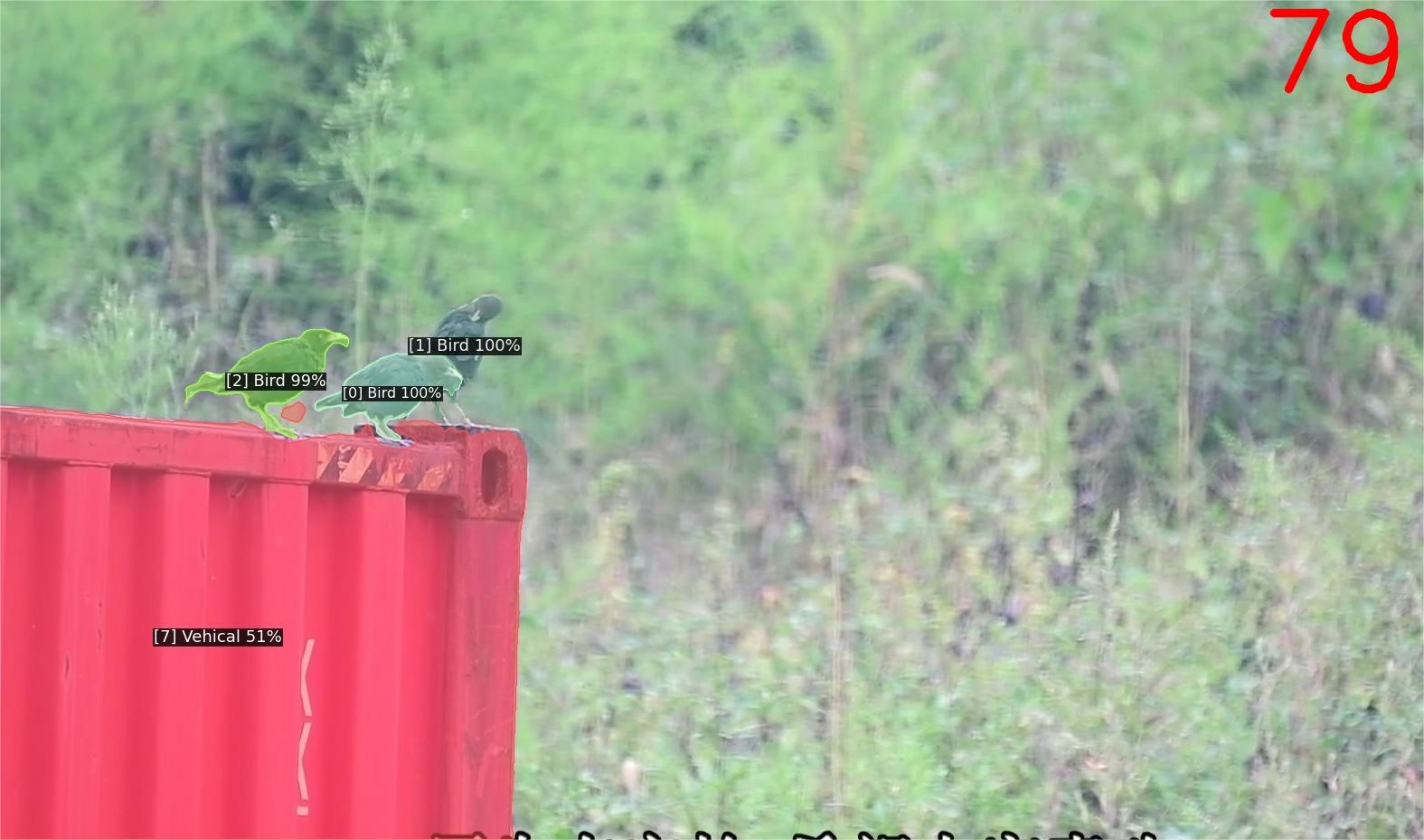}
\includegraphics[width=0.163\linewidth]{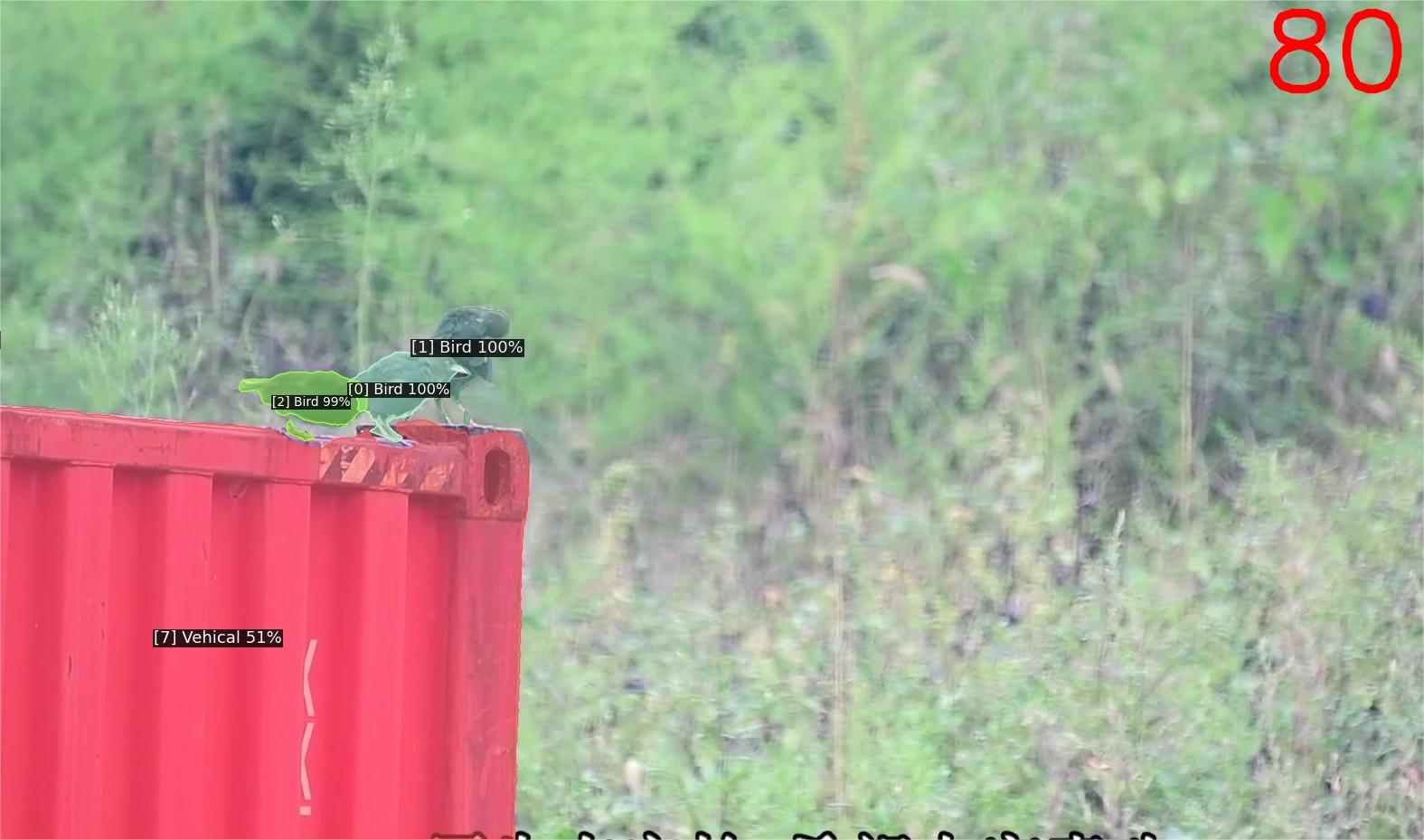}
\includegraphics[width=0.163\linewidth]{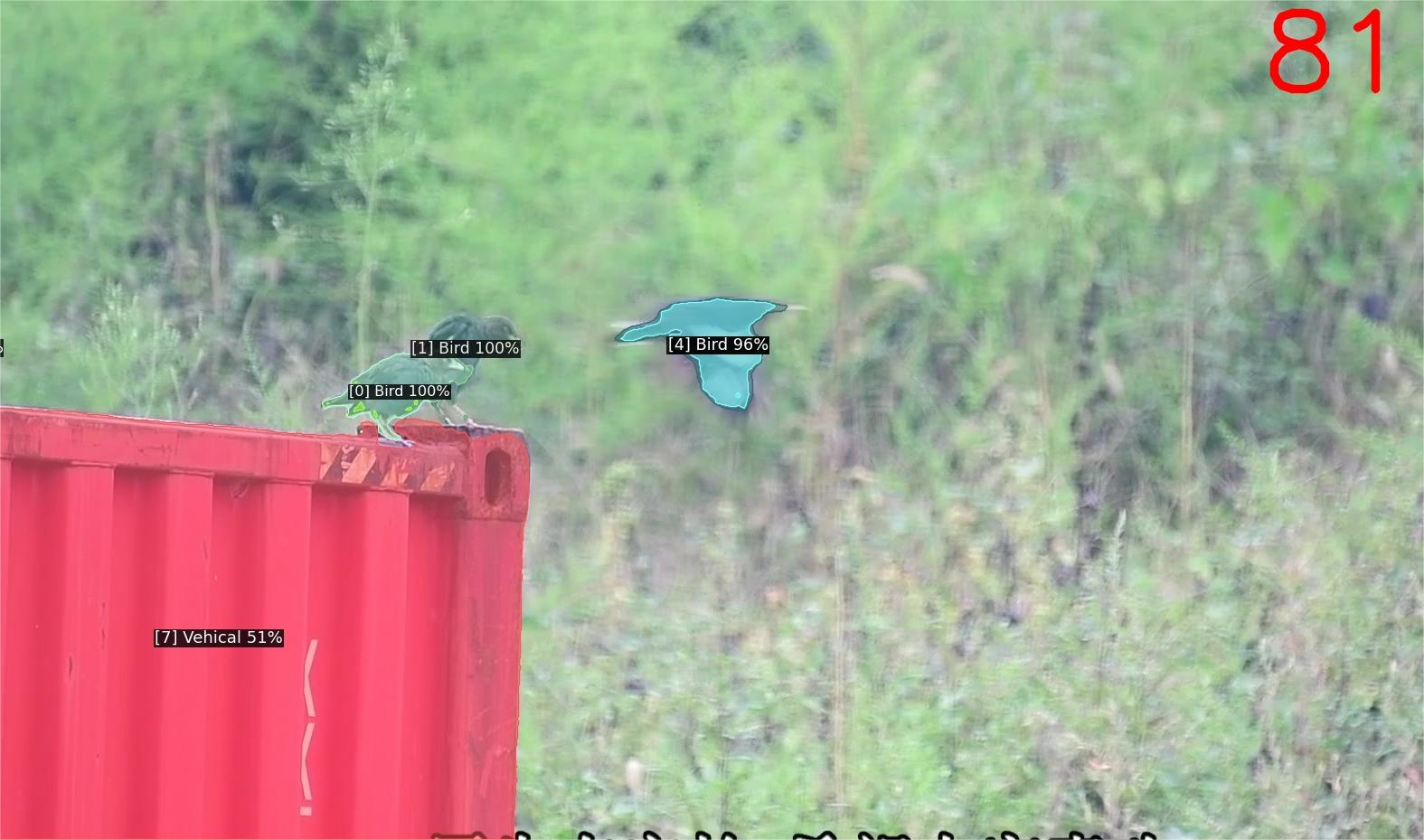}
\includegraphics[width=0.163\linewidth]{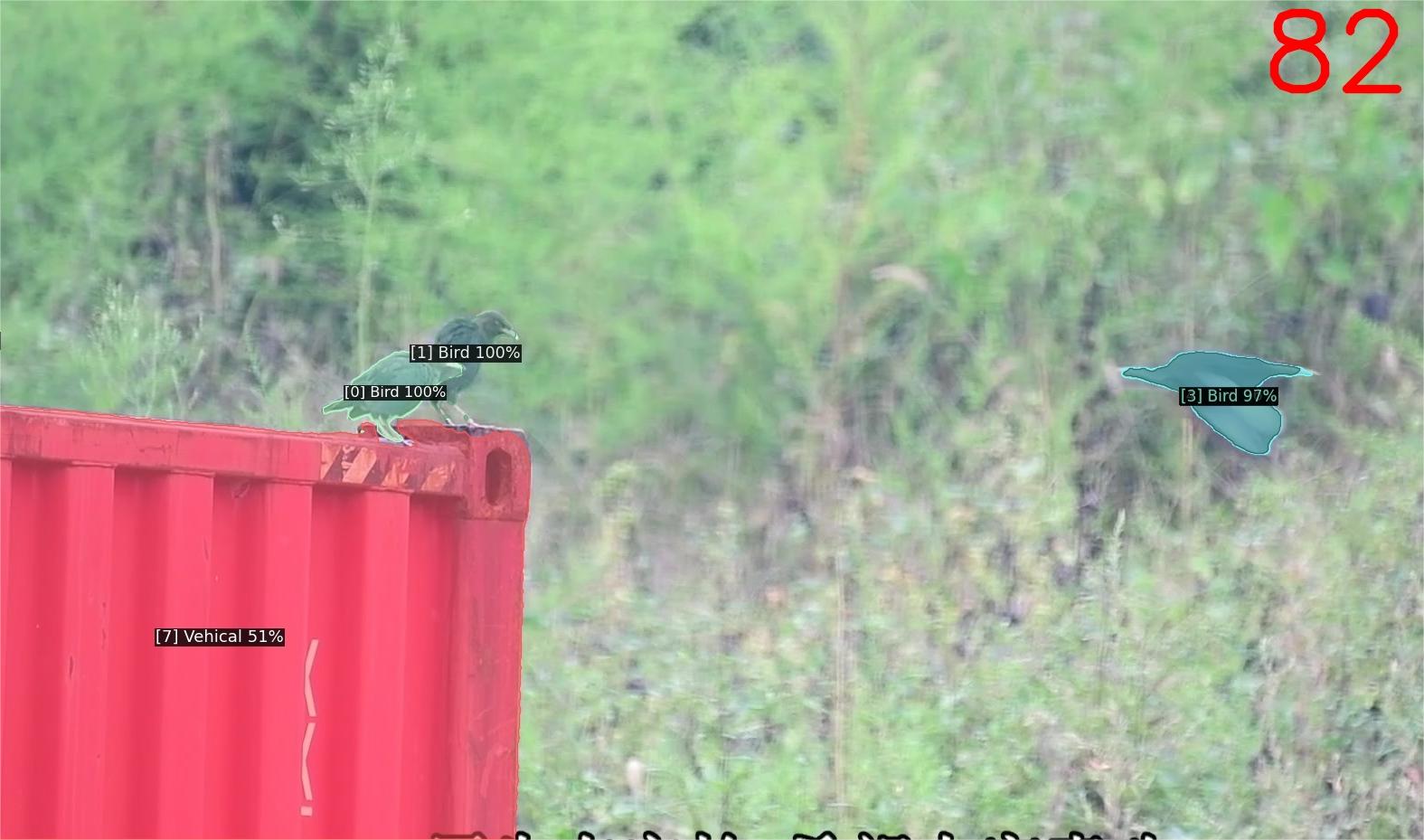}
\includegraphics[width=0.163\linewidth]{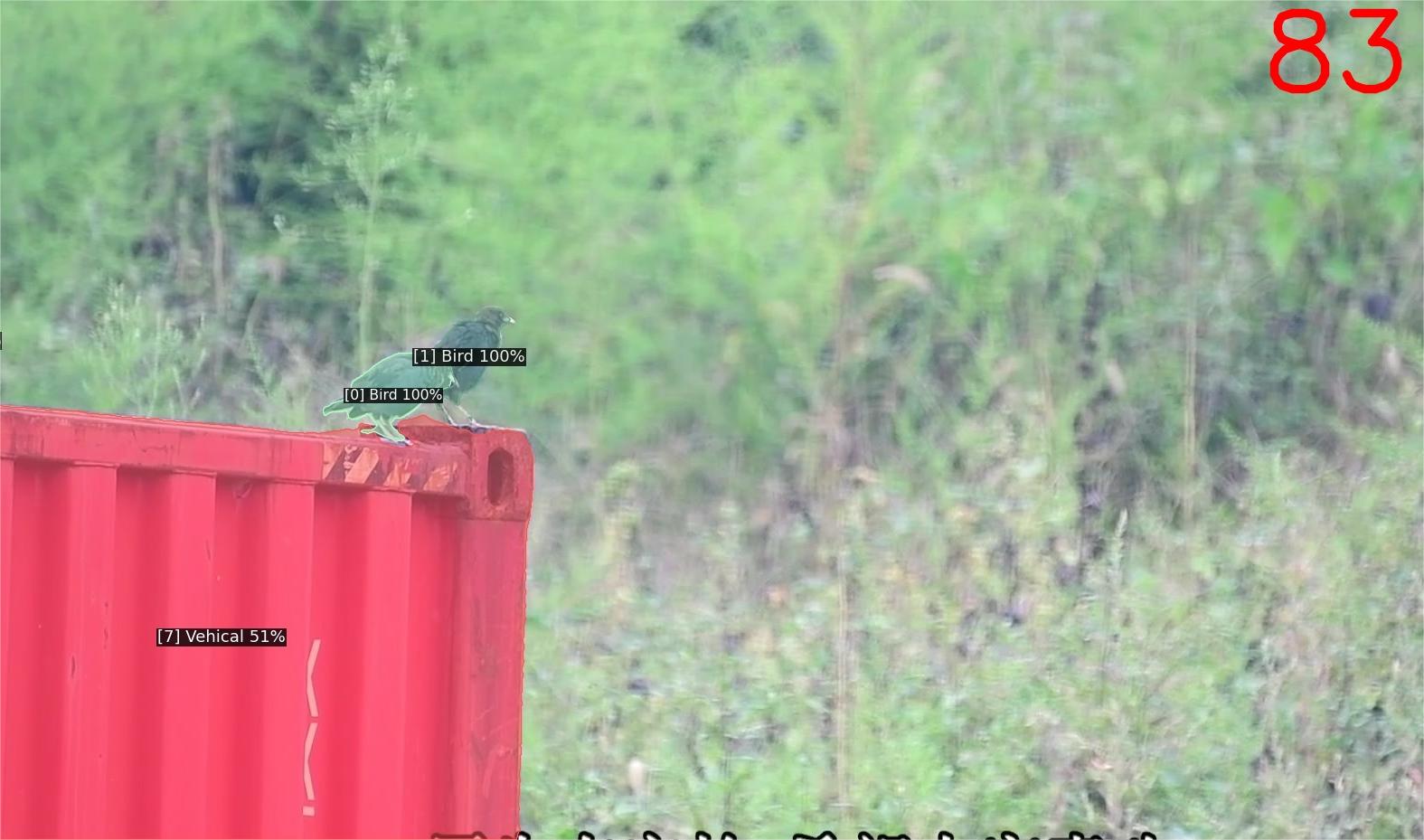}
\end{minipage}\hfill\vspace{1mm}

\begin{minipage}[c]{1.00\linewidth}
\includegraphics[width=0.163\linewidth]{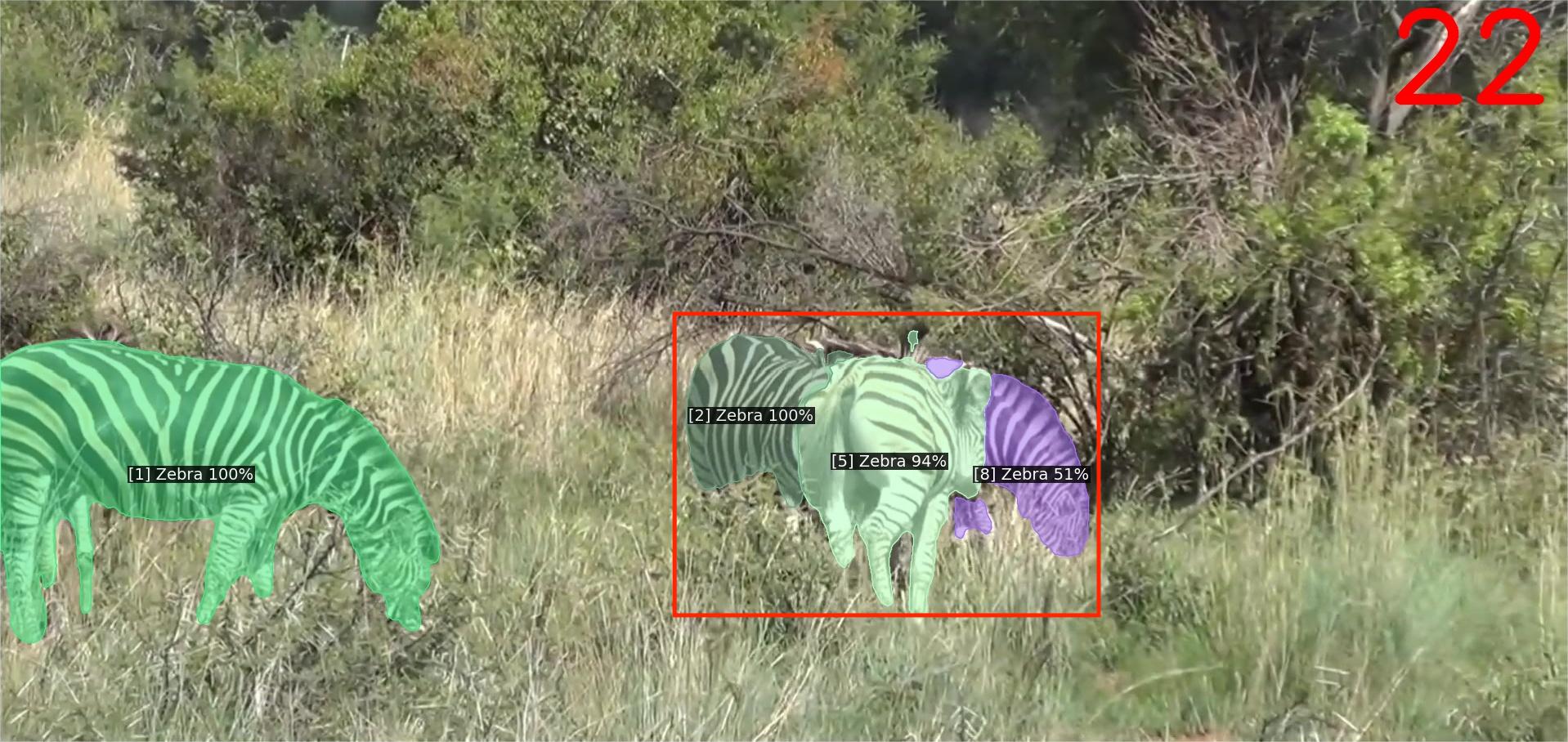}
\includegraphics[width=0.163\linewidth]{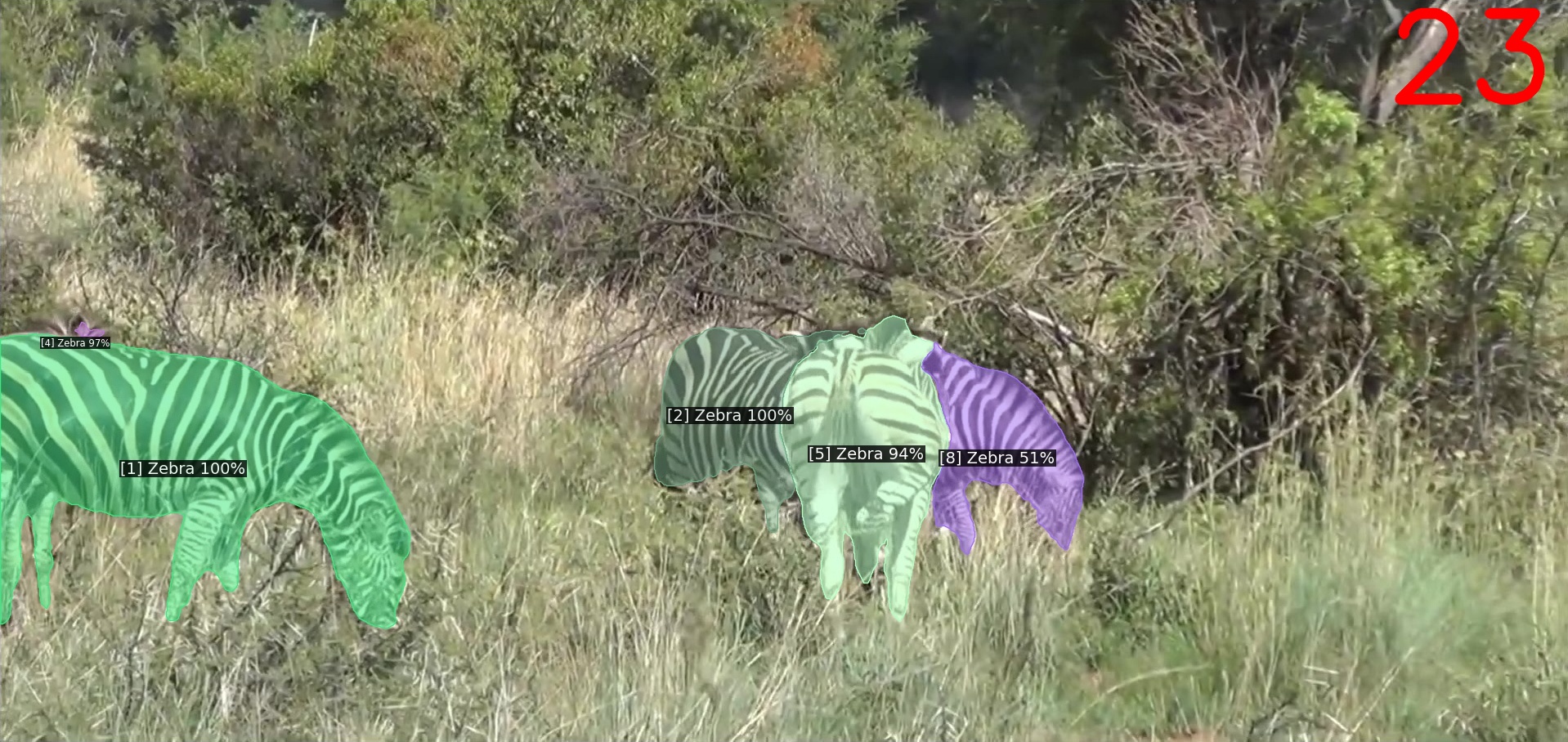}
\includegraphics[width=0.163\linewidth]{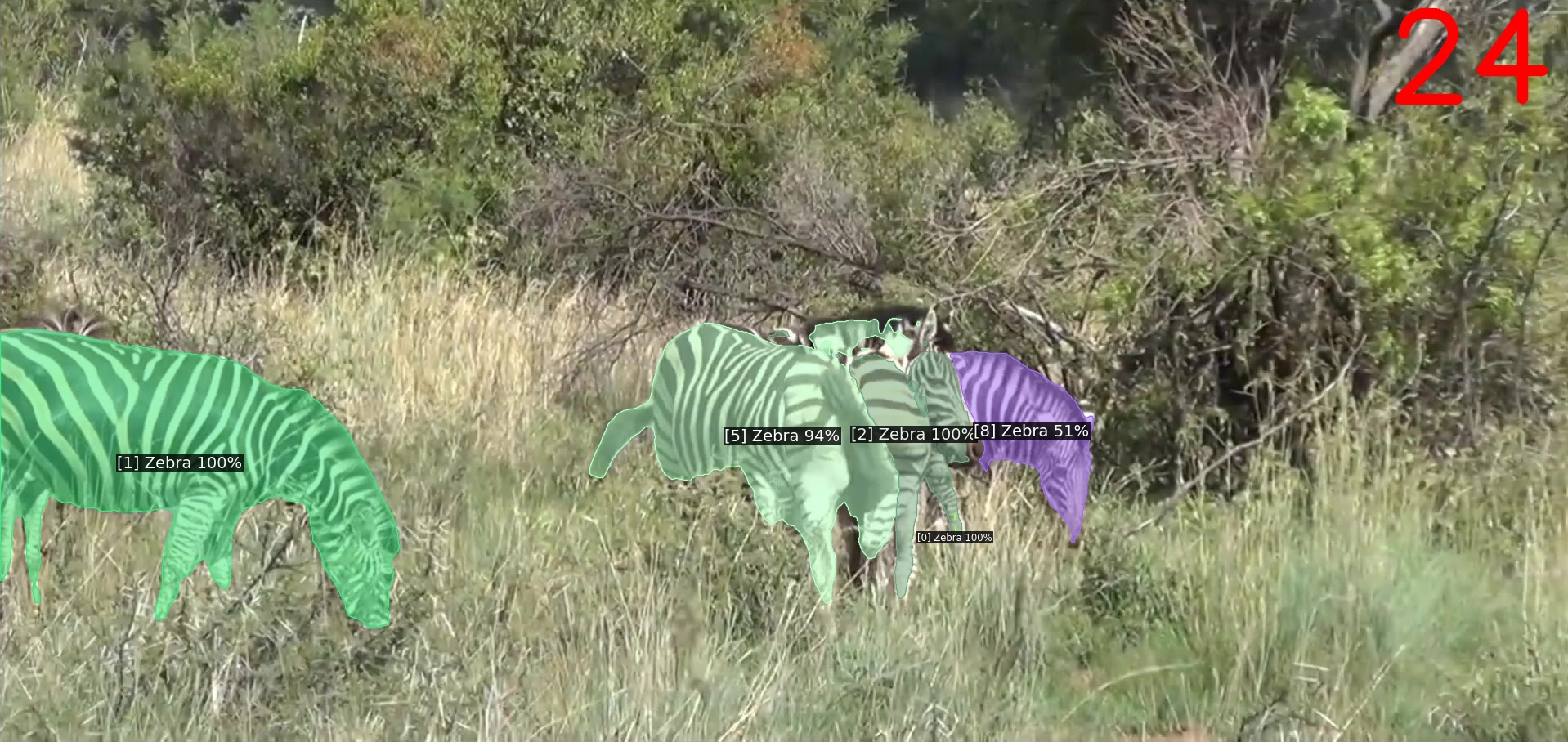}
\includegraphics[width=0.163\linewidth]{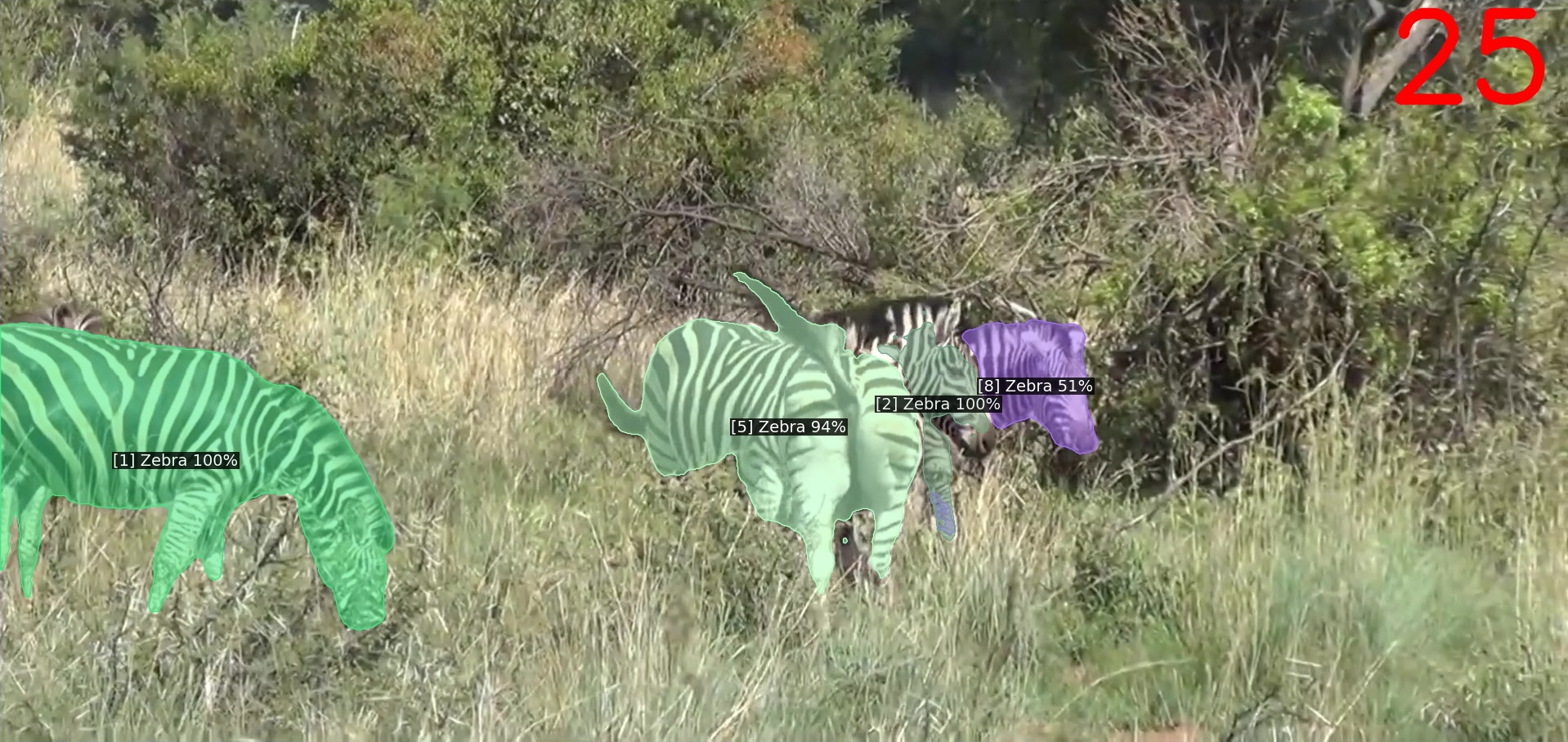}
\includegraphics[width=0.163\linewidth]{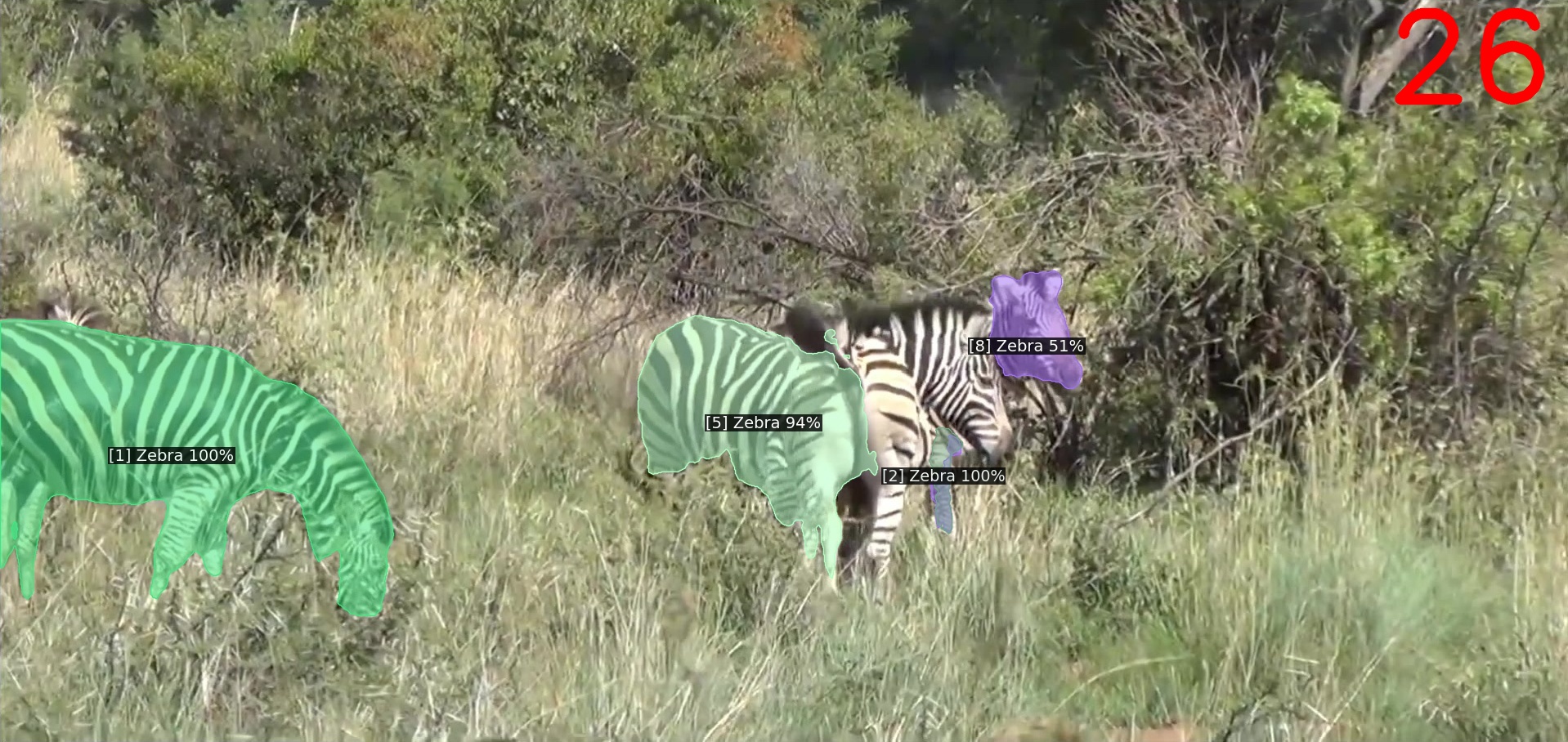}
\includegraphics[width=0.163\linewidth]{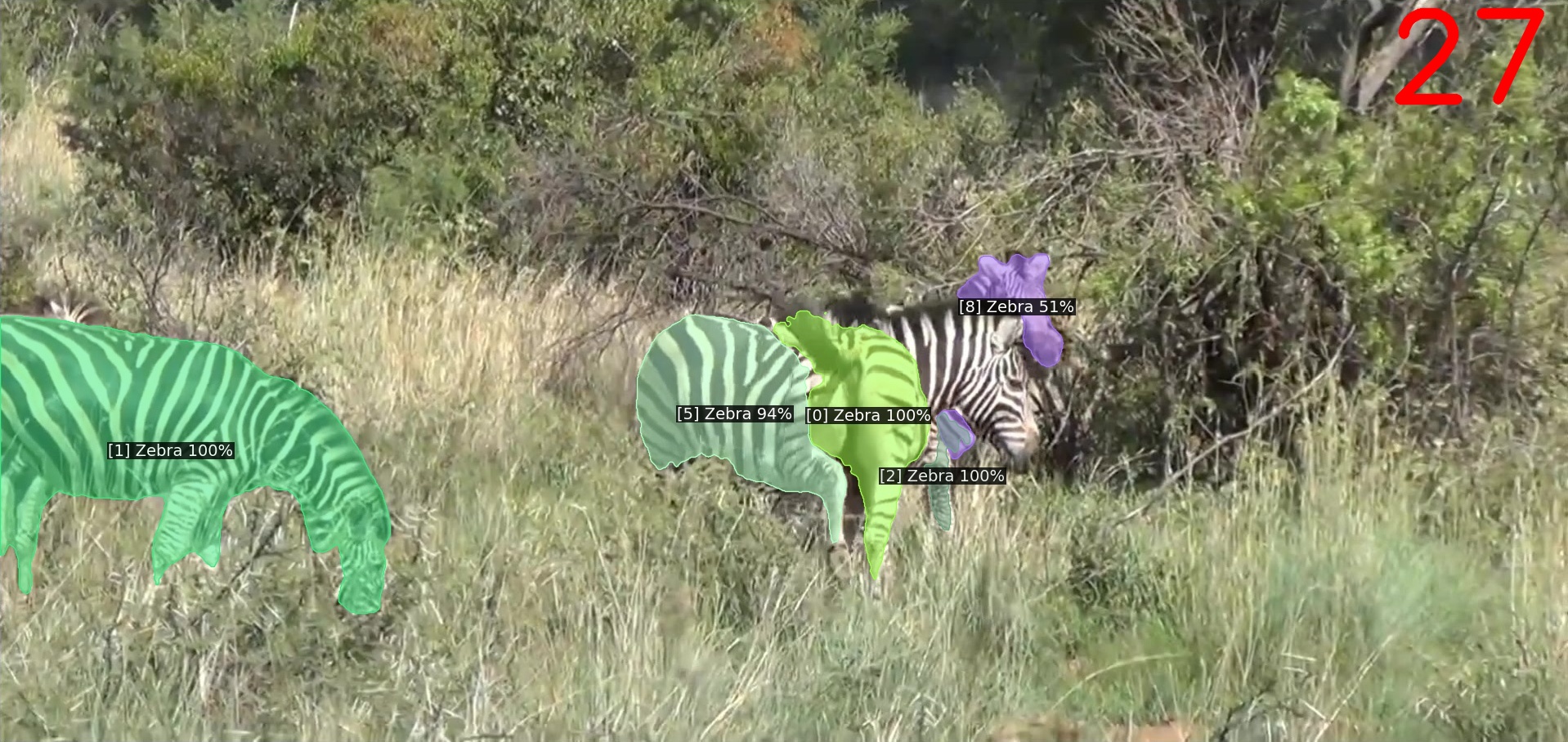}
\end{minipage}\hfill
\begin{minipage}[c]{1.00\linewidth}
\includegraphics[width=0.163\linewidth]{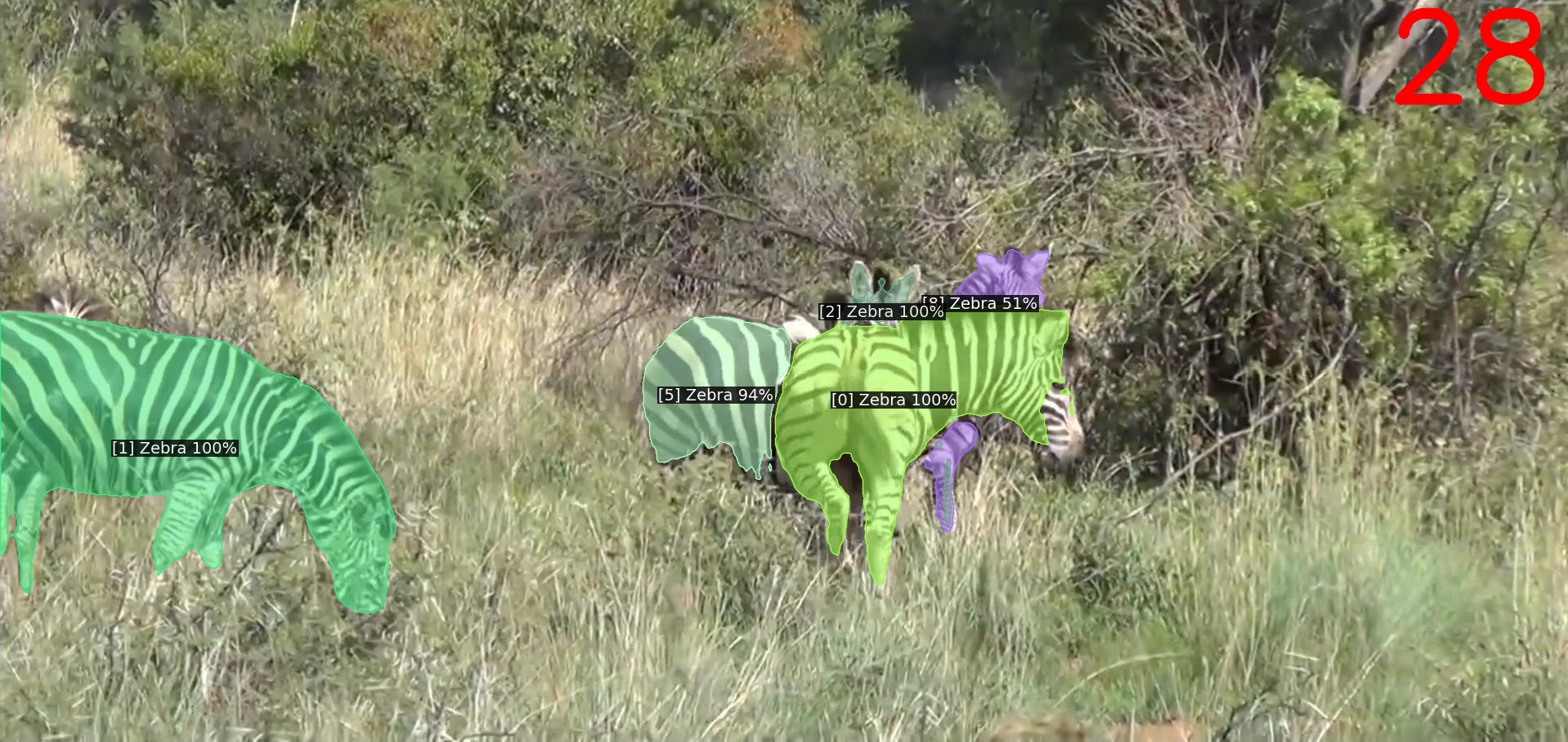}
\includegraphics[width=0.163\linewidth]{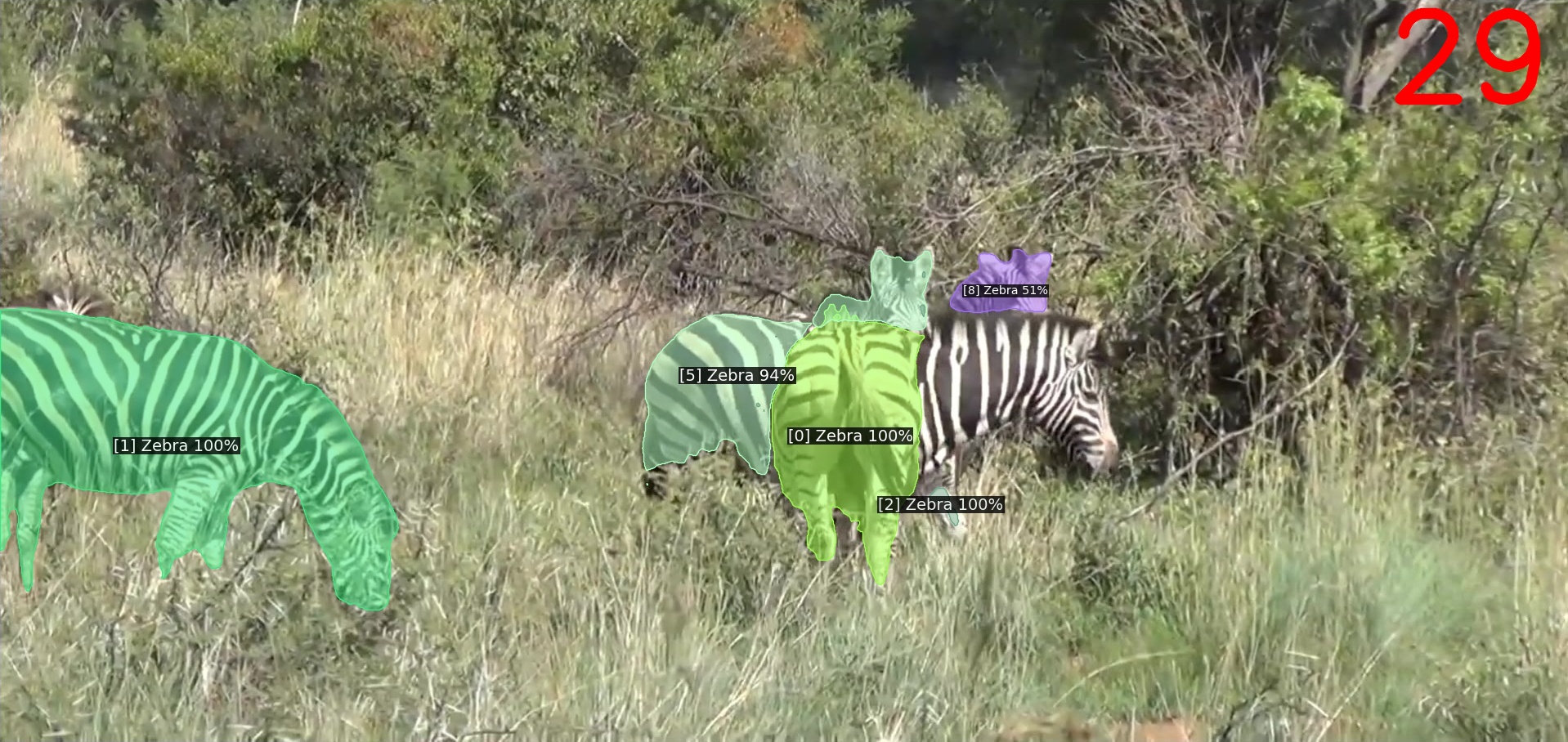}
\includegraphics[width=0.163\linewidth]{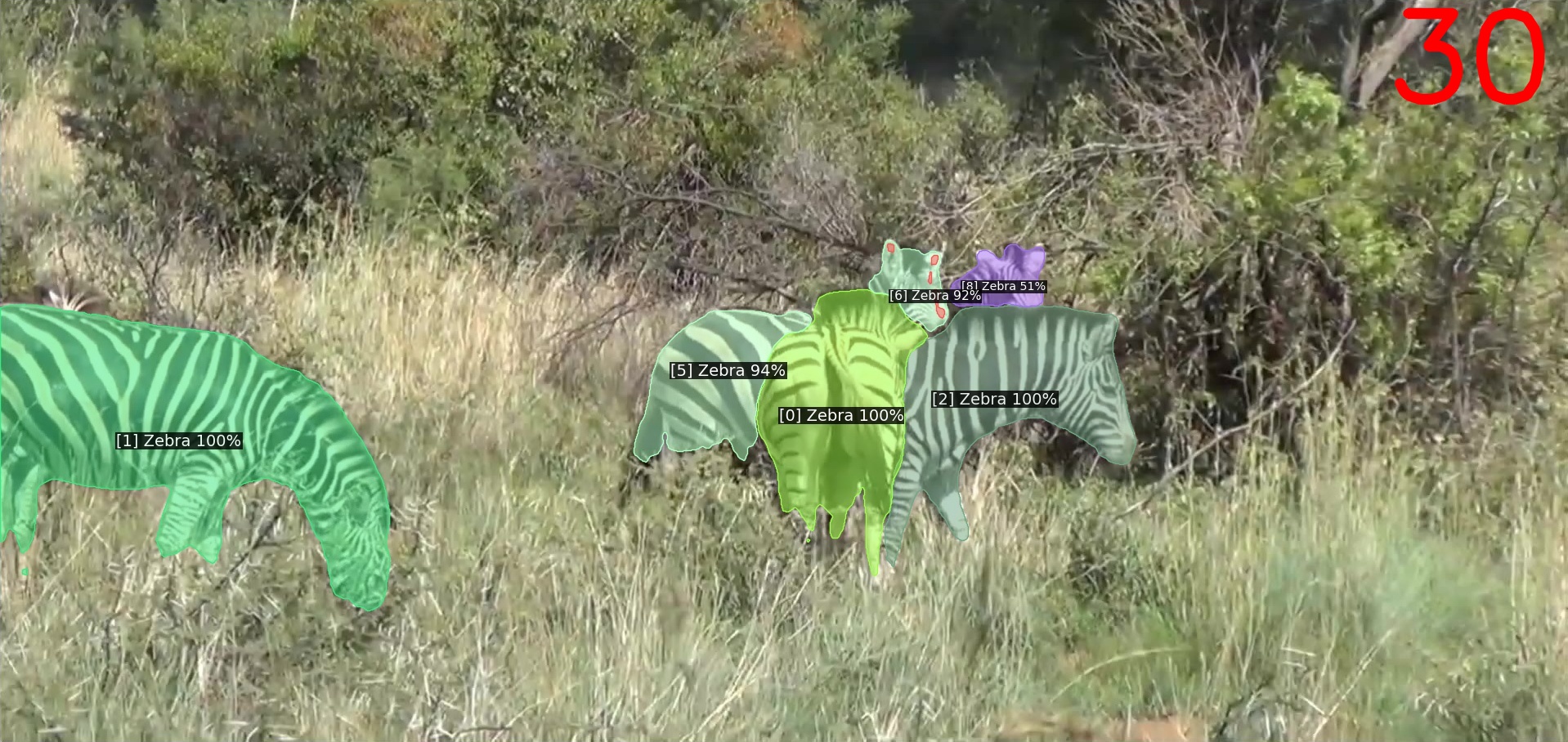}
\includegraphics[width=0.163\linewidth]{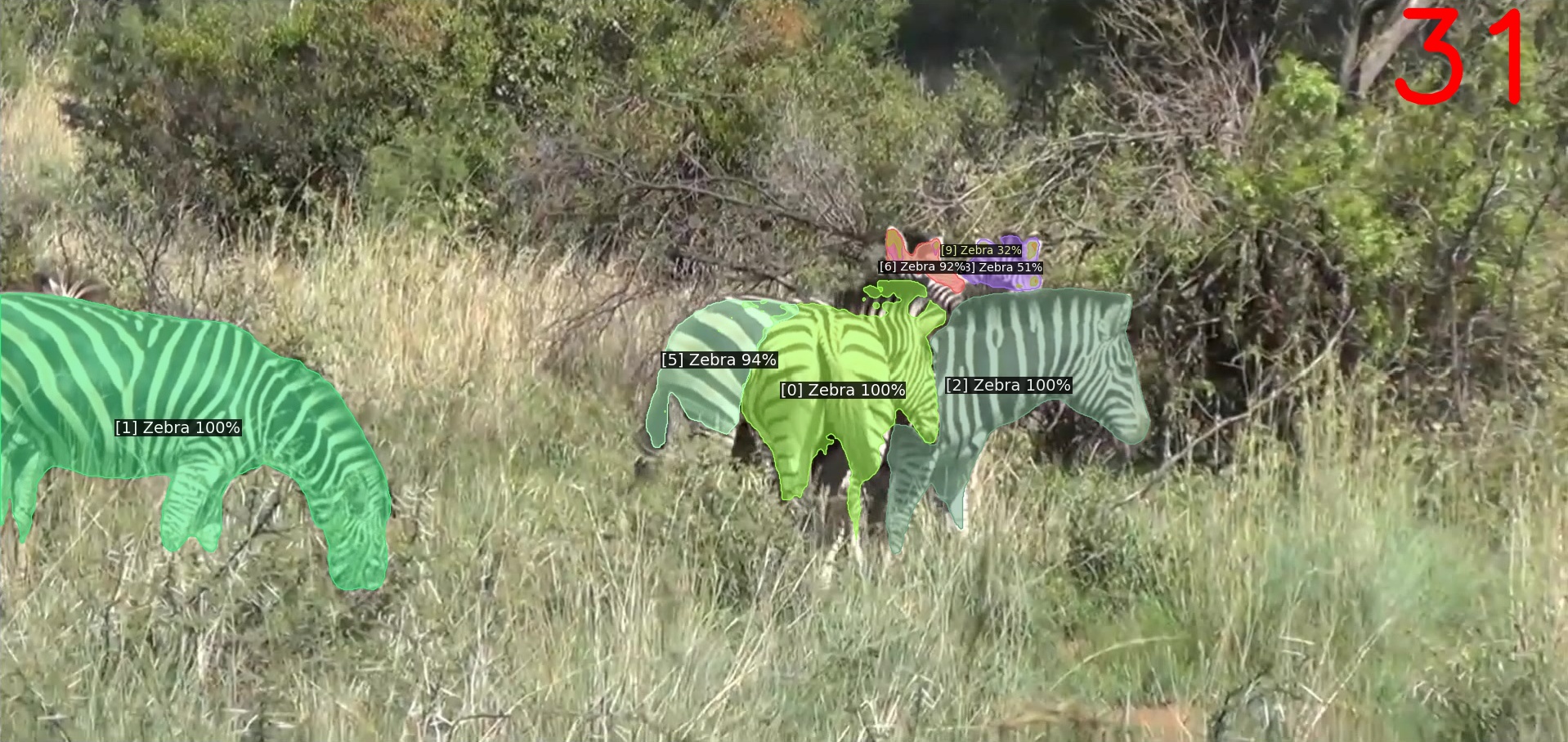}
\includegraphics[width=0.163\linewidth]{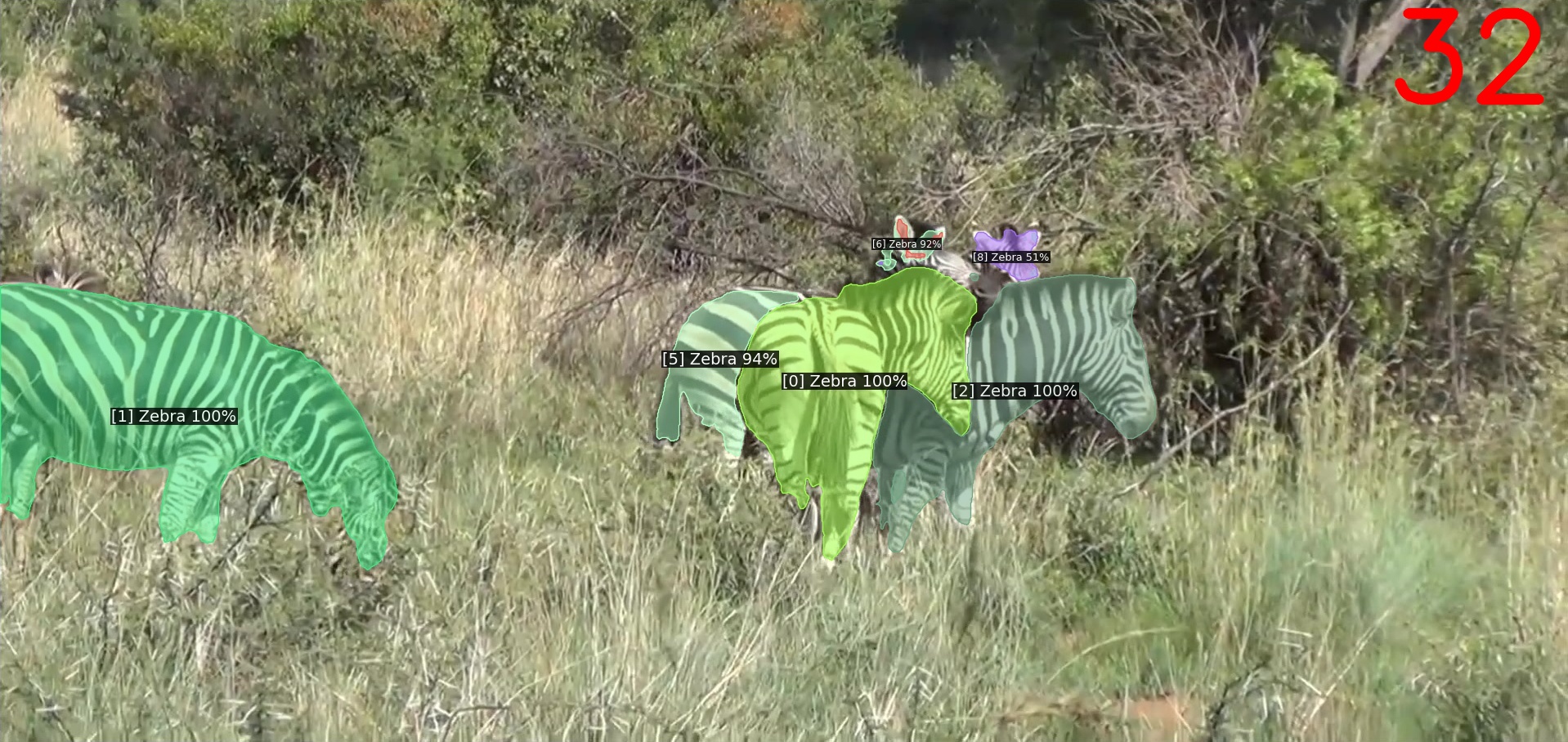}
\includegraphics[width=0.163\linewidth]{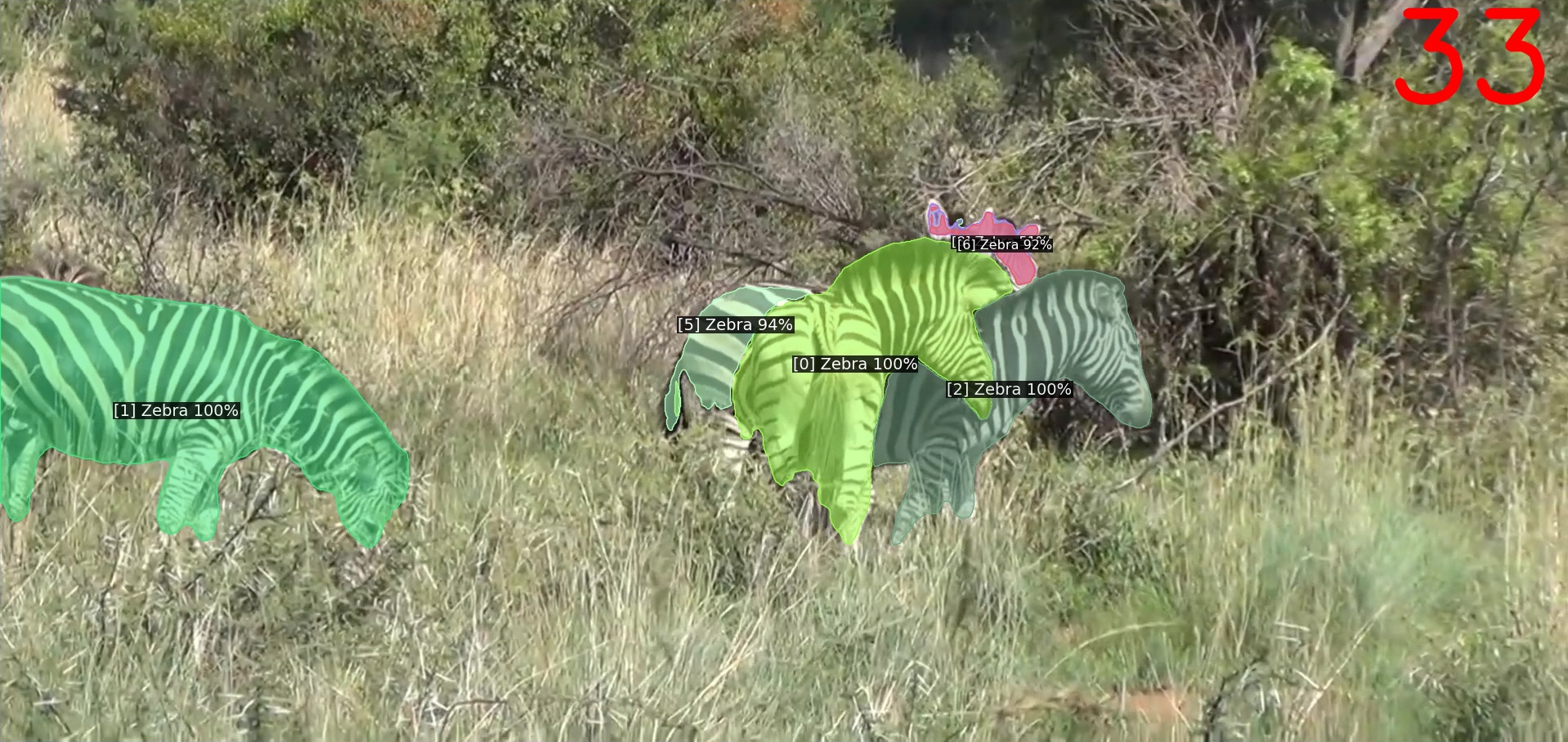}
\end{minipage}\hfill

\caption{The failure cases of DVIS++. In the first image of the videos, the failed segmentation objects are highlighted with red bounding boxes. DVIS++ fails to obtain accurate tracking results when objects move at high speeds (case 1). Similarly, when the segmentation model cannot correctly differentiate extremely similar objects (case 2), DVIS++ also fails to achieve accurate segmentation results.}
\label{fig:demos_failure}
\end{figure*}

We conduct ablation experiments on the denoising training strategy to explore the effects of noise simulation strategies, noise injection probability, and iteration number. The results are shown in Table~\ref{tab:denoising strategy}. Firstly, different noise simulation strategies, including random weighted averaging (WA), random cropping and concatenation (CC), and random shuffling (RS), are used, and their performance is shown in $S0$ to $S3$. Regardless of the noise simulation strategy used, we observe performance improvement. Among them, the WA strategy achieves the best performance, with a performance increase of 2.2 AP compared to no noise added. The RS strategy can be considered an extreme case of WA and CC, but its performance improvement is lower than that of the WA strategy. This is because although the RS strategy introduces strong noise, it significantly reduces the noise sampling space. The improvement of the denoising strategy is mainly reflected in occluded objects (with an increase of 3.3 AP$_{\rm m}$ and 2.9 AP$_{\rm h}$), while it has no significant effect on slightly occluded objects (with a decrease of 0.4 AP$_{\rm l}$).

In addition, we are surprised to find that the denoising training strategy significantly improves performance as the number of training iterations increases. When the training iterations are extended from 40K to 160K, the WA strategy achieves 36.7 AP, resulting in a performance gain of 1.4 AP.

The probability of adding noise, $P$, also influences the effectiveness of the denoising training strategy. As demonstrated in $S6$ to $S8$, the optimal performance is attained when $P$ is set to 0.5. However, when longer training iterations are employed, a higher probability of adding noise leads to improved network training. Ultimately, by adopting the WA strategy, setting $P$ to 0.8 and the iteration number to 160K, the model performance obtains a notable enhancement of 4.1 AP compared to not utilizing the denoising training strategy.

\noindent\textbf{Contrastive Losses.} We investigate the impact of contrastive learning on the segmenter, referring tracker, and temporal refiner. As shown in Table~\ref{tab:ablation}, utilizing contrastive loss during the training process of the segmenter results in more distinct object representations, leading to performance improvements of 0.8 AP, 8.0 AP$_{\rm l}$, 1.3 AP$_{\rm m}$, and 0.4 AP$_{\rm h}$ ($\mathcal{M}1$ \textit{vs.} $\mathcal{M}0$). Notably, contrastive learning has proved highly effective for objects with minor deformations, such as lightly occluded objects (8.0 AP$_{\rm l}$ improvement). However, it does not yield for heavily occluded objects with significant deformations (only 0.4 AP$_{\rm h}$ improvement).

When contrastive loss is implemented in the training process of the referring tracker to enhance the consistency of adjacent frame references, it results in improvements of 0.7 AP, 1.1 AP$_{\rm l}$, 1.2 AP$_{\rm m}$ ($\mathcal{M}4$ \textit{vs.} $\mathcal{M}3$). The utilization of contrastive loss in the training of both the segmenter and referring tracker significantly improve the segmentation results for lightly occluded objects. However, contrastive loss does not enable satisfactory results for heavily occluded objects with large deformations.

Unfortunately, the use of contrastive loss during the training process of the temporal refiner results in a decrease in performance ($\mathcal{M}6$ \textit{vs.} $\mathcal{M}5$), with 0.6 AP, 0.7 AP$_{\rm m}$, and 2.2 AP$_{\rm h}$, despite a performance increase of 3.3 AP$_{\rm l}$. The implementation of contrastive loss has the unintended consequence of suppressing the distinctions in those heavily occluded object representations across temporal frames. However, it has a beneficial impact on lightly occluded objects.

\noindent\textbf{Qualitative Analysis.} The video segmentation results for DVIS++ are presented in Figure~\ref{fig:demos}. In the VIS prediction results, the three horses undergo significant deformation and severe occlusion. Despite these challenges, DVIS++ still manages to achieve flawless results. The perfect prediction results for VPS demonstrate DVIS++'s excellent capability in handling both 'thing' and 'stuff' objects. Furthermore, the prediction results for VSS underscore DVIS++'s exceptional segmentation quality and its remarkable temporal consistency.

The open-vocabulary segmentation results of OV-DVIS++ are shown in Figure~\ref{fig:demos_ov}. It can be observed that the model is capable of effectively tracking and segmenting new categories, such as ``carrot", ``hay", ``lantern", etc., even if they are not present in the training data.

However, some failure cases still exist in the prediction results of DVIS++, as shown in Figure~\ref{fig:demos_failure}. Firstly, DVIS++ cannot track fast-moving objects well. In the first video, when the bird with ID 2 suddenly takes off and moves quickly, DVIS++ mistakenly identifies it as a new object in subsequent frames. We think this problem can be alleviated by properly introducing the trajectory model information. Additionally, DVIS++ relies on a segmenter to perceive individual images, and when the segmenter does not work well, DVIS++ will fail. For example, in the second video, the segmenter fails to distinguish the closely clustered zebras, resulting in incorrect output from DVIS++. We will make efforts to address these issues in future work.

\section{Conclusion}
We introduce DVIS++, a novel universal video segmentation framework that effectively models the spatio-temporal representation of video objects through a decoupled design. It performs SOTA on six mainstream VIS, VSS, and VPS benchmarks. By leveraging CLIP, we also implement OV-DVIS++, an open-vocabulary universal video segmentation framework that achieves SOTA performance in zero-shot inference. Specifically, we decompose the video segmentation task into segmentation, tracking, and refinement. We propose novel referring tracker and temporal refiner modules to handle the tracking and refinement subtasks, respectively. To enhance the tracking capability of the referring tracker, we design a denoising training strategy. Furthermore, we investigate the impact of contrastive learning on the segmenter, referring tracker, and temporal refiner, highlighting its significance in video segmentation networks. Moreover, combining the visual foundation models, DVIS++ is evaluated under various settings. In the open-vocabulary setting, OV-DVIS++ achieves SOTA performance. Additionally, when evaluated with a frozen pre-trained backbone, DVIS++ works well and achieves higher performance.

We believe that DVIS, DVIS++, and OV-DVIS++ will serve as strong baselines for video universal segmentation, fostering future research in the fields of VIS, VSS, VPS, and related areas.


%

%
%
%
%
%

\ifCLASSOPTIONcaptionsoff
  \newpage
\fi



%

{
	\bibliographystyle{IEEEtran}
	\bibliography{IEEEabrv,main}

\begin{thebibliography}{10}
\providecommand{\url}[1]{#1}
\csname url@samestyle\endcsname
\providecommand{\newblock}{\relax}
\providecommand{\bibinfo}[2]{#2}
\providecommand{\BIBentrySTDinterwordspacing}{\spaceskip=0pt\relax}
\providecommand{\BIBentryALTinterwordstretchfactor}{4}
\providecommand{\BIBentryALTinterwordspacing}{\spaceskip=\fontdimen2\font plus
\BIBentryALTinterwordstretchfactor\fontdimen3\font minus
  \fontdimen4\font\relax}
\providecommand{\BIBforeignlanguage}[2]{{%
\expandafter\ifx\csname l@#1\endcsname\relax
\typeout{** WARNING: IEEEtran.bst: No hyphenation pattern has been}%
\typeout{** loaded for the language `#1'. Using the pattern for}%
\typeout{** the default language instead.}%
\else
\language=\csname l@#1\endcsname
\fi
#2}}
\providecommand{\BIBdecl}{\relax}
\BIBdecl

\bibitem{zhang2016instance}
Z.~Zhang, S.~Fidler, and R.~Urtasun, ``Instance-level segmentation for
  autonomous driving with deep densely connected mrfs,'' in \emph{Proceedings
  of the IEEE Conference on Computer Vision and Pattern Recognition}, 2016, pp.
  669--677.

\bibitem{zhou2022survey}
T.~Zhou, F.~Porikli, D.~J. Crandall, L.~Van~Gool, and W.~Wang, ``A survey on
  deep learning technique for video segmentation,'' \emph{IEEE Transactions on
  Pattern Analysis and Machine Intelligence}, vol.~45, no.~6, pp. 7099--7122,
  2022.

\bibitem{yang2019video}
L.~Yang, Y.~Fan, and N.~Xu, ``Video instance segmentation,'' in
  \emph{Proceedings of the IEEE/CVF International Conference on Computer
  Vision}, 2019, pp. 5188--5197.

\bibitem{wu2022seqformer}
J.~Wu, Y.~Jiang, S.~Bai, W.~Zhang, and X.~Bai, ``Seqformer: Sequential
  transformer for video instance segmentation,'' in \emph{European Conference
  on Computer Vision}.\hskip 1em plus 0.5em minus 0.4em\relax Springer, 2022,
  pp. 553--569.

\bibitem{weng2023mask}
Y.~Weng, M.~Han, H.~He, M.~Li, L.~Yao, X.~Chang, and B.~Zhuang, ``Mask
  propagation for efficient video semantic segmentation,'' \emph{arXiv preprint
  arXiv:2310.18954}, 2023.

\bibitem{athar2023tarvis}
A.~Athar, A.~Hermans, J.~Luiten, D.~Ramanan, and B.~Leibe, ``Tarvis: A unified
  approach for target-based video segmentation,'' in \emph{Proceedings of the
  IEEE/CVF Conference on Computer Vision and Pattern Recognition}, 2023, pp.
  18\,738--18\,748.

\bibitem{li2023tube}
X.~Li, H.~Yuan, W.~Zhang, G.~Cheng, J.~Pang, and C.~C. Loy, ``Tube-link: A
  flexible cross tube framework for universal video segmentation,'' in
  \emph{Proceedings of the IEEE/CVF International Conference on Computer Vision
  (ICCV)}, October 2023, pp. 13\,923--13\,933.

\bibitem{kim2020video}
D.~Kim, S.~Woo, J.-Y. Lee, and I.~S. Kweon, ``Video panoptic segmentation,'' in
  \emph{Proceedings of the IEEE/CVF Conference on Computer Vision and Pattern
  Recognition}, 2020, pp. 9859--9868.

\bibitem{wang2021end}
Y.~Wang, Z.~Xu, X.~Wang, C.~Shen, B.~Cheng, H.~Shen, and H.~Xia, ``End-to-end
  video instance segmentation with transformers,'' in \emph{Proceedings of the
  IEEE/CVF conference on computer vision and pattern recognition}, 2021, pp.
  8741--8750.

\bibitem{hwang2021video}
S.~Hwang, M.~Heo, S.~W. Oh, and S.~J. Kim, ``Video instance segmentation using
  inter-frame communication transformers,'' \emph{Advances in Neural
  Information Processing Systems}, vol.~34, pp. 13\,352--13\,363, 2021.

\bibitem{cheng2021mask2former}
B.~Cheng, A.~Choudhuri, I.~Misra, A.~Kirillov, R.~Girdhar, and A.~G. Schwing,
  ``Mask2former for video instance segmentation,'' \emph{arXiv preprint
  arXiv:2112.10764}, 2021.

\bibitem{heo2022vita}
M.~Heo, S.~Hwang, S.~W. Oh, J.-Y. Lee, and S.~J. Kim, ``Vita: Video instance
  segmentation via object token association,'' \emph{Advances in Neural
  Information Processing Systems}, vol.~35, pp. 23\,109--23\,120, 2022.

\bibitem{huang2022minvis}
D.-A. Huang, Z.~Yu, and A.~Anandkumar, ``Minvis: A minimal video instance
  segmentation framework without video-based training,'' \emph{Advances in
  Neural Information Processing Systems}, vol.~35, pp. 31\,265--31\,277, 2022.

\bibitem{wu2022defense}
J.~Wu, Q.~Liu, Y.~Jiang, S.~Bai, A.~Yuille, and X.~Bai, ``In defense of online
  models for video instance segmentation,'' in \emph{European Conference on
  Computer Vision}.\hskip 1em plus 0.5em minus 0.4em\relax Springer, 2022, pp.
  588--605.

\bibitem{heo2023generalized}
M.~Heo, S.~Hwang, J.~Hyun, H.~Kim, S.~W. Oh, J.-Y. Lee, and S.~J. Kim, ``A
  generalized framework for video instance segmentation,'' in \emph{Proceedings
  of the IEEE/CVF Conference on Computer Vision and Pattern Recognition}, 2023,
  pp. 14\,623--14\,632.

\bibitem{ying2023ctvis}
K.~Ying, Q.~Zhong, W.~Mao, Z.~Wang, H.~Chen, L.~Y. Wu, Y.~Liu, C.~Fan,
  Y.~Zhuge, and C.~Shen, ``Ctvis: Consistent training for online video instance
  segmentation,'' in \emph{Proceedings of the IEEE/CVF International Conference
  on Computer Vision (ICCV)}, October 2023, pp. 899--908.

\bibitem{qi2022occluded}
J.~Qi, Y.~Gao, Y.~Hu, X.~Wang, X.~Liu, X.~Bai, S.~Belongie, A.~Yuille, P.~H.
  Torr, and S.~Bai, ``Occluded video instance segmentation: A benchmark,''
  \emph{International Journal of Computer Vision}, vol. 130, no.~8, pp.
  2022--2039, 2022.

\bibitem{cheng2021per}
B.~Cheng, A.~Schwing, and A.~Kirillov, ``Per-pixel classification is not all
  you need for semantic segmentation,'' \emph{Advances in Neural Information
  Processing Systems}, vol.~34, pp. 17\,864--17\,875, 2021.

\bibitem{cheng2022masked}
B.~Cheng, I.~Misra, A.~G. Schwing, A.~Kirillov, and R.~Girdhar,
  ``Masked-attention mask transformer for universal image segmentation,'' in
  \emph{Proceedings of the IEEE/CVF conference on computer vision and pattern
  recognition}, 2022, pp. 1290--1299.

\bibitem{li2023mask}
F.~Li, H.~Zhang, H.~Xu, S.~Liu, L.~Zhang, L.~M. Ni, and H.-Y. Shum, ``Mask
  dino: Towards a unified transformer-based framework for object detection and
  segmentation,'' in \emph{Proceedings of the IEEE/CVF Conference on Computer
  Vision and Pattern Recognition}, 2023, pp. 3041--3050.

\bibitem{vaswani2017attention}
A.~Vaswani, N.~Shazeer, N.~Parmar, J.~Uszkoreit, L.~Jones, A.~N. Gomez,
  {\L}.~Kaiser, and I.~Polosukhin, ``Attention is all you need,''
  \emph{Advances in neural information processing systems}, vol.~30, 2017.

\bibitem{zhang2023dvis}
T.~Zhang, X.~Tian, Y.~Wu, S.~Ji, X.~Wang, Y.~Zhang, and P.~Wan, ``Dvis:
  Decoupled video instance segmentation framework,'' in \emph{Proceedings of
  the IEEE/CVF International Conference on Computer Vision (ICCV)}, October
  2023, pp. 1282--1291.

\bibitem{oquab2023dinov2}
M.~Oquab, T.~Darcet, T.~Moutakanni, H.~Vo, M.~Szafraniec, V.~Khalidov,
  P.~Fernandez, D.~Haziza, F.~Massa, A.~El-Nouby \emph{et~al.}, ``Dinov2:
  Learning robust visual features without supervision,'' \emph{arXiv preprint
  arXiv:2304.07193}, 2023.

\bibitem{radford2021learning}
A.~Radford, J.~W. Kim, C.~Hallacy, A.~Ramesh, G.~Goh, S.~Agarwal, G.~Sastry,
  A.~Askell, P.~Mishkin, J.~Clark \emph{et~al.}, ``Learning transferable visual
  models from natural language supervision,'' in \emph{International conference
  on machine learning}.\hskip 1em plus 0.5em minus 0.4em\relax PMLR, 2021, pp.
  8748--8763.

\bibitem{miao2022large}
J.~Miao, X.~Wang, Y.~Wu, W.~Li, X.~Zhang, Y.~Wei, and Y.~Yang, ``Large-scale
  video panoptic segmentation in the wild: A benchmark,'' in \emph{Proceedings
  of the IEEE/CVF Conference on Computer Vision and Pattern Recognition}, 2022,
  pp. 21\,033--21\,043.

\bibitem{miao2021vspw}
J.~Miao, Y.~Wei, Y.~Wu, C.~Liang, G.~Li, and Y.~Yang, ``Vspw: A large-scale
  dataset for video scene parsing in the wild,'' in \emph{Proceedings of the
  IEEE/CVF conference on computer vision and pattern recognition}, 2021, pp.
  4133--4143.

\bibitem{zhang20231stvps}
T.~Zhang, X.~Tian, H.~Wei, Y.~Wu, S.~Ji, X.~Wang, Y.~Zhang, and P.~Wan, ``1st
  place solution for pvuw challenge 2023: Video panoptic segmentation,''
  \emph{arXiv preprint arXiv:2306.04091}, 2023.

\bibitem{zhang20231stvis}
T.~Zhang, X.~Tian, Y.~Zhou, Y.~Wu, S.~Ji, C.~Yan, X.~Wang, X.~Tao, Y.~Zhang,
  and P.~Wan, ``1st place solution for the 5th lsvos challenge: Video instance
  segmentation,'' \emph{arXiv preprint arXiv:2308.14392}, 2023.

\bibitem{bai2009video}
X.~Bai, J.~Wang, D.~Simons, and G.~Sapiro, ``Video snapcut: robust video object
  cutout using localized classifiers,'' \emph{ACM Transactions on Graphics
  (ToG)}, vol.~28, no.~3, pp. 1--11, 2009.

\bibitem{bai2009geodesic}
X.~Bai and G.~Sapiro, ``Geodesic matting: A framework for fast interactive
  image and video segmentation and matting,'' \emph{International journal of
  computer vision}, vol.~82, pp. 113--132, 2009.

\bibitem{mu2007automatic}
Y.~Mu, H.~Zhang, H.~Wang, and W.~Zuo, ``Automatic video object segmentation
  using graph cut,'' in \emph{2007 IEEE International Conference on Image
  Processing}, vol.~3.\hskip 1em plus 0.5em minus 0.4em\relax IEEE, 2007, pp.
  III--377.

\bibitem{shelhamer2016clockwork}
E.~Shelhamer, K.~Rakelly, J.~Hoffman, and T.~Darrell, ``Clockwork convnets for
  video semantic segmentation,'' in \emph{Computer Vision--ECCV 2016 Workshops:
  Amsterdam, The Netherlands, October 8-10 and 15-16, 2016, Proceedings, Part
  III 14}.\hskip 1em plus 0.5em minus 0.4em\relax Springer, 2016, pp. 852--868.

\bibitem{sun2022coarse}
G.~Sun, Y.~Liu, H.~Ding, T.~Probst, and L.~Van~Gool, ``Coarse-to-fine feature
  mining for video semantic segmentation,'' in \emph{Proceedings of the
  IEEE/CVF Conference on Computer Vision and Pattern Recognition}, 2022, pp.
  3126--3137.

\bibitem{sun2022mining}
G.~Sun, Y.~Liu, H.~Tang, A.~Chhatkuli, L.~Zhang, and L.~Van~Gool, ``Mining
  relations among cross-frame affinities for video semantic segmentation,'' in
  \emph{European Conference on Computer Vision}.\hskip 1em plus 0.5em minus
  0.4em\relax Springer, 2022, pp. 522--539.

\bibitem{zhu2017deep}
X.~Zhu, Y.~Xiong, J.~Dai, L.~Yuan, and Y.~Wei, ``Deep feature flow for video
  recognition,'' in \emph{Proceedings of the IEEE conference on computer vision
  and pattern recognition}, 2017, pp. 2349--2358.

\bibitem{he2017mask}
K.~He, G.~Gkioxari, P.~Doll{\'a}r, and R.~Girshick, ``Mask r-cnn,'' in
  \emph{Proceedings of the IEEE international conference on computer vision},
  2017, pp. 2961--2969.

\bibitem{cao2020sipmask}
J.~Cao, R.~M. Anwer, H.~Cholakkal, F.~S. Khan, Y.~Pang, and L.~Shao, ``Sipmask:
  Spatial information preservation for fast image and video instance
  segmentation,'' in \emph{Computer Vision--ECCV 2020: 16th European
  Conference, Glasgow, UK, August 23--28, 2020, Proceedings, Part XIV
  16}.\hskip 1em plus 0.5em minus 0.4em\relax Springer, 2020, pp. 1--18.

\bibitem{yang2021crossover}
S.~Yang, Y.~Fang, X.~Wang, Y.~Li, C.~Fang, Y.~Shan, B.~Feng, and W.~Liu,
  ``Crossover learning for fast online video instance segmentation,'' in
  \emph{Proceedings of the IEEE/CVF International Conference on Computer
  Vision}, 2021, pp. 8043--8052.

\bibitem{carion2020end}
N.~Carion, F.~Massa, G.~Synnaeve, N.~Usunier, A.~Kirillov, and S.~Zagoruyko,
  ``End-to-end object detection with transformers,'' in \emph{European
  conference on computer vision}.\hskip 1em plus 0.5em minus 0.4em\relax
  Springer, 2020, pp. 213--229.

\bibitem{hannan2023gratt}
T.~Hannan, R.~Koner, M.~Bernhard, S.~Shit, B.~Menze, V.~Tresp, M.~Schubert, and
  T.~Seidl, ``Gratt-vis: Gated residual attention for auto rectifying video
  instance segmentation,'' \emph{arXiv preprint arXiv:2305.17096}, 2023.

\bibitem{meinhardt2023novis}
T.~Meinhardt, M.~Feiszli, Y.~Fan, L.~Leal-Taixe, and R.~Ranjan, ``Novis: A case
  for end-to-end near-online video instance segmentation,'' \emph{arXiv
  preprint arXiv:2308.15266}, 2023.

\bibitem{wang2021max}
H.~Wang, Y.~Zhu, H.~Adam, A.~Yuille, and L.-C. Chen, ``Max-deeplab: End-to-end
  panoptic segmentation with mask transformers,'' in \emph{Proceedings of the
  IEEE/CVF conference on computer vision and pattern recognition}, 2021, pp.
  5463--5474.

\bibitem{zhang2021k}
W.~Zhang, J.~Pang, K.~Chen, and C.~C. Loy, ``K-net: Towards unified image
  segmentation,'' \emph{Advances in Neural Information Processing Systems},
  vol.~34, pp. 10\,326--10\,338, 2021.

\bibitem{yu2022k}
Q.~Yu, H.~Wang, S.~Qiao, M.~Collins, Y.~Zhu, H.~Adam, A.~Yuille, and L.-C.
  Chen, ``k-means mask transformer,'' in \emph{European Conference on Computer
  Vision}.\hskip 1em plus 0.5em minus 0.4em\relax Springer, 2022, pp. 288--307.

\bibitem{li2022video}
X.~Li, W.~Zhang, J.~Pang, K.~Chen, G.~Cheng, Y.~Tong, and C.~C. Loy, ``Video
  k-net: A simple, strong, and unified baseline for video segmentation,'' in
  \emph{Proceedings of the IEEE/CVF Conference on Computer Vision and Pattern
  Recognition}, 2022, pp. 18\,847--18\,857.

\bibitem{kim2022tubeformer}
D.~Kim, J.~Xie, H.~Wang, S.~Qiao, Q.~Yu, H.-S. Kim, H.~Adam, I.~S. Kweon, and
  L.-C. Chen, ``Tubeformer-deeplab: Video mask transformer,'' in
  \emph{Proceedings of the IEEE/CVF Conference on Computer Vision and Pattern
  Recognition}, 2022, pp. 13\,914--13\,924.

\bibitem{zhang2022dino}
H.~Zhang, F.~Li, S.~Liu, L.~Zhang, H.~Su, J.~Zhu, L.~M. Ni, and H.-Y. Shum,
  ``Dino: Detr with improved denoising anchor boxes for end-to-end object
  detection,'' \emph{arXiv preprint arXiv:2203.03605}, 2022.

\bibitem{tian2020conditional}
Z.~Tian, C.~Shen, and H.~Chen, ``Conditional convolutions for instance
  segmentation,'' in \emph{Computer Vision--ECCV 2020: 16th European
  Conference, Glasgow, UK, August 23--28, 2020, Proceedings, Part I 16}.\hskip
  1em plus 0.5em minus 0.4em\relax Springer, 2020, pp. 282--298.

\bibitem{bolya2019yolact}
D.~Bolya, C.~Zhou, F.~Xiao, and Y.~J. Lee, ``Yolact: Real-time instance
  segmentation,'' in \emph{Proceedings of the IEEE/CVF international conference
  on computer vision}, 2019, pp. 9157--9166.

\bibitem{jain2023oneformer}
J.~Jain, J.~Li, M.~T. Chiu, A.~Hassani, N.~Orlov, and H.~Shi, ``Oneformer: One
  transformer to rule universal image segmentation,'' in \emph{Proceedings of
  the IEEE/CVF Conference on Computer Vision and Pattern Recognition}, 2023,
  pp. 2989--2998.

\bibitem{zhang2023simple}
H.~Zhang, F.~Li, X.~Zou, S.~Liu, C.~Li, J.~Gao, J.~Yang, and L.~Zhang, ``A
  simple framework for open-vocabulary segmentation and detection,''
  \emph{arXiv preprint arXiv:2303.08131}, 2023.

\bibitem{li2023semantic}
F.~Li, H.~Zhang, P.~Sun, X.~Zou, S.~Liu, J.~Yang, C.~Li, L.~Zhang, and J.~Gao,
  ``Semantic-sam: Segment and recognize anything at any granularity,''
  \emph{arXiv preprint arXiv:2307.04767}, 2023.

\bibitem{ghiasi2022scaling}
G.~Ghiasi, X.~Gu, Y.~Cui, and T.-Y. Lin, ``Scaling open-vocabulary image
  segmentation with image-level labels,'' in \emph{European Conference on
  Computer Vision}.\hskip 1em plus 0.5em minus 0.4em\relax Springer, 2022, pp.
  540--557.

\bibitem{xu2022simple}
M.~Xu, Z.~Zhang, F.~Wei, Y.~Lin, Y.~Cao, H.~Hu, and X.~Bai, ``A simple baseline
  for open-vocabulary semantic segmentation with pre-trained vision-language
  model,'' in \emph{European Conference on Computer Vision}.\hskip 1em plus
  0.5em minus 0.4em\relax Springer, 2022, pp. 736--753.

\bibitem{liang2023open}
F.~Liang, B.~Wu, X.~Dai, K.~Li, Y.~Zhao, H.~Zhang, P.~Zhang, P.~Vajda, and
  D.~Marculescu, ``Open-vocabulary semantic segmentation with mask-adapted
  clip,'' in \emph{Proceedings of the IEEE/CVF Conference on Computer Vision
  and Pattern Recognition}, 2023, pp. 7061--7070.

\bibitem{ding2022decoupling}
J.~Ding, N.~Xue, G.-S. Xia, and D.~Dai, ``Decoupling zero-shot semantic
  segmentation,'' in \emph{Proceedings of the IEEE/CVF Conference on Computer
  Vision and Pattern Recognition}, 2022, pp. 11\,583--11\,592.

\bibitem{xu2022groupvit}
J.~Xu, S.~De~Mello, S.~Liu, W.~Byeon, T.~Breuel, J.~Kautz, and X.~Wang,
  ``Groupvit: Semantic segmentation emerges from text supervision,'' in
  \emph{Proceedings of the IEEE/CVF Conference on Computer Vision and Pattern
  Recognition}, 2022, pp. 18\,134--18\,144.

\bibitem{zhou2022extract}
C.~Zhou, C.~C. Loy, and B.~Dai, ``Extract free dense labels from clip,'' in
  \emph{European Conference on Computer Vision}.\hskip 1em plus 0.5em minus
  0.4em\relax Springer, 2022, pp. 696--712.

\bibitem{zou2023generalized}
X.~Zou, Z.-Y. Dou, J.~Yang, Z.~Gan, L.~Li, C.~Li, X.~Dai, H.~Behl, J.~Wang,
  L.~Yuan \emph{et~al.}, ``Generalized decoding for pixel, image, and
  language,'' in \emph{Proceedings of the IEEE/CVF Conference on Computer
  Vision and Pattern Recognition}, 2023, pp. 15\,116--15\,127.

\bibitem{jia2021scaling}
C.~Jia, Y.~Yang, Y.~Xia, Y.-T. Chen, Z.~Parekh, H.~Pham, Q.~Le, Y.-H. Sung,
  Z.~Li, and T.~Duerig, ``Scaling up visual and vision-language representation
  learning with noisy text supervision,'' in \emph{International conference on
  machine learning}.\hskip 1em plus 0.5em minus 0.4em\relax PMLR, 2021, pp.
  4904--4916.

\bibitem{rombach2022high}
R.~Rombach, A.~Blattmann, D.~Lorenz, P.~Esser, and B.~Ommer, ``High-resolution
  image synthesis with latent diffusion models,'' in \emph{Proceedings of the
  IEEE/CVF conference on computer vision and pattern recognition}, 2022, pp.
  10\,684--10\,695.

\bibitem{rao2022denseclip}
Y.~Rao, W.~Zhao, G.~Chen, Y.~Tang, Z.~Zhu, G.~Huang, J.~Zhou, and J.~Lu,
  ``Denseclip: Language-guided dense prediction with context-aware prompting,''
  in \emph{Proceedings of the IEEE/CVF Conference on Computer Vision and
  Pattern Recognition}, 2022, pp. 18\,082--18\,091.

\bibitem{xu2023open}
J.~Xu, S.~Liu, A.~Vahdat, W.~Byeon, X.~Wang, and S.~De~Mello, ``Open-vocabulary
  panoptic segmentation with text-to-image diffusion models,'' in
  \emph{Proceedings of the IEEE/CVF Conference on Computer Vision and Pattern
  Recognition}, 2023, pp. 2955--2966.

\bibitem{yu2023convolutions}
Q.~Yu, J.~He, X.~Deng, X.~Shen, and L.-C. Chen, ``Convolutions die hard:
  Open-vocabulary segmentation with single frozen convolutional clip,'' in
  \emph{NeurIPS}, 2023.

\bibitem{kuhn1955hungarian}
H.~W. Kuhn, ``The hungarian method for the assignment problem,'' \emph{Naval
  research logistics quarterly}, vol.~2, no. 1-2, pp. 83--97, 1955.

\bibitem{dosovitskiy2020image}
A.~Dosovitskiy, L.~Beyer, A.~Kolesnikov, D.~Weissenborn, X.~Zhai,
  T.~Unterthiner, M.~Dehghani, M.~Minderer, G.~Heigold, S.~Gelly \emph{et~al.},
  ``An image is worth 16x16 words: Transformers for image recognition at
  scale,'' \emph{arXiv preprint arXiv:2010.11929}, 2020.

\bibitem{chen2022vision}
Z.~Chen, Y.~Duan, W.~Wang, J.~He, T.~Lu, J.~Dai, and Y.~Qiao, ``Vision
  transformer adapter for dense predictions,'' \emph{arXiv preprint
  arXiv:2205.08534}, 2022.

\bibitem{milletari2016v}
F.~Milletari, N.~Navab, and S.-A. Ahmadi, ``V-net: Fully convolutional neural
  networks for volumetric medical image segmentation,'' in \emph{2016 fourth
  international conference on 3D vision (3DV)}.\hskip 1em plus 0.5em minus
  0.4em\relax Ieee, 2016, pp. 565--571.

\bibitem{loshchilov2017decoupled}
I.~Loshchilov and F.~Hutter, ``Decoupled weight decay regularization,''
  \emph{arXiv preprint arXiv:1711.05101}, 2017.

\bibitem{lin2014microsoft}
T.-Y. Lin, M.~Maire, S.~Belongie, J.~Hays, P.~Perona, D.~Ramanan,
  P.~Doll{\'a}r, and C.~L. Zitnick, ``Microsoft coco: Common objects in
  context,'' in \emph{Computer Vision--ECCV 2014: 13th European Conference,
  Zurich, Switzerland, September 6-12, 2014, Proceedings, Part V 13}.\hskip 1em
  plus 0.5em minus 0.4em\relax Springer, 2014, pp. 740--755.

\bibitem{han2022visolo}
S.~H. Han, S.~Hwang, S.~W. Oh, Y.~Park, H.~Kim, M.-J. Kim, and S.~J. Kim,
  ``Visolo: Grid-based space-time aggregation for efficient online video
  instance segmentation,'' in \emph{Proceedings of the IEEE/CVF Conference on
  Computer Vision and Pattern Recognition}, 2022, pp. 2896--2905.

\bibitem{li2023tcovis}
J.~Li, B.~Yu, Y.~Rao, J.~Zhou, and J.~Lu, ``Tcovis: Temporally consistent
  online video instance segmentation,'' in \emph{Proceedings of the IEEE/CVF
  International Conference on Computer Vision (ICCV)}, October 2023, pp.
  1097--1107.

\bibitem{yan2023universal}
B.~Yan, Y.~Jiang, J.~Wu, D.~Wang, P.~Luo, Z.~Yuan, and H.~Lu, ``Universal
  instance perception as object discovery and retrieval,'' in \emph{Proceedings
  of the IEEE/CVF Conference on Computer Vision and Pattern Recognition}, 2023,
  pp. 15\,325--15\,336.

\bibitem{abrantes2023refinevis}
A.~Abrantes, J.~Wang, P.~Chu, Q.~You, and Z.~Liu, ``Refinevis: Video instance
  segmentation with temporal attention refinement,'' \emph{arXiv preprint
  arXiv:2306.04774}, 2023.

\bibitem{wu2022efficient}
J.~Wu, S.~Yarram, H.~Liang, T.~Lan, J.~Yuan, J.~Eledath, and G.~Medioni,
  ``Efficient video instance segmentation via tracklet query and proposal,'' in
  \emph{Proceedings of the IEEE/CVF Conference on Computer Vision and Pattern
  Recognition}, 2022, pp. 959--968.

\bibitem{li2023mdqe}
M.~Li, S.~Li, W.~Xiang, and L.~Zhang, ``Mdqe: Mining discriminative query
  embeddings to segment occluded instances on challenging videos,'' in
  \emph{Proceedings of the IEEE/CVF Conference on Computer Vision and Pattern
  Recognition}, 2023, pp. 10\,524--10\,533.

\bibitem{he2016deep}
K.~He, X.~Zhang, S.~Ren, and J.~Sun, ``Deep residual learning for image
  recognition,'' in \emph{Proceedings of the IEEE conference on computer vision
  and pattern recognition}, 2016, pp. 770--778.

\bibitem{liu2021swin}
Z.~Liu, Y.~Lin, Y.~Cao, H.~Hu, Y.~Wei, Z.~Zhang, S.~Lin, and B.~Guo, ``Swin
  transformer: Hierarchical vision transformer using shifted windows,'' in
  \emph{Proceedings of the IEEE/CVF international conference on computer
  vision}, 2021, pp. 10\,012--10\,022.

\bibitem{shin2023video}
I.~Shin, D.~Kim, Q.~Yu, J.~Xie, H.-S. Kim, B.~Green, I.~S. Kweon, K.-J. Yoon,
  and L.-C. Chen, ``Video-kmax: A simple unified approach for online and
  near-online video panoptic segmentation,'' \emph{arXiv preprint
  arXiv:2304.04694}, 2023.

\bibitem{deeplabv3plus2018}
L.-C. Chen, Y.~Zhu, G.~Papandreou, F.~Schroff, and H.~Adam, ``Encoder-decoder
  with atrous separable convolution for semantic image segmentation,'' in
  \emph{ECCV}, 2018.

\bibitem{woo2021learning}
S.~Woo, D.~Kim, J.-Y. Lee, and I.~S. Kweon, ``Learning to associate every
  segment for video panoptic segmentation,'' in \emph{Proceedings of the
  IEEE/CVF Conference on Computer Vision and Pattern Recognition}, 2021, pp.
  2705--2714.

\bibitem{qiao2021vip}
S.~Qiao, Y.~Zhu, H.~Adam, A.~Yuille, and L.-C. Chen, ``Vip-deeplab: Learning
  visual perception with depth-aware video panoptic segmentation,'' in
  \emph{Proceedings of the IEEE/CVF Conference on Computer Vision and Pattern
  Recognition}, 2021, pp. 3997--4008.

\bibitem{zhou2022detecting}
X.~Zhou, R.~Girdhar, A.~Joulin, P.~Kr{\"a}henb{\"u}hl, and I.~Misra,
  ``Detecting twenty-thousand classes using image-level supervision,'' in
  \emph{European Conference on Computer Vision}.\hskip 1em plus 0.5em minus
  0.4em\relax Springer, 2022, pp. 350--368.

\bibitem{bewley2016simple}
A.~Bewley, Z.~Ge, L.~Ott, F.~Ramos, and B.~Upcroft, ``Simple online and
  realtime tracking,'' in \emph{2016 IEEE international conference on image
  processing (ICIP)}.\hskip 1em plus 0.5em minus 0.4em\relax IEEE, 2016, pp.
  3464--3468.

\bibitem{liu2022opening}
Y.~Liu, I.~E. Zulfikar, J.~Luiten, A.~Dave, D.~Ramanan, B.~Leibe,
  A.~O{\v{s}}ep, and L.~Leal-Taix{\'e}, ``Opening up open world tracking,'' in
  \emph{Proceedings of the IEEE/CVF Conference on Computer Vision and Pattern
  Recognition}, 2022, pp. 19\,045--19\,055.

\bibitem{wang2023towards}
H.~Wang, C.~Yan, S.~Wang, X.~Jiang, X.~Tang, Y.~Hu, W.~Xie, and E.~Gavves,
  ``Towards open-vocabulary video instance segmentation,'' in \emph{Proceedings
  of the IEEE/CVF International Conference on Computer Vision}, 2023, pp.
  4057--4066.

\bibitem{gupta2019lvis}
A.~Gupta, P.~Dollar, and R.~Girshick, ``Lvis: A dataset for large vocabulary
  instance segmentation,'' in \emph{Proceedings of the IEEE/CVF conference on
  computer vision and pattern recognition}, 2019, pp. 5356--5364.

\bibitem{liu2022convnet}
Z.~Liu, H.~Mao, C.-Y. Wu, C.~Feichtenhofer, T.~Darrell, and S.~Xie, ``A convnet
  for the 2020s,'' in \emph{Proceedings of the IEEE/CVF conference on computer
  vision and pattern recognition}, 2022, pp. 11\,976--11\,986.

\end{thebibliography}
}

%
%
%
%
%
%
%




\end{document}